\documentclass{amsart}
\usepackage[foot]{amsaddr}

\usepackage{graphicx}

\usepackage{multicol}
\usepackage[bottom]{footmisc}
\usepackage{bm}
\usepackage{cite}
\usepackage{url}

\usepackage[margin=3cm]{geometry}
\usepackage{lipsum}

\usepackage{float}

\newcommand{\pa}{\partial}
\newcommand{\lt}{\left}
\newcommand{\rt}{\right}

\newcommand{\etal}{\textit{et al}. }
\newcommand{\ph}{\phi}
\newcommand{\phh}{H}

\newtheorem{theorem}{Theorem}[section]

\theoremstyle{definition}

\newtheoremstyle{examplestyle}
{}{}{}{}{\bfseries}{.}{.5em}{{\thmname{#1 }}{\thmnumber{#2}}{\thmnote{ (#3)}}}
\theoremstyle{examplestyle}

\usepackage{mathtools}
\usepackage{graphicx}
\usepackage{multirow}
\usepackage[mathscr]{euscript}
\usepackage{enumitem}
\usepackage{changepage}
\usepackage{bbm}
\usepackage{array}
\usepackage{booktabs}
\usepackage[dvipsnames]{xcolor}
\usepackage{tikz}
\usepackage[export]{adjustbox}
\usepackage{pdflscape}

\newcolumntype{C}{>{\centering\arraybackslash}m{.3\textwidth}}
\usepackage{subcaption}

\usepackage[T1]{fontenc}
\usepackage{imakeidx}
\makeindex[columns=3, title =Alphabetical Index, intoc]
\usepackage{xcolor}
\usepackage{mathtools}
\usepackage{multirow}
\usepackage{diffcoeff}

\usepackage{hyperref}
\hypersetup{
	colorlinks=true,
	linkcolor=blue,
	citecolor=blue,
}
\usepackage[noabbrev,capitalize]{cleveref}

\usepackage{mdframed}

\theoremstyle{remark}
\newtheorem{remark}[theorem]{Remark}
\crefname{remark}{remark}{remarks}

\begin{document}

	\title[  ]{Enhancement of damaged-image prediction through Cahn-Hilliard Image Inpainting}
	
	\author{Jos\'e A. Carrillo}
	\address[Jos\'e A. Carrillo]{Mathematical Institute, University of Oxford, Oxford OX2 6GG, UK}
	\curraddr{}
	\email{carrillo@maths.ox.ac.uk}
	\thanks{}
	
	\author{Serafim Kalliadasis}
	\address[Serafim Kalliadasis]{Department of Chemical Engineering, Imperial College London, SW7 2AZ, UK}
	\curraddr{}
	\email{s.kalliadasis@imperial.ac.uk}
	\thanks{}
	
	\author{Fuyue Liang}
	\address[Fuyue Liang]{Department of Chemical Engineering, Imperial College London, SW7 2AZ, UK}
	\curraddr{}
	\email{fuyue.liang18@imperial.ac.uk}
	\thanks{}
	
	\author{Sergio P. Perez}
	\address[Sergio P. Perez]{Departments of Chemical Engineering and Mathematics, Imperial College London, SW7 2AZ, UK}
	\curraddr{}
	\email{sergio.perez15@imperial.ac.uk}
	\thanks{}

	\begin{abstract} We assess the benefit of including an image inpainting
		filter before passing damaged images into a classification neural network. We
		follow Bertozzi \textit{et al.} [IEEE {\bf 16} 285-291 (2007)] and we employ
		an appropriately modified Cahn-Hilliard equation as an image inpainting
		filter which is solved numerically with a finite-volume scheme exhibiting
		reduced computational cost and the properties of energy stability and
		boundedness. {The benchmark dataset employed here is MNIST, which
			consists of binary images of handwritten digits and is a standard dataset to validate image-processing methodologies}. We train a neural network based on dense
		layers with MNIST, and subsequently we contaminate the test set with damages
		of different types and intensities. We then compare the prediction accuracy
		of the neural network with and without applying the Cahn-Hilliard filter to
		the damaged images test. Our results quantify the significant improvement of
		damaged-image prediction due to applying the Cahn-Hilliard filter, which for
		specific damages can increase up to 50\% and is advantageous for low to
		moderate damage.
	\end{abstract}
	
	\maketitle
	

	%
	%
	
	\section{Introduction}\label{sec:intro}
	
	Image inpainting consists in filling damaged or missing areas of an image,
	with the ultimate objective of restoring it and making it appear as the true
	and original image. There are multiple applications of image inpainting,
	ranging from restoration of the missing areas of oil paintings and removal
	scratches in photographs to noisy MRI scans and blurred satellite images of
	the earth. Manual image inpainting techniques have been employed for many
	centuries by art conservators and professional restorers, but it was not
	until the turn of the $21^{\text{st}}$ century that digital image inpainting
	models based on PDEs and variational methods were
	introduced~\cite{caselles1998axiomatic,masnou1998level,bertalmio2000image}.
	These methods are usually referred to as non-texture, geometrical or
	structural inpainting since they focus on restoring the structural
	information in the inpainted domain such as edges, corners or curvatures.
	This is done by performing an image interpolation of the damaged areas based
	on the information collected from the surrounding environment only, leading to
	appealing images for the human vision system. On the contrary, texture
	inpainting is based on recovering global patterns of the image for the
	inpainted region \cite{criminisi2004region}, and a popular tool in this
	category is the exemplar-based inpainting
	methods\cite{criminisi2003object,lee2012robust}. Associated with these
	developments, a field that has gained a lot of traction in recent years is
	the so-called generative image inpainting, where deep learning based
	approaches have proven to be successful even for blind inpainting in which
	the inpaited region is not provided a priori
	\cite{yeh2017semantic,xie2012image,yu2018generative}. In this work we focus
	on non-texture image inpainting methods based on PDEs, and we refer the
	reader to Ref.~\cite{schonlieb2015partial} for a general review of the topic.
	
	There have been multiple PDE models for image inpainting proposed since the
	initial work of Bertalmio \etal \cite{bertalmio2000image} nearly 20
	years ago. Their trailblazing model is able to propagate isotopes, i.e.
	contours of uniform grayscale image intensity, through the inpainted region,
	a common technique employed by museum artists in restoration. As it also
	turns out that the original model bears close connection to fluid dynamics
	through the Navier-Stokes equation with the image intensity function acting
	as the stream function~\cite{bertalmio2001navier}. Another fluid dynamic
	equation that has played a pivotal role in image inpainting is the
	Cahn-Hilliard (CH) equation, initially proposed in
	\cite{cahn1958free} for phase separation in binary alloys. It has been
	employed in a wide spectrum of applications from wetting phenomena
	\cite{aymard2019linear,schmuck2012upscaled} and polymer science
	\cite{choksi2009phase} to tumour growth \cite{wu2014stabilized}. This
	equation satisfies a gradient-flow structure employing a $H^{-1}$ norm,
	\begin{equation}\label{eq:ch}
	\frac{\pa \ph (\bm{x},t)}{\pa t}=\nabla \cdot \left(M(\ph) \nabla \frac{\delta \mathcal{F}[\ph]}{\delta \ph}\right),
	\end{equation}
	where $\ph$ is an order parameter widely referred to as the phase field which
	for a binary system takes on the value $\ph=1$ in one of the phases and
	$\ph=-1$ in the other, while varying smoothly in the interface region with a width of
	$O(\epsilon)$, {with $\epsilon$ being a positive parameter related to the interface thickness coming from the derivation of diffuse-interface models \cite{cates2005physical}}. $M(\ph)$ is the mobility obtained here
	from the one-sided model $M(\ph)=1$ and $\mathcal{F}[\ph]$ is the free energy
	satisfying
	\begin{equation}\label{eq:freeenergy}
	\mathcal{F}[\ph]=\int_{\Omega}\left(\phh(\ph)+\frac{\epsilon^2}{2}|\nabla \ph |^2\right)dx,
	\end{equation}
	with the variation of the free energy denoted as $\xi$ and given by
	{\begin{equation}\label{eq:varfreeenergy}
		\xi = \frac{\delta \mathcal{F}[\ph]}{\delta \ph}= \phh'(\ph)-\epsilon^2 \Delta \ph,
		\end{equation}}
	and $\phh(\ph)$ taken here as the Ginzburg-Landau double-well potential with
	the two wells corresponding to the two phases,
	\begin{equation}\label{eq:doublewellpot}
	\phh(\ph)=\frac{1}{4}\left(\ph^2-1\right)^2\;\text{for}\;\ph \in [-1,1].
	\end{equation}
	{The boundary conditions imposed for the CH equation in \eqref{eq:ch} are no-flux for both the phase field and  for the variation of the free energy,
		\begin{equation}\label{eq:noflux}
		\epsilon^2 \nabla \ph \cdot \bm{n} = 0,\quad M(\ph) \nabla \xi \cdot \bm{n} = 0,
		\end{equation}
		where $\bm{n}$ is defined in the normal direction to and into the wall.}

	{There is a clear physical interpretation of the terms forming the free energy in \eqref{eq:freeenergy}: on the one hand, the potential $\phh(\ph)$ contains the equilibrium information of the system, and each of the minima of the choice \eqref{eq:doublewellpot} correspond to an
		equilibrium phase; on the other hand, the gradient term in \eqref{eq:freeenergy} accounts for the cost of spatial inhomogeneities, resulting in a surface tension between different phases. As a result, the equilibrum state balances the tendency of separation from the hydrophobic potential $\phh(\ph)$ and the tendency of mixing from the hydrophilic gradient term. The mobility term impacts the rate of phase separation and the coarsening process, and some works employ nonconstant degenerate mobilities that cancel when the phase field corresponds to one of the two wells in \eqref{eq:doublewellpot}
		(see e.g.~\cite{Ala-Nissila2004,Marc2012,refCH} for further discussions of the physical
		significance of the various terms of the CH equation)}.
	
	The CH equation was
	firstly proposed in the context of image inpainting by \cite{bertozzi2006inpainting}. Specifically, the
	authors adopted a modified CH equation for binary images with inpainting
	quality as accurate as the state-of-art inpainting models but with a much
	faster computational speed taking advantage of the efficient computational
	techniques already available for the CH
	equation~\cite{eyre1998unconditionally,vollmayr2003fast}. Since then several
	authors have extended the applicability of the CH equation in the field of
	image inpaiting, for instance by taking into account grayvalue images
	\cite{burger2009cahn,cherfils2017complex}, nonsmooth potentials instead of
	the double-well potential \eqref{eq:doublewellpot} \cite{bosch2014fast}, and
	considering color image inpainting \cite{cherfils2016cahn}. The modified CH
	equation in \cite{bertozzi2006inpainting} introduces a fidelity term
	$\lambda(\bm{x})$ to avoid modifying the original image outside of the
	inpainted region $D$, and the CH equation in \eqref{eq:ch} becomes
	\begin{equation}\label{eq:ch_modif}
	\frac{\pa \ph (\bm{x},t)}{\pa t}= -\nabla^{2} \lt(\epsilon^2 \nabla^{2} \ph -  \phh'(\ph) \rt) + \lambda(\textbf{x})\lt(\ph (\bm{x},t=0) - \ph\rt),
	\end{equation}
	where
	\begin{equation}\label{eq:lambda}
	\lambda(\textbf{x}) =\begin{cases} 0 & \text{if} \quad \bm{x} \in D,\\
	\lambda_{0} & \text{if} \quad \bm{x} \notin D, \end{cases}
	\end{equation}
	and $\ph (\bm{x},t=0)$ refers to the original damaged image. The parameter
	$\epsilon$ plays a similar role as in the original CH equation, and here it
	is related to the interface between the two phases or colours presented in
	the image. The two parameters $\epsilon$ and $\lambda_{0}$ are essential to
	achieve an adequate image inpainting outcome, and it is usually necessary to
	iterate until finding appropriate tunings for their values, which typically
	depend on the image specifications. {Furthermore, the new modified CH equation \eqref{eq:ch_modif} is not stricly a gradient flow:
		although the original CH equation satisfies a gradient-flow structure under an $H^{-1}$ norm and the fidelity term in \eqref{eq:ch_modif} can be derived from a gradient flow
		under an $L^{2}$ norm, the combined modified CH equation is neither a gradient flow in $H^{-1}$ nor $L^{2}$.}
	
	{One of the main advantages of employing the CH
		equation for image inpainting is the myriad of fast and reliable numerical
		methods available for its solution.} A pivotal contribution was the
	convex-splitting scheme developed by Eyre \cite{eyre1998unconditionally} which is
	unconditionally energy-stable by treating as implicit the convex
	terms of the free energy in \eqref{eq:freeenergy}, while keeping the concave
	terms explicit. The design of energy-stable and maximum-principle satisfying
	schemes for the CH equation has been a really active area of research
	\cite{tierra2015numerical}, and several authors have proposed schemes based
	on finite differences \cite{furihata2001stable,guo2016h2,wise2009energy},
	finite elements \cite{barrett1999finite,elliott1989cahn}, finite volumes
	\cite{cueto2008time} or discontinuous Galerkin
	\cite{xia2007local,choo2005discontinuous,wells2006discontinuous,liu2015stabilized}.
	These schemes have also proven effective for degenerate mobilities  or
	logarithmic potentials. Schemes satisfying the maximum principle condition
	for specific choices of free energy have also been constructed
	\cite{chen2019positivity}. We refer the reader to \cite{shen2019new} for
	a recent work discussing the state-of-the-art numerical techniques for
	nonlinear gradient flows.

	{In a recent effort~\cite{refCH} we constructed a robust semi-implicit
		finite-volume scheme for the CH equation that offers crucial advantages when
		applied to the field of image inpainting:}
	
	\begin{itemize}
		\item {Firstly, finite volumes are a straightforward discretization
			when dealing with images, which often consist of rectangular-shaped
			pixel cells with an average color intensity. This is exactly the
			starting point of finite-volume schemes, and as a result it is
			conceptually simpler to apply finite volumes in comparison with finite
			elements, finite differences or discontinuous Galerkin (which would be
			more suitable for other more complex and rare pixel shapes such as
			triangular ones).}
		
		\item {Secondly, our scheme is based on a
			dimensional-splitting approach: instead of solving the full
			two-dimensional (2D) image altogether, this technique initially solves row
			by row and then column by column. This has a massive benefit in
			computational cost, which is reduced from
			$\mathcal{O}(N^{d\gamma})$
			for an image with $N$ cells in $d$ dimensions to $\mathcal{O}(d N^{d+\gamma-1})$, with
			$2<\gamma<3$. The reason for this is that the cost of inverting a
			$N\times N$ matrix is $\mathcal{O}(N^{d\gamma})$, with a value of
			$2<\gamma<3$ that slightly varies depending on the inversion algorithm and
			matrix structure (see \cite{coppersmith1987matrix} for details). For images with $N$ cells per
			dimension, the solution of the full 2D scheme involves inverting a
			$N^d\times N^d$ Jacobian matrix, with a subsequent cost of
			$\mathcal{O}(N^{d\gamma})$. In contrast, the dimensional-splitting
			technique requires inverting $dN^{d-1}$ Jacobians of size $N\times N$,
			amounting for a total computational cost of $\mathcal{O}(d
			N^{d+\gamma-1})$.  This is already
			advantageous for a 2D image, and the computational cost is further reduced
			for high-dimensional images, such as the ones for \cite{bosch2015fractional} or X-ray computed tomography \cite{gu2006x} in
			medical image analysis. To add more, such dimensional-splitting
			technique allows for parallelization, and it is possible to half the total
			computational cost by solving nonadjacent rows and columns in parrallel. }
		
		\item {Thirdly, our scheme has been extensively tested in \cite{refCH} for challenging configurations of the original CH equation \eqref{eq:ch}. In addition, in \cite{refCH} we
			prove that the scheme unconditionally satisfies the discrete decay of
			the free energy for different choices of potentials \eqref{eq:doublewellpot}
			\cite{bosch2014fast}, while at the same time we prove the phase-field boundedness for
			mobilities of the form $M(\ph)=1-\ph^2$. Even though the modified CH
			equation \eqref{eq:ch_modif} is not strictly a gradient flow due to the inclusion of the
			fidelity term, our scheme preserves its robustness for all the
			image-inpainting test cases presented in this work.}
	\end{itemize}
	{The combination of these properties
		and reduced computational cost, together with the versatility of finite
		volumes, make our scheme efficient and robust for the solution of
		the modified CH equation in
		\eqref{eq:ch_modif} for a variety of applications in image inpainting.}
	
	The objective of this work is to show precisely the applicability of our
	numerical framework in \cite{refCH} for a benchmark dataset of images in need
	of restoration through image inpainting. For this task we purposely add
	different types and intensities of damage to the popular MNIST dataset
	\cite{deng2012mnist}, and then apply image inpainting by solving the modified
	CH equation \eqref{eq:ch_modif} with our finite-volume scheme in
	\cite{refCH}. { The MNIST database (Modified National Institute of Standards and Technology database) is a standard dataset to validate image-processing methodologies, and it contains binary handwritten images of numbers from zero to nine. We choose this dataset since the modified CH equation \eqref{eq:ch_modif} is applicable to binary images such as the ones in the MNIST dataset.  The extension of this work to non-binary images is also relevant and will be explored elsewhere.} We also assess the improvement in pattern recognition accuracy
	of the restored MNIST images, and for this we construct a neural network for
	the task of classification. A key objective of our study is to
	quantify the benefits of including a CH filter before introducing a damaged
	image into a neural network. Our results demonstrate that accuracies in
	classification can increase up to 50\% for particular damages in the images,
	and, in general, applying the CH filter improves the accuracy prediction for
	a wide range of low to moderate image damage. {This increase in accuracy is obtained by applying our image-inpainting methodology to the MNIST dataset exclusively, and as a disclaimer we remark that other datasets or types of damage may result in different increases of accuracy. The application of other types of filters, such as texture inpainting or generative inpainting, may also result in different increases of accuracy, but overall the accuracy should be higher compared to the case where no filter is applied to the damaged image.} {The code to reproduce the results of this work is available at \cite{GithubSergio}}.
	
	In \autoref{sec:methodo} we outline the methodology: in
	\autoref{subsec:fvm_mod} we adapt our finite-volume scheme in \cite{refCH}
	for the modified CH equation in \eqref{eq:ch_modif}; in
	\autoref{subsec:two-step method} we recall the two-step method for image
	inpainting in \cite{bertozzi2006inpainting}; in \autoref{subsec:neural
		network} we detail the neural network architecture for the classification
	task; and lastly in \autoref{subsec:Integrated} we explain the structure of
	the integrated algorithm which takes a damaged image, applies a CH filter to
	it, and then classifies the image through a neural network. Subsequently in
	\autoref{sec:results} we present the results of the integrated algorithm
	applied to the MNIST dataset: in \autoref{subsec:crossline} we begin by
	identofying appropiate tunings for the values of $\epsilon$ and $\lambda_0$; in
	\autoref{subsec:damage} we present the different types of damage introduced
	into the MNIST testset of images; and finally in \autoref{subsec:prediction}
	we quantify the improvement in accuracy of applying the CH filter to the
	damaged MNIST images before introducing them into the neural network. A
	discussion and final remarks are offered in \autoref{sec:con_fwork}.

	%
	%
	
	\section{Integrated algorithm with image inpainting and pattern recognition}\label{sec:methodo}
	
	We detail the construction of an integrated algorithm that firstly applies
	image inpainting and subsequently conducts pattern recognition for the restored image.
	In \autoref{subsec:fvm_mod} we begin by presenting the finite-volume scheme
	employed to solve the modified CH equation, based on the work in
	\cite{refCH}. Then in \autoref{subsec:two-step method} we illustrate the
	two-step method for image inpainting, based on tuning the parameters
	$\epsilon$ and $\lambda_0$ of the modified CH equation. In
	\autoref{subsec:neural network} we present  the neural network employed for
	pattern recognition, detailing its architecture and training parameters.
	Finally, in \autoref{subsec:Integrated} we gather all previous elements to
	formulate an integrated algorithm for prediction with an image-inpainting
	filter.
	
	%
	%
	
	\subsection{2D finite-volume scheme for the modified CH equation}\label{subsec:fvm_mod}
	
	We summarise the 2D finite-volume scheme constructed for the original CH
	equation in our previous work \cite{refCH}. {This scheme satisfies an unconditional decay of the
		discrete free energy of the original CH equation in \eqref{eq:freeenergy}, for no matter what choice of the time step, and for specific choice of mobility $M(\phi)=1-\phi^2$ it ensures the
		unconditional boundedness of the phase field}. {The scheme can be straightforwardly
		extended to the modified CH equation in \eqref{eq:ch_modif} proposed in
		\cite{bertozzi2006inpainting} as we show here. Even though the modified CH equation does not possess some of the properties of the original CH equation, such as
		the gradient-flow structure, our finite-volume scheme preserves its robustness for all the image-inpainting test cases presented in this work. In \cref{rem:deltat}
		we explain how to choose the time step and the mesh size, and in
		\cref{rem:dim} we detail how to turn the scheme into a dimensional splitting
		one, with promising applicability in high-dimensional images such as medical ones. We refer the reader to \cite{refCH, bailo2018fully} for further details about
		dimensional-splitting schemes.}

	{For simplicity, let's rewrite \eqref{eq:ch_modif} in 2D by introducing $\bm{u} = (v, w)$ as the physical velocity term and $\xi$ as the variation of the free energy with respect to the density,
		\begin{equation}\label{eq:ch_modif_fv}
		\frac{\pa \ph (x,y,t)}{\pa t}= -\nabla \cdot \bm{u} + \lambda(x,y)\lt(\ph (x,y,t=0) - \ph(x,y,t)\rt),
		\end{equation}
		where
		\begin{equation}\label{eq:free_energy_fv}
		\xi = \frac{\delta \mathcal{F}[\ph]}{\delta \ph}= \epsilon^2 \nabla^{2} \ph -  \phh'(\ph), \quad
		\bm{u}=\nabla \xi, \quad
		v = \frac{\partial \xi}{\partial x}, \quad \text{and} \quad   w = \frac{\partial \xi}{\partial y}.
		\end{equation}}
	For the finite-volume formulation we begin by dividing the computational domain $[0,L]\times [0,L]$ in $N\times N$
	cells $C_{i,j}:=[x_{i-1/2},x_{i+1/2}]\times[y_{j-1/2},y_{j+1/2}]$, all with uniform size $\Delta x \Delta y$
	so that $x_{i+1/2}-x_{i-1/2}=\Delta x$ and $y_{j+1/2}-y_{j-1/2}=\Delta y$. {The time step is denoted as $\Delta t$. In each of the cells we define the cell average $\ph_{i,j}^n$ at time $t=n\Delta t$ as
		\begin{equation}
		\ph_{i,j}^n=\frac{1}{\Delta x \Delta y}\int_{C_{i,j}}\ph(x,y,t=n \Delta t)\,dx\,dy,
		\end{equation}
		where $\ph_{i,j}^{0}$ is the phase field at $t=0$, which corresponds to the normalized pixel intensities of the initial damaged image to be inpainted.}

	The finite-volume scheme is then derived by integrating the modified CH equation \eqref{eq:ch_modif_fv}
	over each of the cells $C_{i,j}$ of the domain, leading to an implicit formulation satisfying
	\begin{equation}\label{eq:fv_2}
	\frac{\ph_{i,j}^{n+1}-\ph_{i,j}^{n}}{\Delta t}=-\frac{F_{i+\frac{1}{2},j}^{n+1}-F_{i-\frac{1}{2},j}^{n+1}}{\Delta x}-\frac{G_{i,j+\frac{1}{2}}^{n+1}-G_{i,j-\frac{1}{2}}^{n+1}}{\Delta y}+\lambda_{i,j}(\ph_{i,j}^{0} - \ph_{i,j}^{n}),
	\end{equation}
	{with $F_{i+1/2,j}^{n+1}$ and $G_{i,j+1/2}^{n+1}$ being flux approximations at the boundaries},
	and $\lambda_{i,j}$ the discrete version of $\lambda(\bm x)$ in \eqref{eq:lambda} satisfying
	\begin{equation}\label{eq:lambda_dis}
	\lambda_{i,j}=
	\begin{cases} 0 & \text{if } (x_i,y_j)\in D, \\
	\lambda_0 & \text{if }(x_i,y_j)\notin D,\\
	\end{cases}
	\end{equation}
	{with $D$ being the inpainted domain where the image damage is located. The impainted domain has to be determined beforehand for each image, and is formed by those finite-volume cells
		to be repaired during the image inpainting.}

	{The flux terms $F_{i+1/2,j}^{n+1}$ and $G_{i,j+1/2}^{n+1}$ are the discrete approximations of the velocity components $v$ and $w$ at the cell interfaces $\lt(x_{i+1/2},y_j\rt)$ and $\lt(x_{i},y_{j+1/2}\rt)$, respectively}. Such approximations
	follow an upwind and implicit approach inspired by
	the works in Refs~\cite{carrillo2015finite, bailo2018fully, bessemoulin2012finite, refCH}, satisfying
	\begin{equation}\label{eq:flux_2}
	\begin{gathered}
	F_{i+\frac{1}{2},j}^{n+1}=\left(v^{n+1}_{i+1/2,j}\right)^++\left(v^{n+1}_{i+1/2,j}\right)^-,\\ G_{i,j+\frac{1}{2}}^{n+1}=\left(w^{n+1}_{i,j+1/2}\right)^++\left(w^{n+1}_{i,j+1/2}\right)^-,
	\end{gathered}
	\end{equation}
	where the velocities $v^{n+1}_{i+\frac{1}{2},j}$ and $w^{n+1}_{i,j+\frac{1}{2}}$ are discretized from \eqref{eq:free_energy_fv} as
	{\begin{equation}\label{eq:vel_2}
		v^{n+1}_{i+\frac{1}{2},j}=-\frac{\xi^{n+1}_{i+1,j}-\xi^{n+1}_{i,j}}{\Delta x},\quad w^{n+1}_{i,j+\frac{1}{2}}=-\frac{\xi^{n+1}_{i,j+1}-\xi^{n+1}_{i,j}}{\Delta y},
		\end{equation}}
	where $\xi^{n+1}_{i,j}$ is the discretized variation of the free energy defined in \eqref{eq:free_energy_fv}. The upwind approach in \eqref{eq:flux_2} follows from
	\begin{equation}\label{eq:velpm_2}
	\begin{gathered}
	\left(v^{n+1}_{i+1/2,j}\right)^+=\max(v^{n+1}_{i+1/2,j},0),\quad \left(v^{n+1}_{i+1/2,j}\right)^-=\min(v^{n+1}_{i+1/2,j},0),\\
	\left(w^{n+1}_{i,j+1/2}\right)^+=\max(w^{n+1}_{i,j+1/2},0),\quad \left(w^{n+1}_{i,j+1/2}\right)^-=\min(w^{n+1}_{i,j+1/2},0).
	\end{gathered}
	\end{equation}
	
	The discretized variation of the free energy $\xi_{i,j}^{n+1}$ in \eqref{eq:free_energy_fv} is computed with a semi-implicit scheme inspired by
	the ideas of \cite{eyre1998unconditionally, vollmayr2003fast}, where the so-called convexity splitting scheme is proposed to construct
	unconditional gradient-stable schemes (i.e. schemes that ensure the decay of the discrete version of the free energy in
	\eqref{eq:freeenergy}). In our recent effort \cite{refCH} we show that our finite-volume scheme unconditionally decreases the discrete free energy of the
	CH equation if the contractive part of the potential, $\phh_c(\ph)$, is taken as implicit; the expansive part of the potential,
	$\phh_e(\ph)$, is taken as explicit; and the Laplacian is taken as an average between the explicit and the implicit second-order
	discretizations, so that
	\begin{equation}\label{eq:varsemi2}
	\begin{gathered}
	\xi_{i,j}^{n+1}=\phh_c'(\ph_{i,j}^{n+1})-\phh_e'(\ph_{i,j}^{n})-\frac{\epsilon^2}{2} \left[(\Delta \ph)^{n}_{i,j}+(\Delta \ph)^{n+1}_{i,j}\right] ,\\
	\phh(\ph)=\phh_c(\ph)-\phh_e(\ph)=\frac{\ph^4+1}{4}-\frac{\ph^2}{2},
	\end{gathered}
	\end{equation}
	{where the potential $\phh(\ph)$ in \eqref{eq:doublewellpot} is decomposed into two convex functions, $\phh_c(\ph)$ and $\phh_e(\ph)$,
		\begin{equation*}
		\phh(\ph)=\phh_c(\ph)-\phh_e(\ph)=\frac{\ph^4+1}{4}-\frac{\ph^2}{2}.
		\end{equation*}}
	{The discrete two-dimensional approximation of the Laplacian $(\Delta \phi)^n_{i,j}$ is chosen to satisfy the second-order form
		\begin{equation}
		(\Delta \ph)_{i,j}^{n}\coloneqq\frac{\ph_{i+1,j}^{n}=-2\ph^{n}_{i,j}+\ph^{n}_{i-1,j}}{\Delta x ^2}+\frac{\ph_{i,j+1}^{n}-2\ph^{n}_{i,j}+\ph^{n}_{i,j-1}}{\Delta y ^2}.
		\end{equation}}
	
	{The modified CH equation in \eqref{eq:ch_modif} employs the no-flux boundary conditions defined in \eqref{eq:noflux}. The numerical implementation of the boundary conditions follows from
		\begin{equation}\label{eq:nofluxnum2}
		\begin{gathered}
		\ph_{i-1,j}^{n}=\ph_{i,j}^{n} \; \text{for}\;i=1,\forall j;\quad \ph_{i+1,j}^{n}=\ph_{i,j}^{n} \; \text{for}\;i=N,\forall j;\\
		\ph_{i,j-1}^{n}=\ph_{i,j}^{n} \; \text{for}\;j=1,\forall i;\quad \ph_{i,j+1}^{n}=\ph_{i,j}^{n} \; \text{for}\;j=N,\forall i;\\
		F_{i-\frac{1}{2},j}^{n+1}=0 \; \text{for}\;i=1,\forall j;\quad F_{i+\frac{1}{2},j}^{n+1}=0 \; \text{for}\;i=N,\forall j;\\
		G_{i,j-\frac{1}{2}}^{n+1}=0 \; \text{for}\;j=1,\forall i;\quad G_{i,j+\frac{1}{2}}^{n+1}=0 \; \text{for}\;j=N,\forall i.
		\end{gathered}
		\end{equation}}
	
	\begin{remark}[{Choice of time step $\Delta t$ and mesh size $\Delta x = \Delta y$}]\label{rem:deltat} {Explicit finite-volume schemes are often stable under a CFL condition imposed over the time step and mesh size. The implicit scheme presented in \autoref{subsec:fvm_mod} does not need of a CFL condition to decay the discrete free energy of the original Cahn-Hilliard equation in \eqref{eq:ch}, as proved in \cite{refCH}. This doesn't mean however that any $\Delta t$ works in practice: in each time step the nonlinear system \eqref{eq:fv_2} has to be solved by iteration, and with a large $\Delta t$ the convergence is likely to fail. In this work we have tested that with the choice of $\Delta t=0.1$ and $\Delta x = \Delta y=1$ our simulations converge to the right solution for the simulations presented in \autoref{sec:results}.}

	\end{remark}

	\begin{remark}[{Dimensional-splitting scheme and parallelisation}]\label{rem:dim}
		{The full 2D finite-volume scheme presented in \autoref{subsec:fvm_mod} can be reformulated by employing a dimensional-splitting methodology: instead of solving the full 2D image altogether, this technique initially solves row by row and then column by column. The detailed construction of such scheme is presented in \cite{refCH} and is based on \cite{bailo2018fully}, and here we briefly illustrate how it works.}
		
		{For the dimensional-splitting approach we firstly update the
			solution along the $x$ axis, for each index $j$ corresponding to a fix value
			of $y_{j}$ where $j\in[1,N]$. Subsequently, we proceed in the same way along
			the $y$ axis, for each index $i$ corresponding to a fixed value a value of
			$x_{i}$ where $i\in[1,N]$. The index $r$, where $r\in[1,N]$, denotes the
			index $j$ of the fixed $y_{j}$ value in every $x$ axis of the first loop, and
			the updated average density for each $x$ axis with $j=r$ is
			$\ph_{i,j}^{n,r}$. Similarly, the index $c\in[1,N]$ denotes the index $i$ for
			every fixed value of $x_{j}$ in each $y$ axis of the second loop, and the
			updated density for each $y$ axis with $i=c$ is $\ph_{i,j}^{n,c}$.}
		
		{In the first place we go through each of the $x$ axes of the domain
			at a fixed $y_{j}$ with $j=r$. The initial conditions for the scheme are
			$\ph_{i,j}^{n,0}\coloneqq \ph_{i,j}^{n}$. The scheme for each $x$ along the
			loop $r=1,\ldots,N$ satisfies
			\begin{equation}
			\ph_{i,j}^{n,r}= \begin{cases}  \ph_{i,j}^{n,r-1}-\frac{\Delta t}{\Delta x} \lt(F_{i+1/2,j}^{n,r}-F_{i-\frac{1}{2},j}^{n,r}\rt) +\lambda_{i,j}(\ph_{i,j}^{0} - \ph_{i,j}^{n,r-1})& \text{if } j=r; \\ \ph_{i,j}^{n,r-1} & \text{otherwise}; \end{cases}
			\end{equation}
			with $F_{i+1/2,j}^{n,r}$ computed in a similar fashion as in
			\eqref{eq:flux_2}. Once the loop for the rows is completed, we define the
			intermediate density values as $\rho^{n+1/2} \coloneqq \rho^{n,N}$.
			Subsequently, we continue through each of the $y$ axes with index
			$c=1,\ldots,N$, each of them at a fixed $x_{i}$ with $i=c$. The initial
			condition for this scheme is $\ph_{i,j}^{n,0}\coloneqq \ph_{i,j}^{n+1/2}$.
			The scheme for each $y$ along the loop $c=1,\ldots,N$ satisfies:
			\begin{equation}
			\ph_{i,j}^{n,c}= \begin{cases}  \ph_{i,j}^{n,c-1}-\frac{\Delta t}{\Delta y} \lt(G_{i,j+1/2}^{n,c}-G_{i,j-\frac{1}{2}}^{n,c}\rt) +\lambda_{i,j}(\ph_{i,j}^{0} - \ph_{i,j}^{n,c-1})& \text{if } i=c; \\ \ph_{i,j}^{n,c-1} & \text{otherwise}. \end{cases}
			\end{equation}
			Once the loop for the columns is completed, we define the final density
			values $\ph_{i,j}^{n+1}$ after a discrete timestep $\Delta t$  as
			$\ph_{i,j}^{n+1} \coloneqq \ph_{i,j}^{n,N}$. This dimensional-splitting
			scheme can be fully parallelized in order to save computational time. This is
			because: (i) the scheme does not take notice the order of updating the
			rows/columns, as long as all of them are updated; (ii) one row or column only
			depends on the values of the directly adjacent rows or columns respectively.
			As a result, a strategy to parallelize the dimensional splitting scheme
			consists of updating at the same time all the odd rows/columns, since they do
			not depend on one another. One can then proceed with all the even
			rows/columns at the same time."}

	\end{remark}
	%
	%
	
	\subsection{Two-step method for the modified CH equation}\label{subsec:two-step method}
	{The two-step method followed here was firstly proposed in \cite{bertozzi2006inpainting}.
		It basically consists of applying the finite-volume scheme
		in \autoref{subsec:fvm_mod} twice with different values of the parameter $\epsilon$}. The first stage consists of taking a
	large $\epsilon$ to execute a large-scale topological reconnection of shapes,
	leading to images with diffused edges. Subsequently, and in order to sharpen
	the edges after the first stage, $\epsilon$ is substantially reduced, and the final
	outcome becomes less blurry and diffused. We denote the corresponding values
	of $\epsilon$ as $\epsilon_1$ and $\epsilon_2$.
	
	Adequately tuning the two values of $\epsilon$ in each stage, as well as $\lambda$, is vital to complete a successful image inpainting.
	Those values have to be chosen empirically and depend on the dataset and type of damage, and in \autoref{subsec:crossline} we conduct a study
	to select them {(with a sensitivity analysis provided in \autoref{app:sen})}. As explained there, the appropriate values for $\epsilon$ are between $0.5$ and $1.5$ for MNIST-like images,
	while $\lambda\in[1,1000]$. The cell sizes are $\Delta x=\Delta y=1$. The reader can find the exact values employed after the analysis
	of \autoref{subsec:crossline} in \autoref{tab:parameters}.
	
	%
	%
	\subsection{Neural network architecture for classification}\label{subsec:neural network}
	
	The prediction of the label in the restored images is performed via a neural network constructed in TensorFlow  \cite{TensorFlow}.
	Its architecture is defined taking into account that in this work we employ the MNIST dataset \cite{deng2012mnist},
	which contains binary images of digits from $0$ to $9$ and has a resolution of $28\times28$ pixels.
	This is a benchmark dataset in the community and is the de facto ``hello world" dataset of computer vision.
	There are consequently plenty of neural network architectures attaining extremely high accuracies for the MNIST dataset,
	and we refer the reader to the Kaggle competition of Digit Recognizer in \cite{Kaggle} for examples of such architectures.
	
	Here, however, our overarching objective aim is to quantify how the prediction of damaged images is
	enhanced once the CH filter is applied to the images beforehand. Hence we do not
	require a highly sophisticated neural network and a cutting-edge architecture
	as in computer vision: our images are not going to be exactly the same
	as in the training set due to the damage and the subsequent restoration. We
	then select a standard architecture for classification based on sequential
	dense layers. Such architecture is formed by:
	
	\begin{enumerate}[label=\arabic*)]
		\item A flatten layer that takes the $28\times28$ image input and turns it into an array with $784$ elements. There are no weights to optimize in this layer.
		\item {A dense layer with $64$ units and the ReLU activation function, defined as $f(x)=\max\{0,x\}$ for $x\in\mathbb{R}$}. There are $784 \times 64$ weights to optimize in this layer, in addition to the bias term in each of the 64 units.
		\item Another dense layer with $64$ units and the ReLU activation function. There are $64 \times 64$ weights to optimize in this layer, in addition to the bias term in each of the 64 units.
		\item A final dense layer with $10$ units and the softmax activation function, which returns the normalized probability distribution for the 10 labels and satisfies $\sigma (z_i)=\exp (z_i)/ \sum_{j=1}^{10} \exp(z_j)$, with $z=(z_1,\ldots,z_{10})$ being the output of the final dense layer with 10 units. There are $64\times 10$ weights to optimize in this layer, in addition to the bias term in each of the 10 units.
	\end{enumerate}
	
\begin{figure}[h]
		\centering
		
		\includegraphics[scale=0.37]{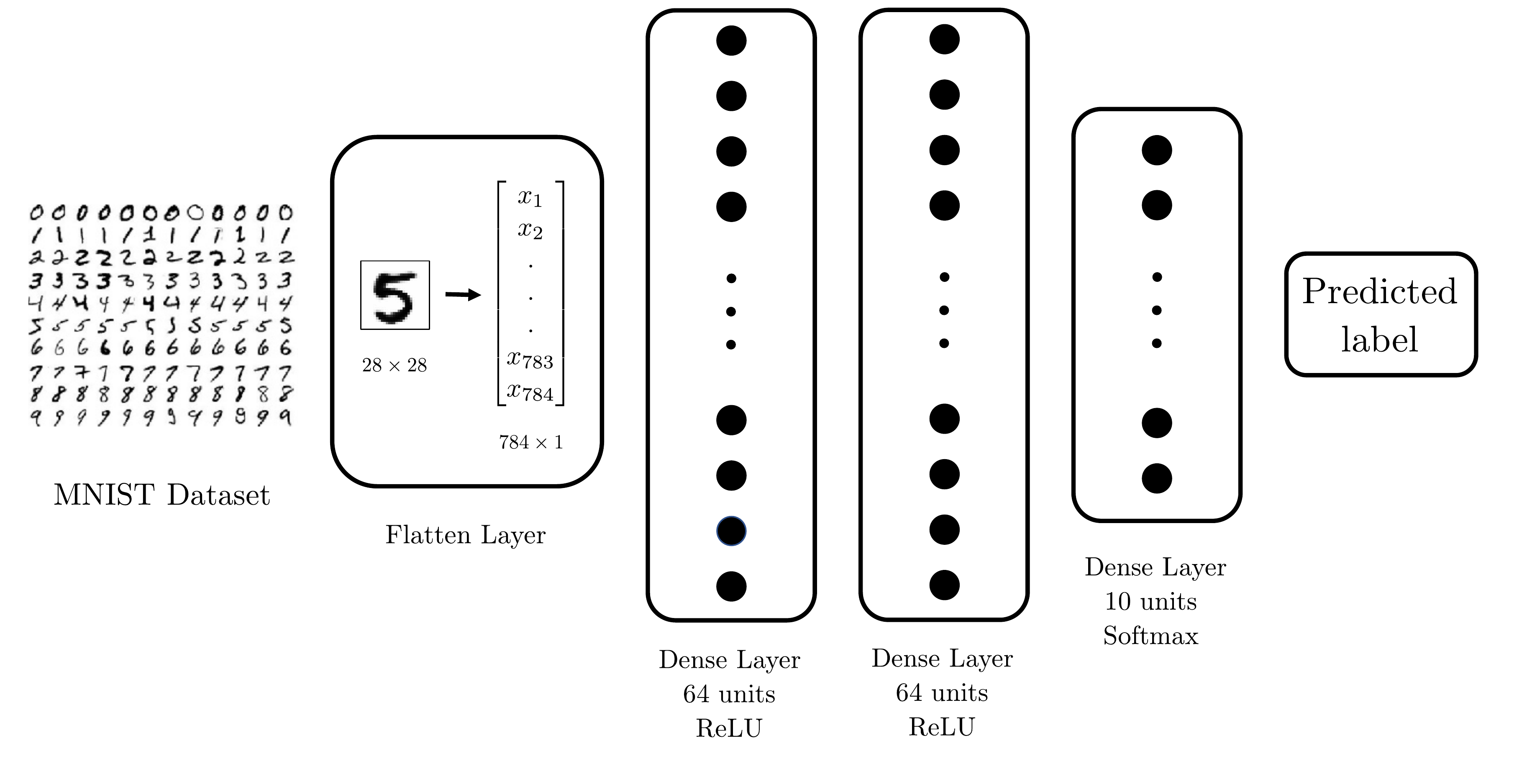}
		\caption{Diagram showing the layers of the neural network of \autoref{subsec:neural network}.}
		\label{fig:NN}
	\end{figure}

	For the training of this network we initially divide the original MNIST dataset in 60000 training images and 10000 testing images. Then we train the neural network for 10 epochs with the Adam optimizer, choosing as loss function the categorical crossentropy defined as
	\begin{equation*}
	J(w)=\frac{1}{N}\sum_{i=1}^N \lt[y_i \log (\hat{y}_i) + (1-y_i) \log(1-\hat{y}_i)\rt],
	\end{equation*}
	with $w$ being the weights to optimize, $y_i$ each of the $N$ true labels
	of the training dataset, and $\hat{y}_i$ each of the $N$ predicted labels.
	After 10 epochs we get an accuracy for the training dataset of $99.02\%$,
	while for the test set the accuracy is $97.47\%$. Once the neural network is
	trained we keep the weights fixed for the comparison of damaged and restored
	images in \autoref{sec:results}. A display of the neural network is depicted
	in \autoref{fig:NN}.
	
	%
	%
	
	\subsection{Integrated Algorithm}\label{subsec:Integrated}
	
	The integrated algorithm proposed and tested in this work takes as input
	a damaged image, applies the CH filter based on \autoref{subsec:fvm_mod}
	and \autoref{subsec:two-step method} to restore it, and finally applies the
	already-trained neural network in \autoref{subsec:neural network} to predict
	its label.
	
\begin{figure}[h]
		\centering
		
		\includegraphics[scale=0.7]{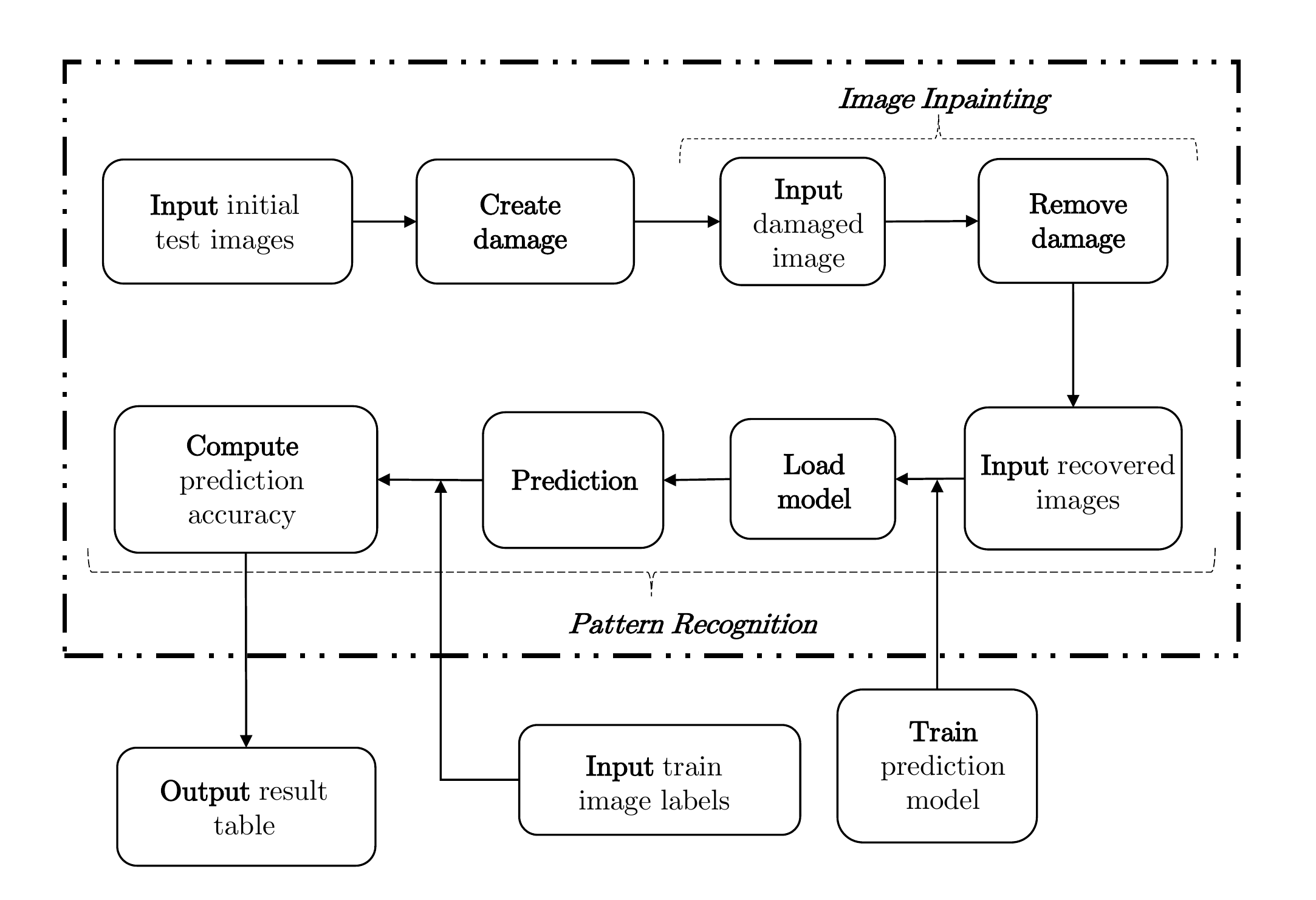}
		\caption{A schematic representation to show the applicability of the integrated algorithm.}
		\label{fig:flow_chart}
	\end{figure}

	To show the applicability of this integrated algorithm we initially create damage in the images of the test set in
	the MNIST dataset \cite{deng2012mnist}. After we apply the image inpainting to the damaged test images,
	we introduce the restored images in the neural network. At that point, and since we have the true labels of the test set,
	we can assess the attained accuracy in comparison to directly introducing the damaged images or the original images into the neural network.
	This procedure is conducted for multiple types of damage in \autoref{sec:results}, and a schematic representation of
	all steps is depicted in \autoref{fig:flow_chart}.
	
	%
	%

	\section{Application of the integrated algorithm to the MNIST dataset}\label{sec:results}
	
	Our focus here is testing the applicability of the integrated algorithm in
	\autoref{subsec:Integrated} to increase predictability in damaged images. In
	\autoref{subsec:crossline} we start by analysing the impact of the parameters
	$\lambda_0$ and $\epsilon$ on the inpainting process, with the objective of
	calibrating them before employing the MNIST dataset. In
	\autoref{subsec:damage} we detail the types of damage that we insert into the
	MNIST dataset, and we also show the restored outcomes of applying the CH
	equation as an image inpainting filter. Finally, in
	\autoref{subsec:prediction} we evaluate how the accuracy of the damaged
	images increases after applying the CH filter to them, for various types and
	degrees of damage.
	
	{The code to reproduce the results of this work is available at \cite{GithubSergio}}.
	%
	%
	
	\subsection{Inpainting of a Crossline}\label{subsec:crossline}
\begin{figure}[h]
		\centering
		\begin{minipage}{.3\textwidth}
			\includegraphics[width=\linewidth, height=4cm]{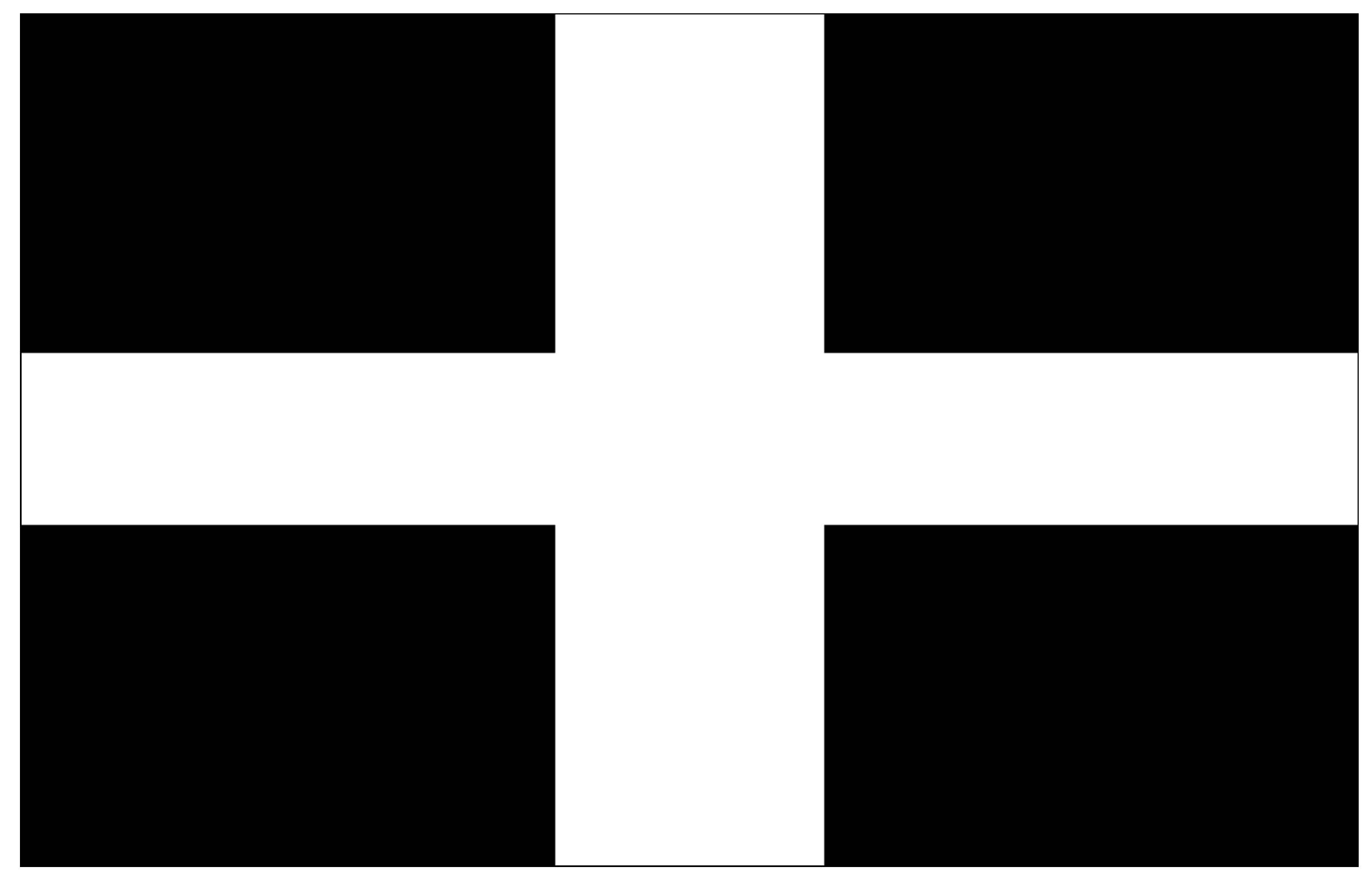}
			\subcaption{Initial crossline}
			\label{fig: initialcrossline}
		\end{minipage}
		\hfill
		\begin{minipage}{0.3\textwidth}
			\includegraphics[width=1\linewidth, height=4cm]{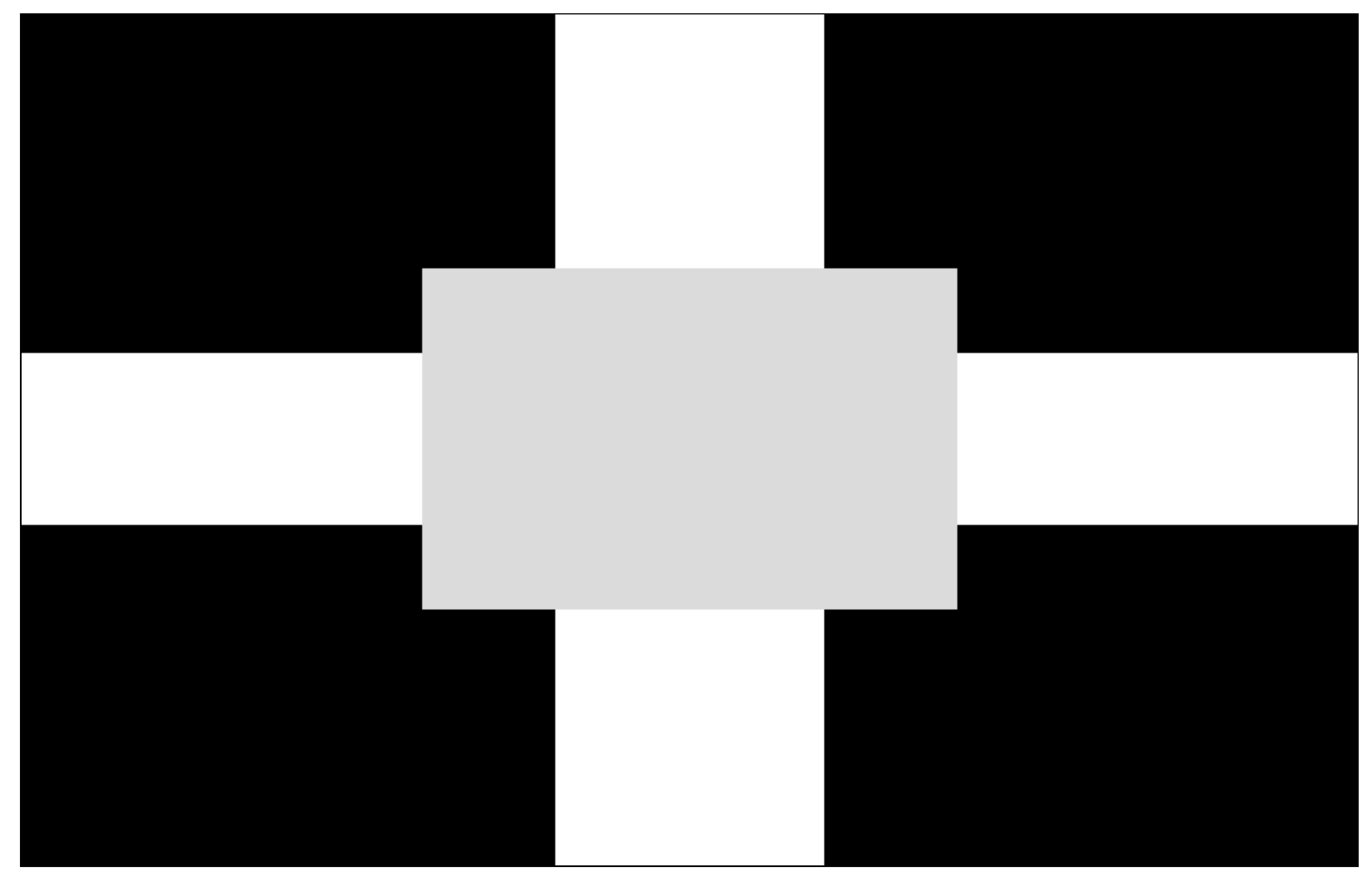}
			\subcaption{Damaged crossline}
			\label{fig:damagedcrossline}
		\end{minipage}
		\hfill
		\begin{minipage}{0.3\textwidth}
			\includegraphics[width=1\linewidth, height=4cm]{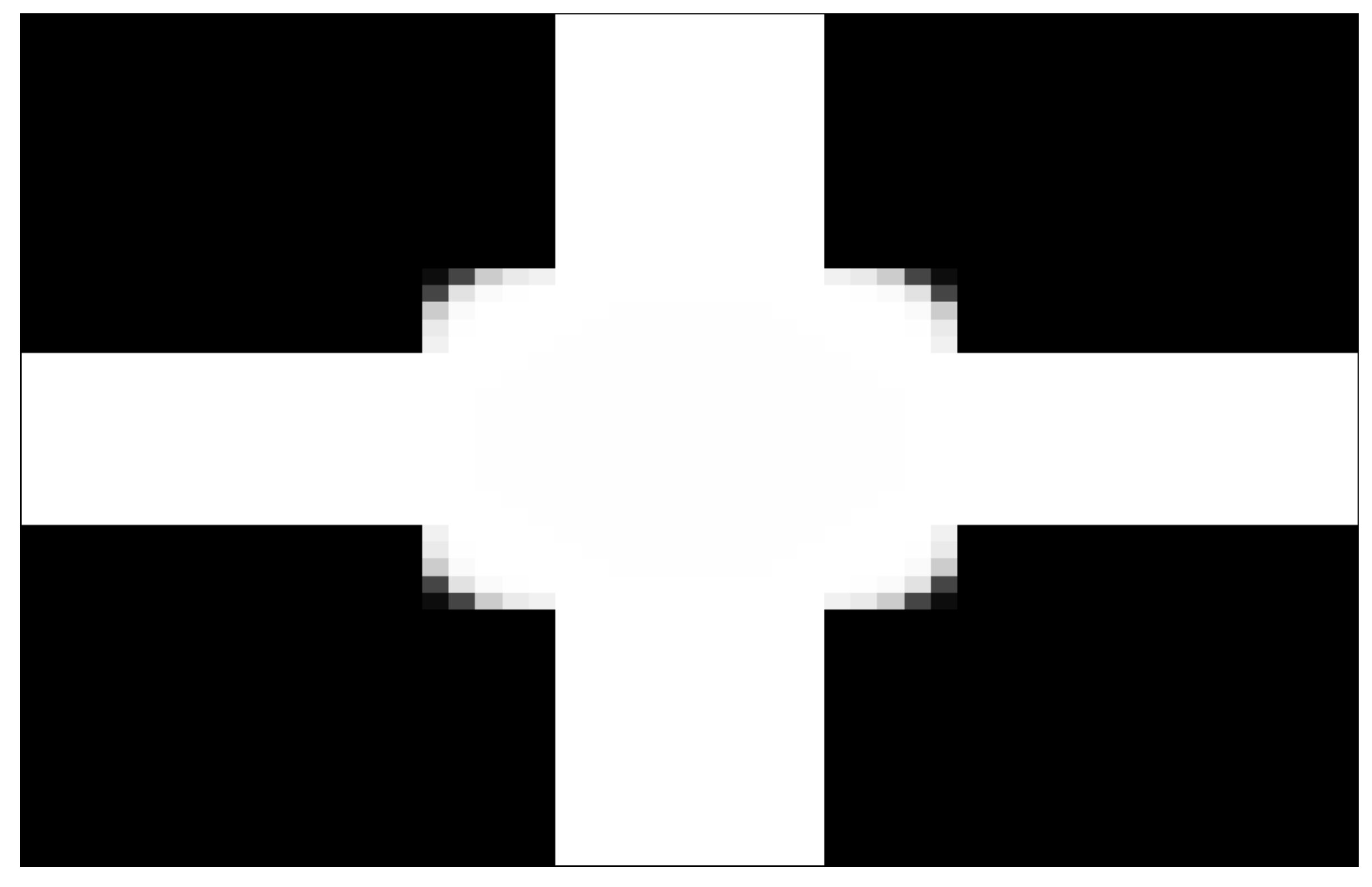}
			\subcaption{Result from image inpainting}
			\label{fig:second_step_final}
		\end{minipage}
		
		\caption{Image inpainting of a crossline, inspired by \cite{bertozzi2006inpainting}.}
		\label{fig:crossline}
		
	\end{figure}

	We employ the crossline example in \cite{bertozzi2006inpainting} to analyse
	the role of the parameters $\epsilon$ and $\lambda_0$ in the finite-volume
	scheme of \autoref{subsec:fvm_mod}. These two parameters crucially determine
	the success of the image inpainting procedure, and consequently appropriate calibrations for
	the parameters must be chosen before running the scheme. The
	original crossline image is depicted in \autoref{fig: initialcrossline}, and
	we add to it a grey damage in the center, as shown in
	\autoref{fig:damagedcrossline}. This image contains $50\times 50$ pixels or
	cells, each with a size of $\Delta x = \Delta y =1$. We apply the
	finite-volume scheme in \autoref{subsec:fvm_mod} to the damaged image in
	\autoref{fig:damagedcrossline}.
	
	We first aim to determine $\lambda_0$ in \eqref{eq:lambda_dis} and we set
	$\epsilon=1$ as an initial guess so that $\epsilon=\Delta x = \Delta y$.
	From \eqref{eq:lambda_dis},
	$\lambda_{i,j}$ is only nonzero for the predefined area of undamaged pixels.
	Indeed the term with  $\lambda_{i,j}$ in the finite-volume scheme in
	\eqref{eq:fv_2} ensures that the undamaged pixels are not modified during the
	image inpainting, but for this $\lambda_0$ has to be sufficiently large to
	counterbalance the fluxes of the scheme. Bearing this in mind we run the
	numerical scheme with a $\Delta t=0.1$  and until the $L^1$ norm between
	successive states is lower than a certain tolerance fixed to be $10^{-4}$.
	Our simulation produces satisfactory results and does not break down for a
	range of $\lambda\in[1,1000]$. \autoref{tab:lambdatime} shows that
	the computational time to reach the required tolerance decreases when increasing the
	value of $\lambda$. In addition, the $L^1$ norm between the final and initial
	state is also lower for greater $\lambda$. It is worth mentioning that
	different choices for $\Delta t$ yield different ranges of valid $\lambda$,
	given that \eqref{eq:fv_2} is a singularly perturbed problem for large
	$\lambda_0$. Hence, greater values  of $\lambda$ are possible if $\Delta
	t$ is refined. In our case, with the choice of $\Delta t=0.1$, our
	finite-volume scheme does not yield any result and breaks down during the
	for values of $\lambda\notin[1,1000]$.

	\begin{table}[h]
		\centering
		\begin{tabular}{lll}
			\toprule
			$\lambda$ & time & $L^1$ norm \\
			\midrule
			1 & 489.8 & 61.2 \\
			10 & 489.7 & 60.8 \\
			100 & 484.8 & 60.8 \\
			1000 & 481.5 & 60.7 \\
			\bottomrule \\
		\end{tabular}
		
		\caption{Comparison for different values of $\lambda$:  computational time before reaching the tolerance and $L^1$ norm between the final and initial state.}
		\label{tab:lambdatime}
	\end{table}
	
	We next consider the tuning of the parameter $\epsilon$ which in turn
	is related to the pixel size $\Delta x$ and
	$\Delta y$. For values of $\epsilon$ larger than the pixel size the outcome
	tends to be diffusive, while for smaller values the edges are sharpened.
	When applying the finite-volume scheme in \eqref{eq:fv_2} with
	$\lambda\in[1,1000]$ we obtain satisfactory results for $\epsilon \in
	[0.5,1.5]$, while for values outside this range the simulation breaks down because of the
	singular nature of~\eqref{eq:fv_2}. As a consequence, for the two-step method in \autoref{subsec:two-step method} we
	first take the value $\epsilon_1=1.5$ for the large-scale
	topological reconnection of shapes, while for the second step we choose the
	value $\epsilon_2=0.5$ to sharpen the edges. The image impainting of the
	damaged image in \autoref{fig:damagedcrossline} resulting from applying this
	choice of parameters is shown in \autoref{fig:second_step_final}. {More details about the choice of $\lambda$, $\epsilon_1$ and $\epsilon_2$ are provided in \autoref{app:sen}, where a sensitivity analysis is carried out depicting the outcome of choosing a not optimal combination of parameters.}
	
	The final outcome after the image inpainting in
	\autoref{fig:damagedcrossline} is not the same as the original image in
	\autoref{fig: initialcrossline}. The reason from this is explained in the
	work of \cite{bertozzi2007analysis}, where multiple steady-state solutions of
	the modified CH equation were shown to exist. As the information under the
	inpainting region has been destroyed, there is no way of knowing that the steady state
	we obtain is less accurate than other viable solutions, in comparison to
	\autoref{fig: initialcrossline}. For further details we refer the reader to
	\cite{bertozzi2007analysis}, where a bifurcation analysis is carried out to
	show that the steady state may vary depending on the choices for
	$\epsilon$ and $\Delta x$, $\Delta y$.

	%
	%
	
	\subsection{Damage introduced in the MNIST dataset}\label{subsec:damage}
\begin{figure}[h!] 
		\centering
		\begin{tabular}[h]{cccccc}
			\toprule
			Damaged rows & Damaged & Inpainted & Damaged pixels & Damaged & Inpainted  \\
			\midrule
			& & & & & \\[-0.5cm]
			8&\includegraphics[width=0.13\linewidth, height=2cm, valign=m]{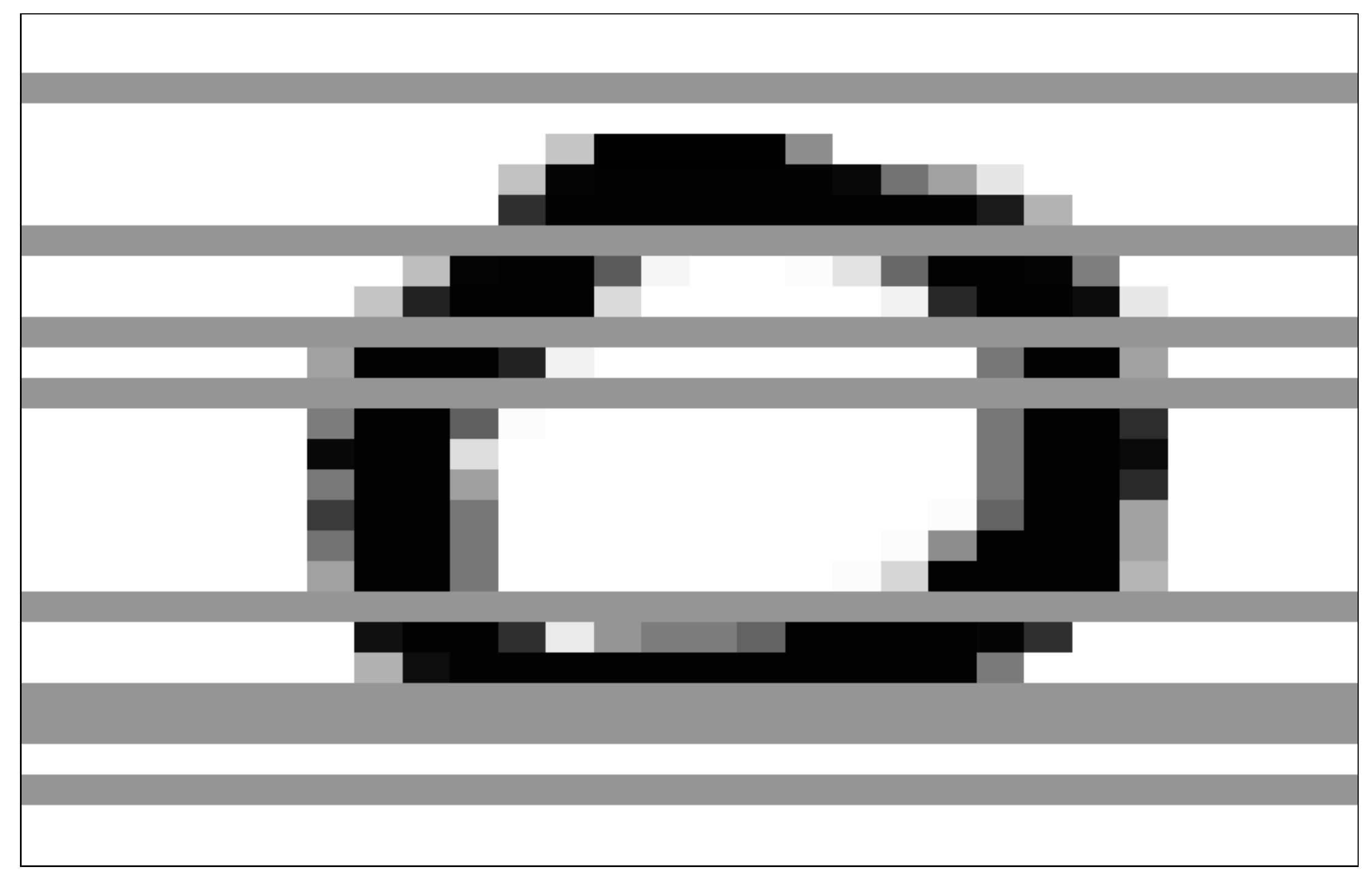}&
			\includegraphics[width=0.13\linewidth, height=2cm, valign=m]{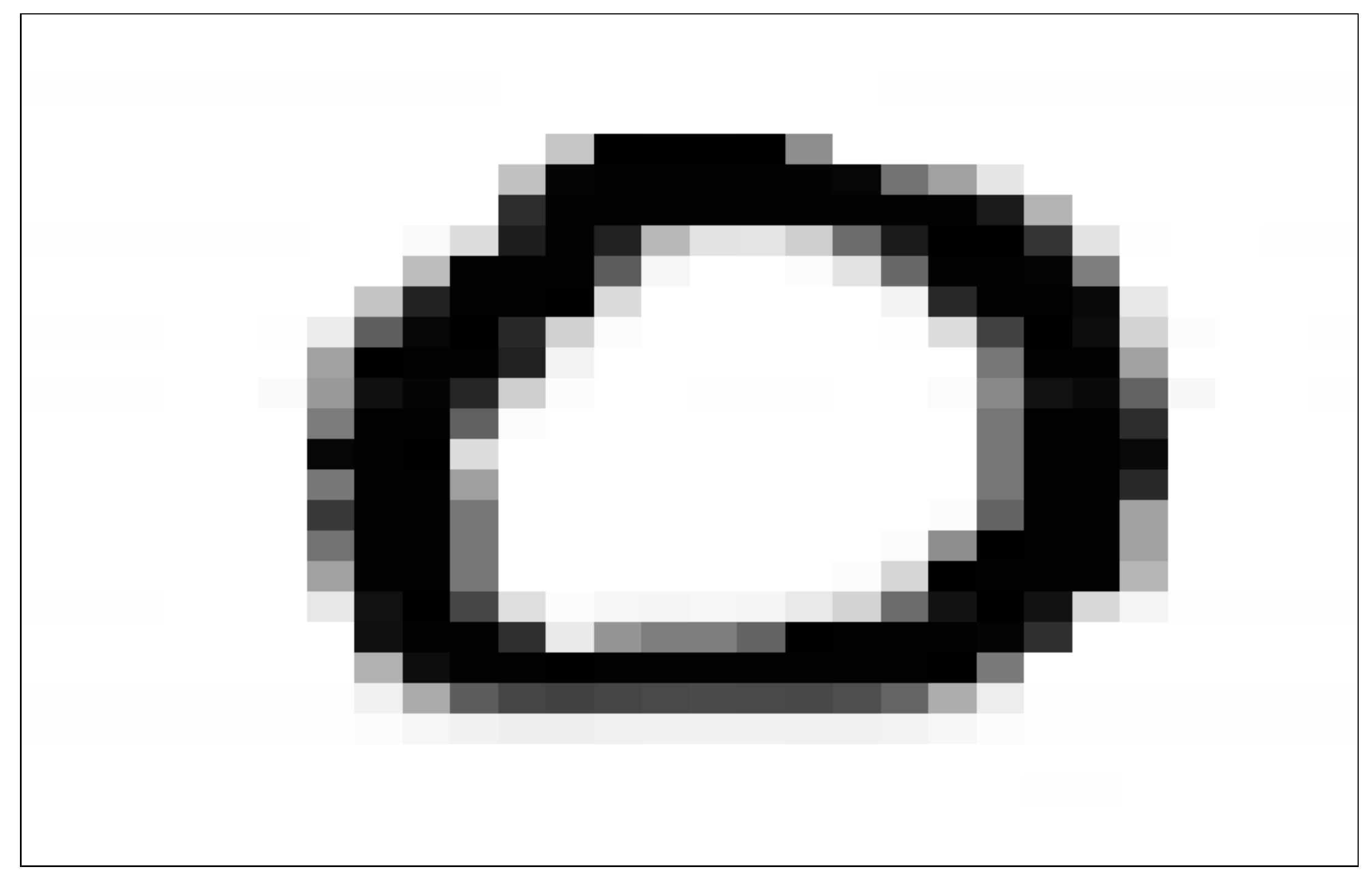}&
			$30\%$&\includegraphics[width=0.13\linewidth, height=2cm, valign=m]{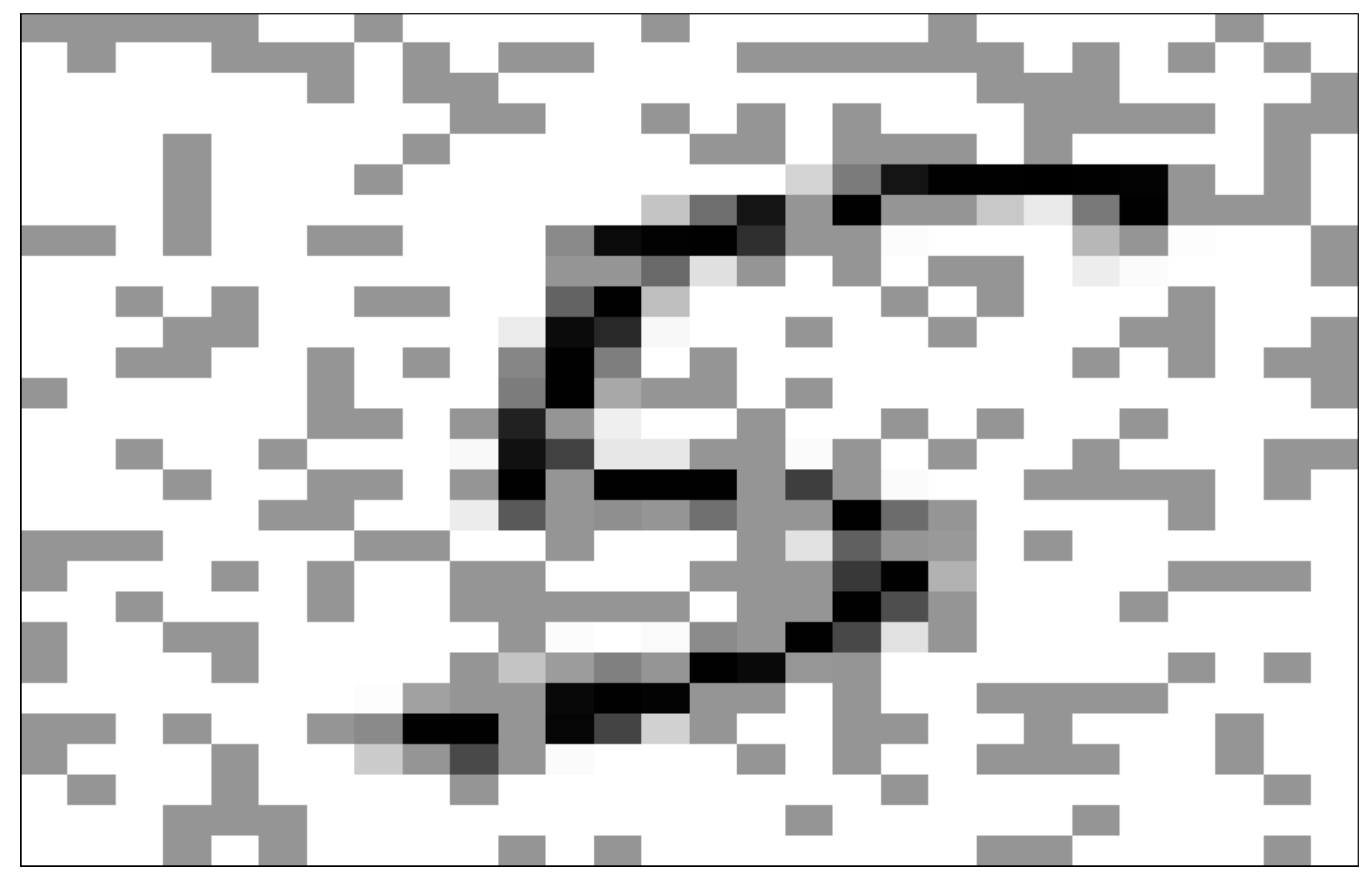}&
			\includegraphics[width=0.13\linewidth, height=2cm, valign=m]{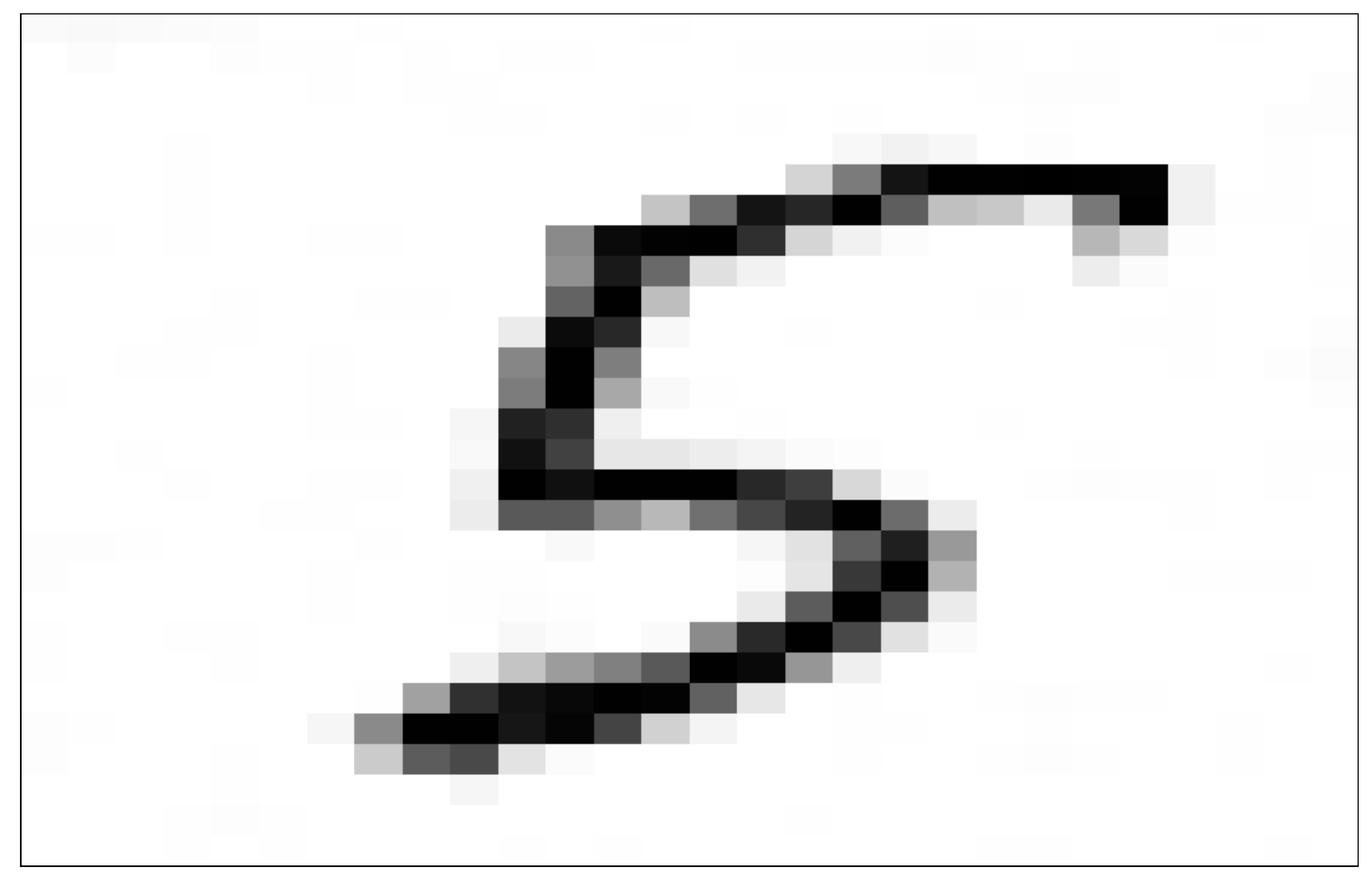}\\[1.5cm]
			10&\includegraphics[width=0.13\linewidth, height=2cm, valign=m]{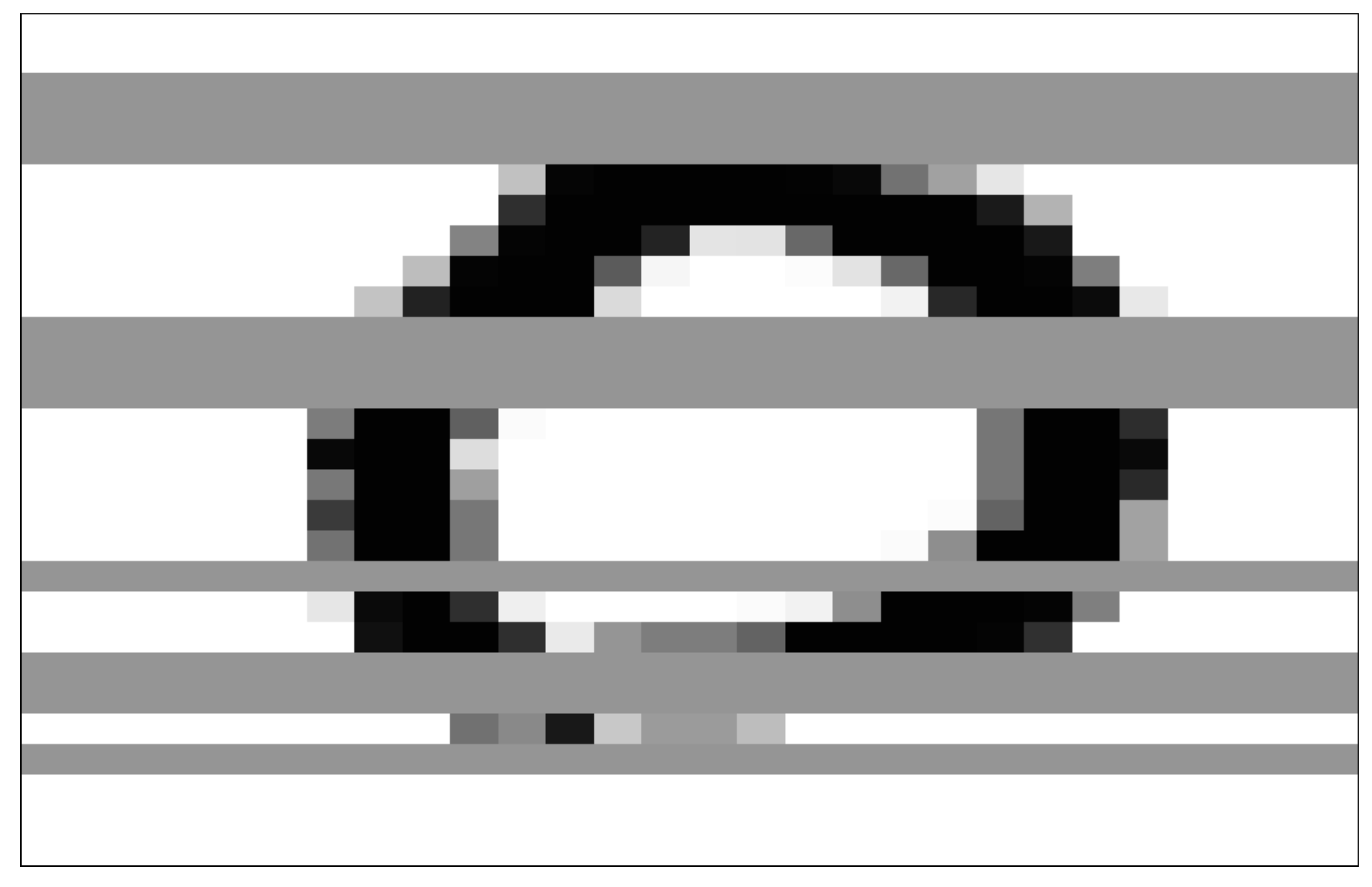}&
			\includegraphics[width=0.13\linewidth, height=2cm, valign=m]{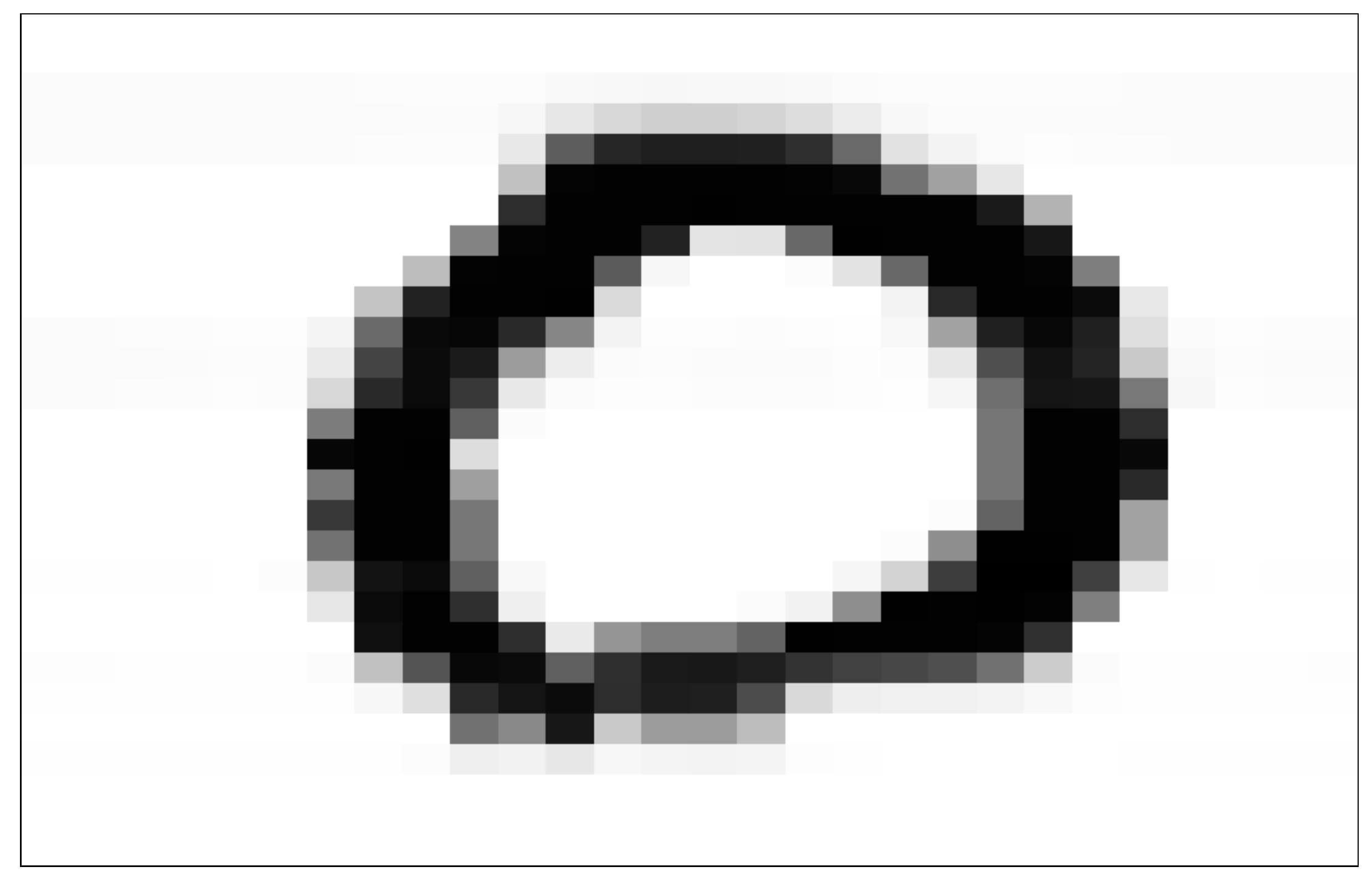}&
			$40\%$&\includegraphics[width=0.13\linewidth, height=2cm, valign=m]{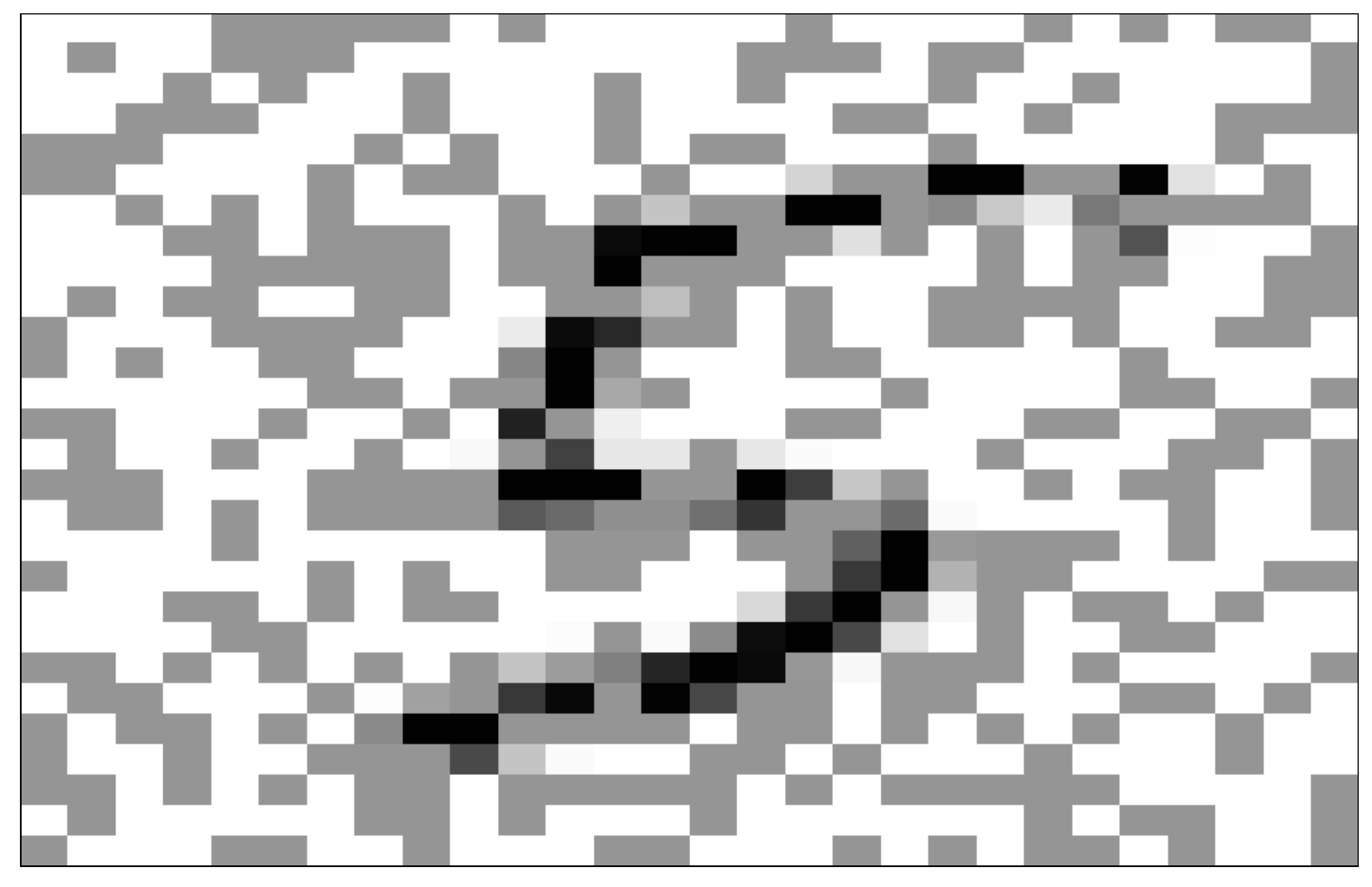}&
			\includegraphics[width=0.13\linewidth, height=2cm, valign=m]{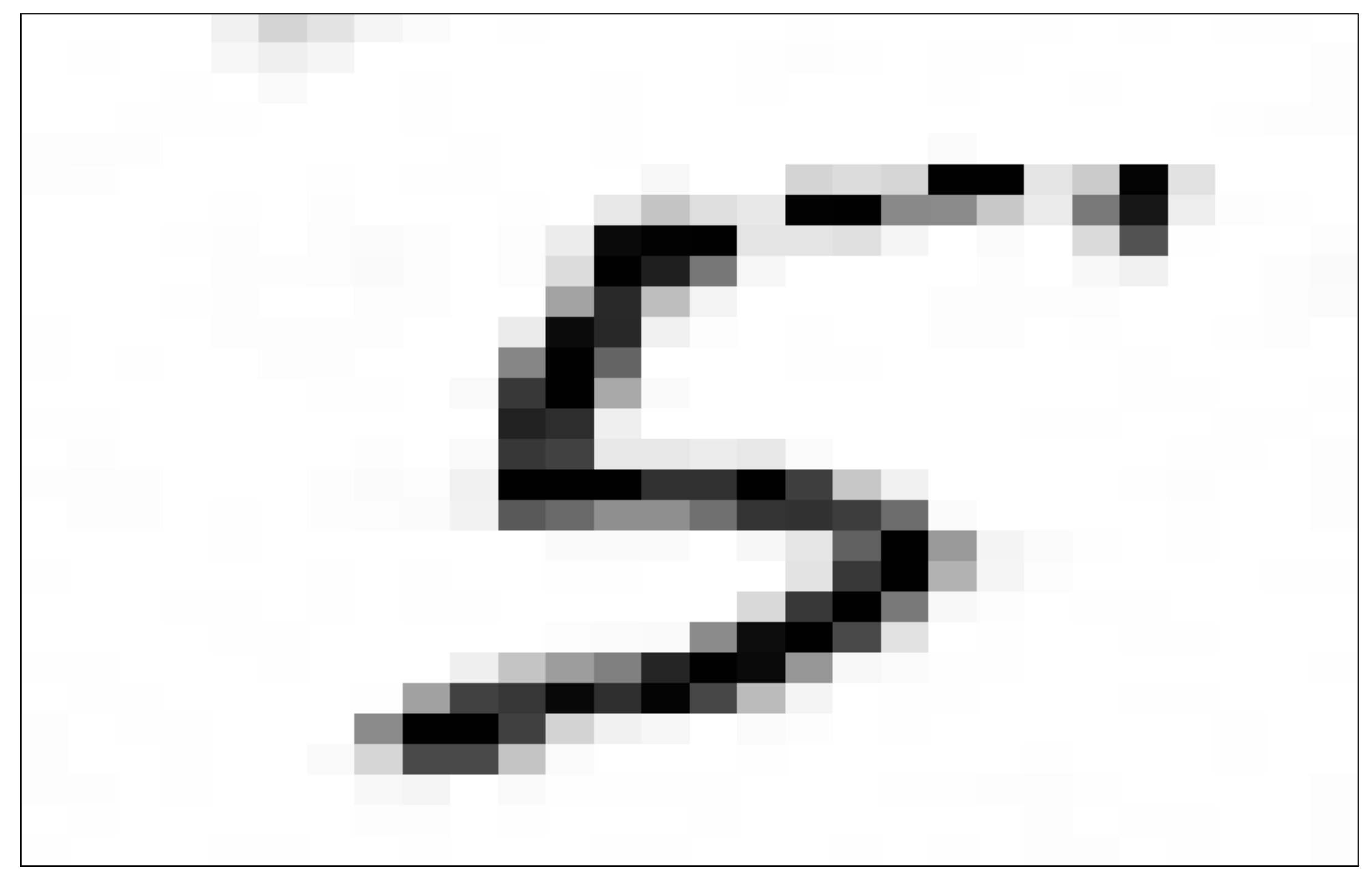}\\[1.5cm]
			12&\includegraphics[width=0.13\linewidth, height=2cm, valign=m]{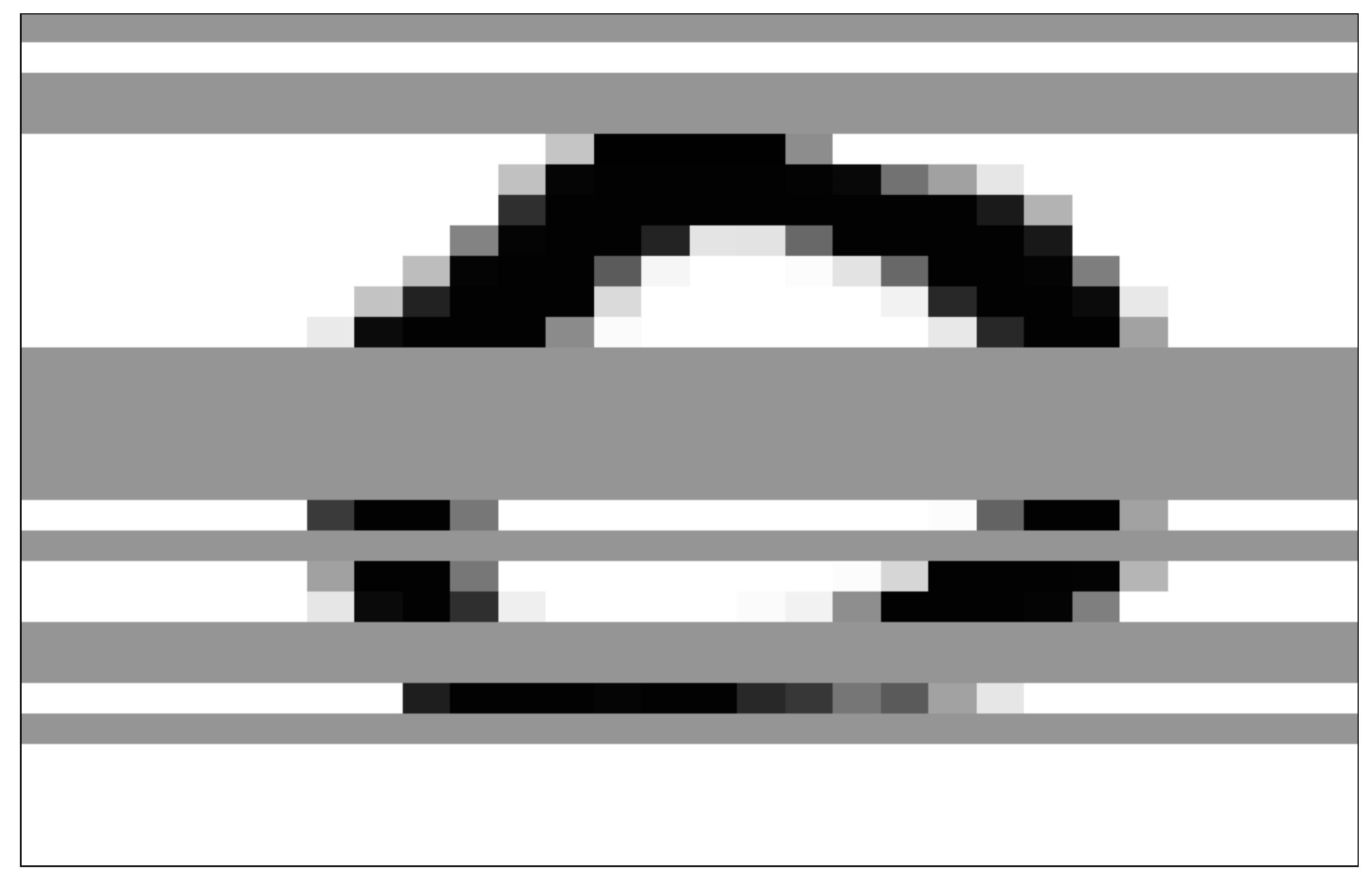}&
			\includegraphics[width=0.13\linewidth, height=2cm, valign=m]{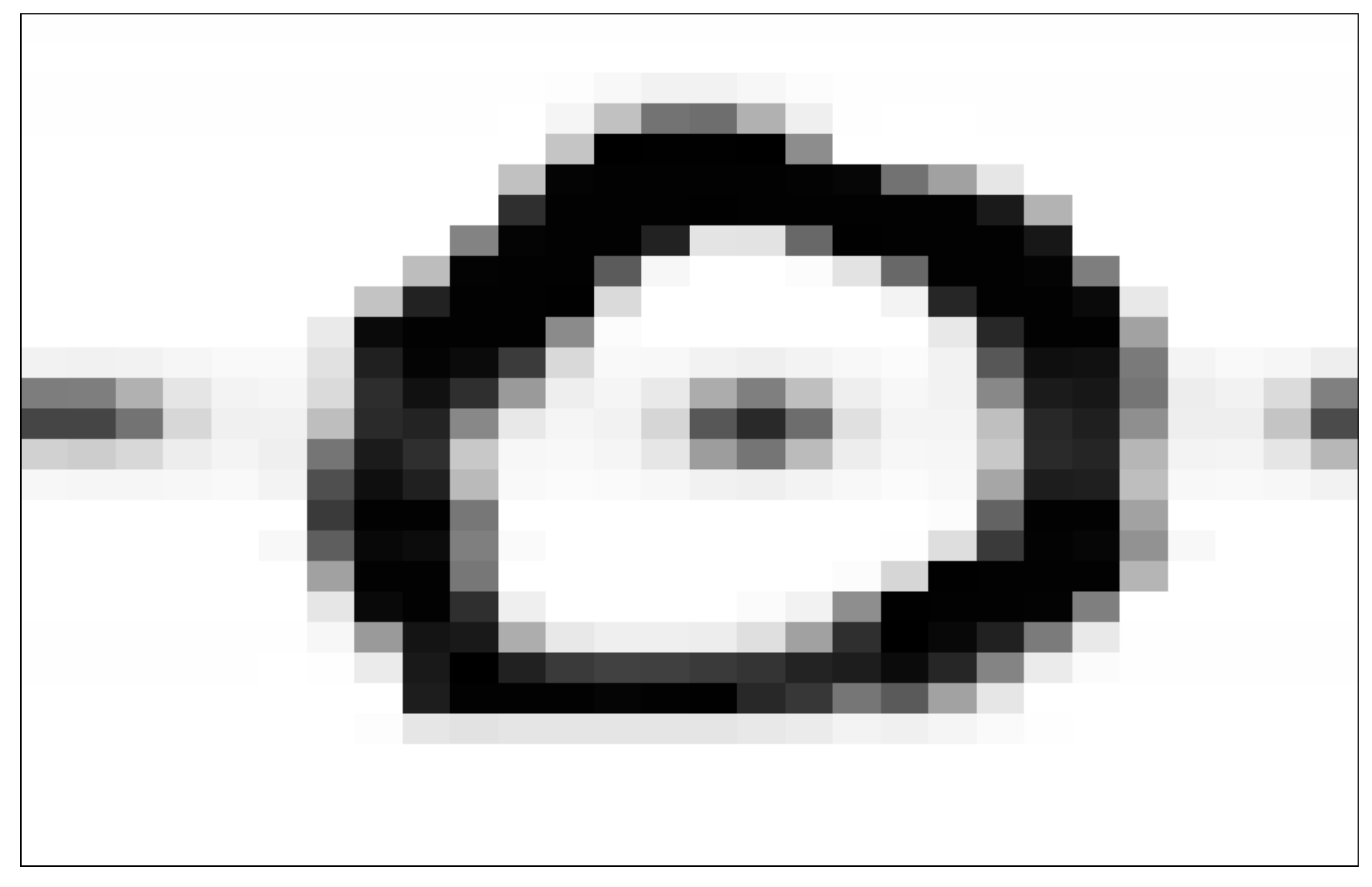}&
			$50\%$&\includegraphics[width=0.13\linewidth, height=2cm, valign=m]{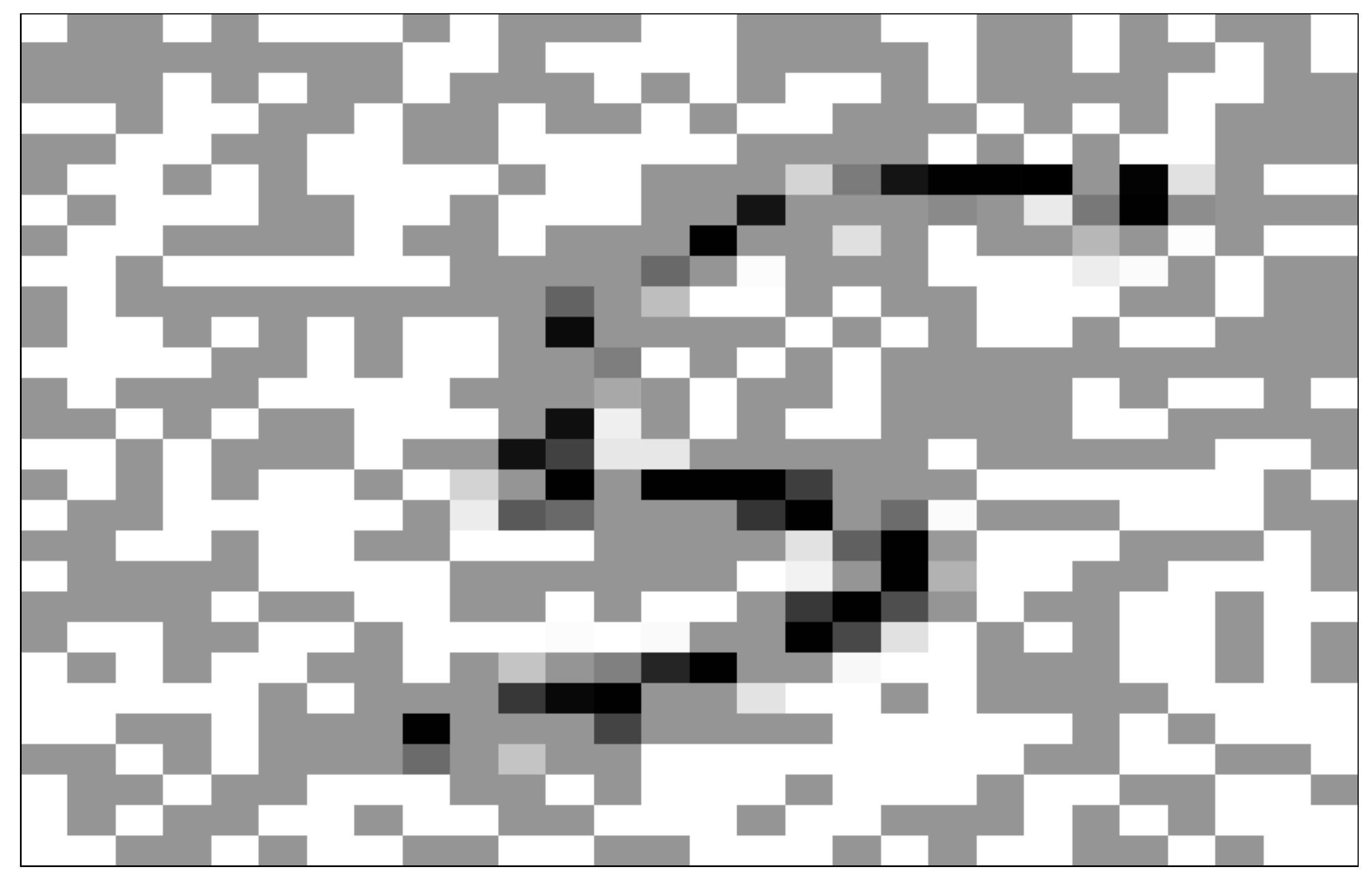}&
			\includegraphics[width=0.13\linewidth, height=2cm, valign=m]{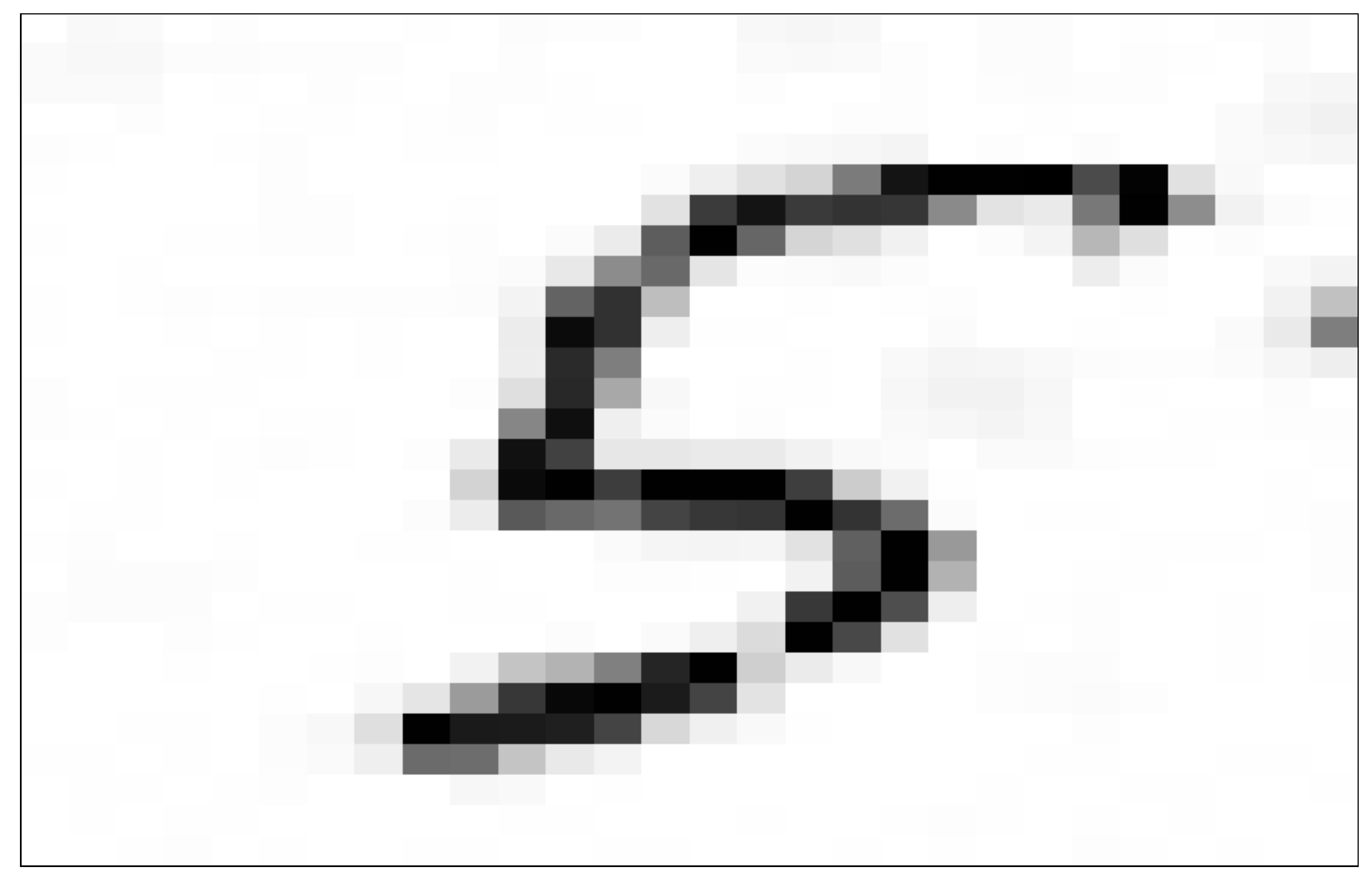}\\[1.5cm]
			14&\includegraphics[width=0.13\linewidth, height=2cm, valign=m]{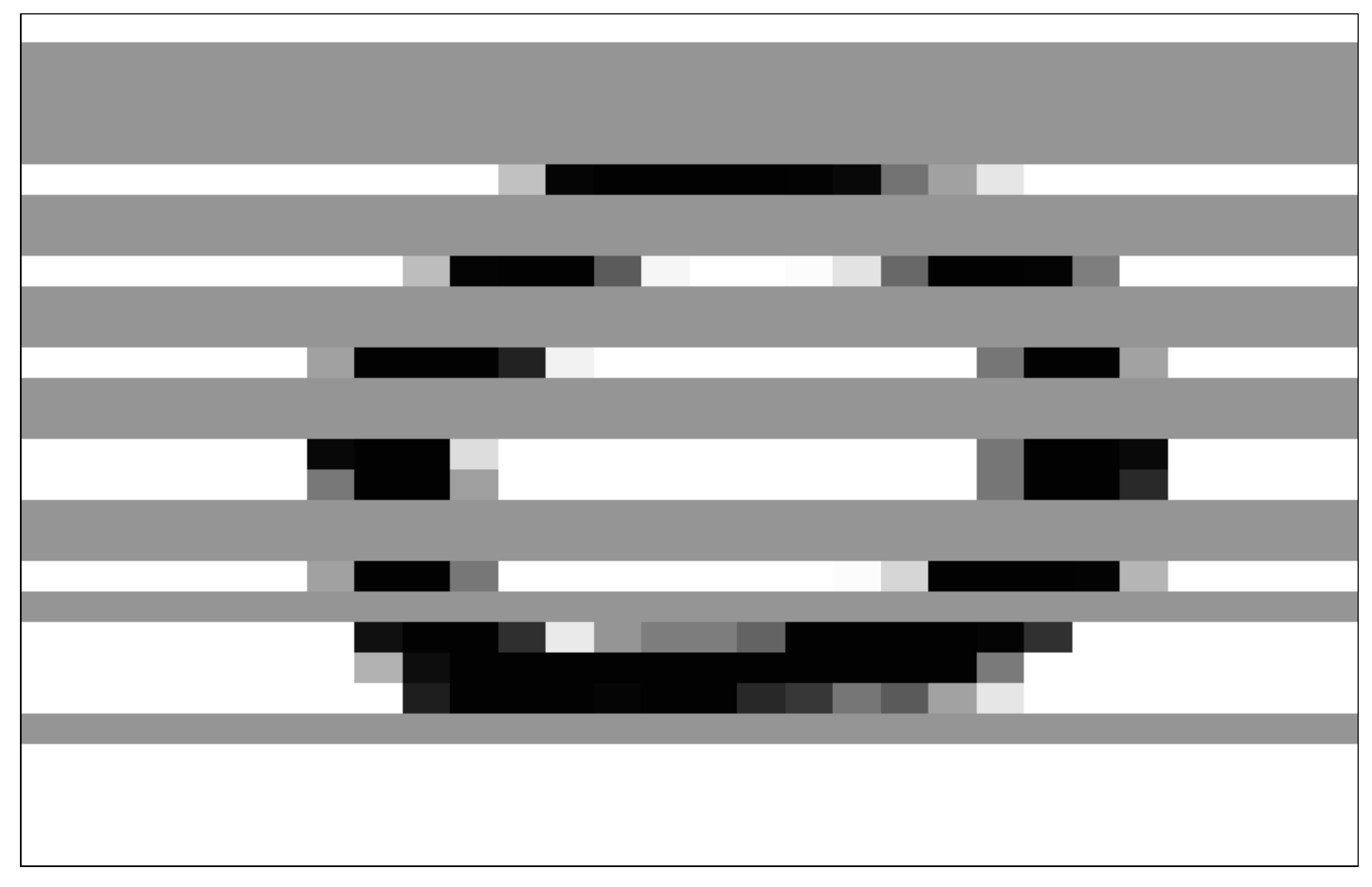}&
			\includegraphics[width=0.13\linewidth, height=2cm, valign=m]{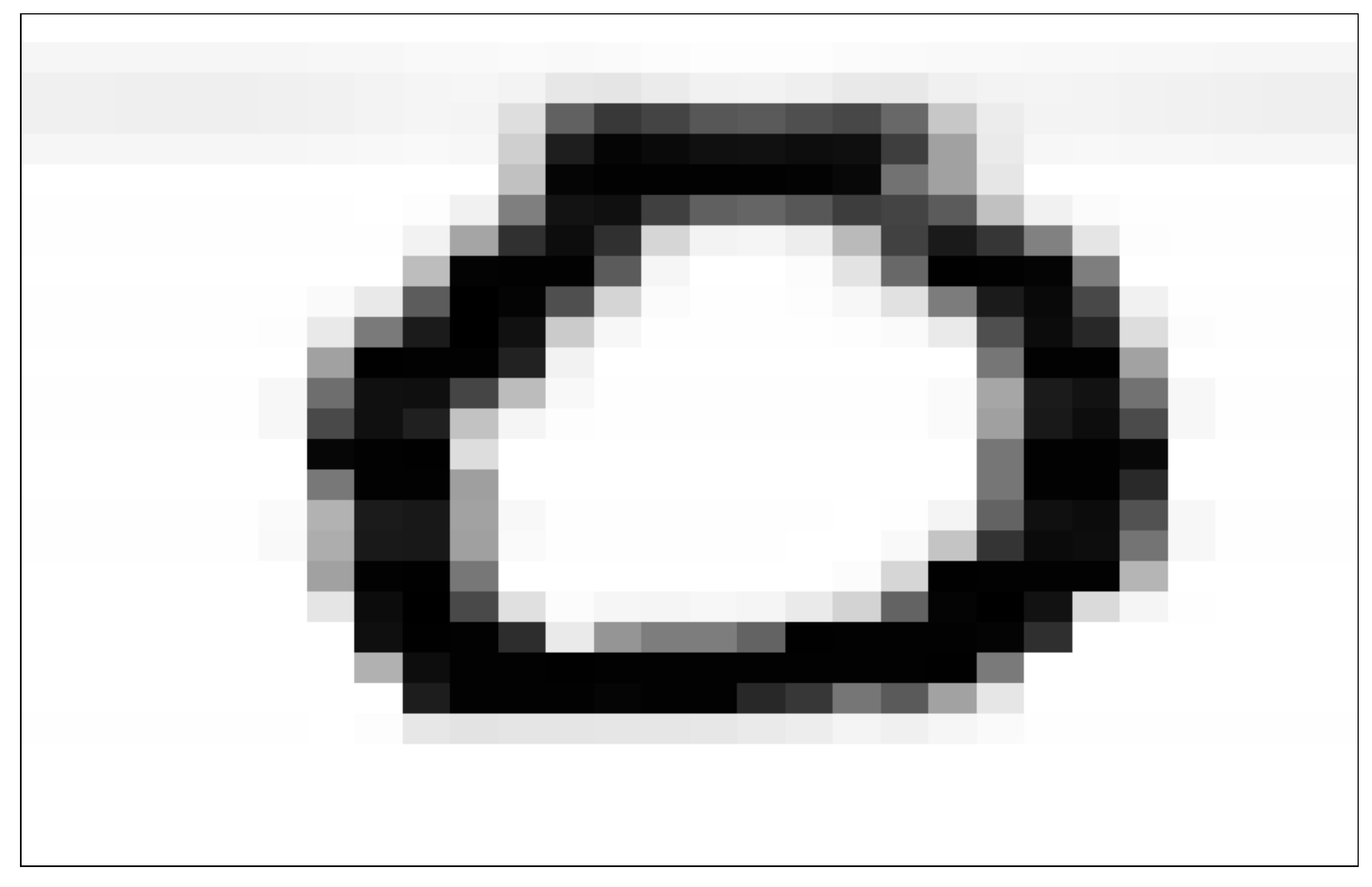}&
			$60\%$&\includegraphics[width=0.13\linewidth, height=2cm, valign=m]{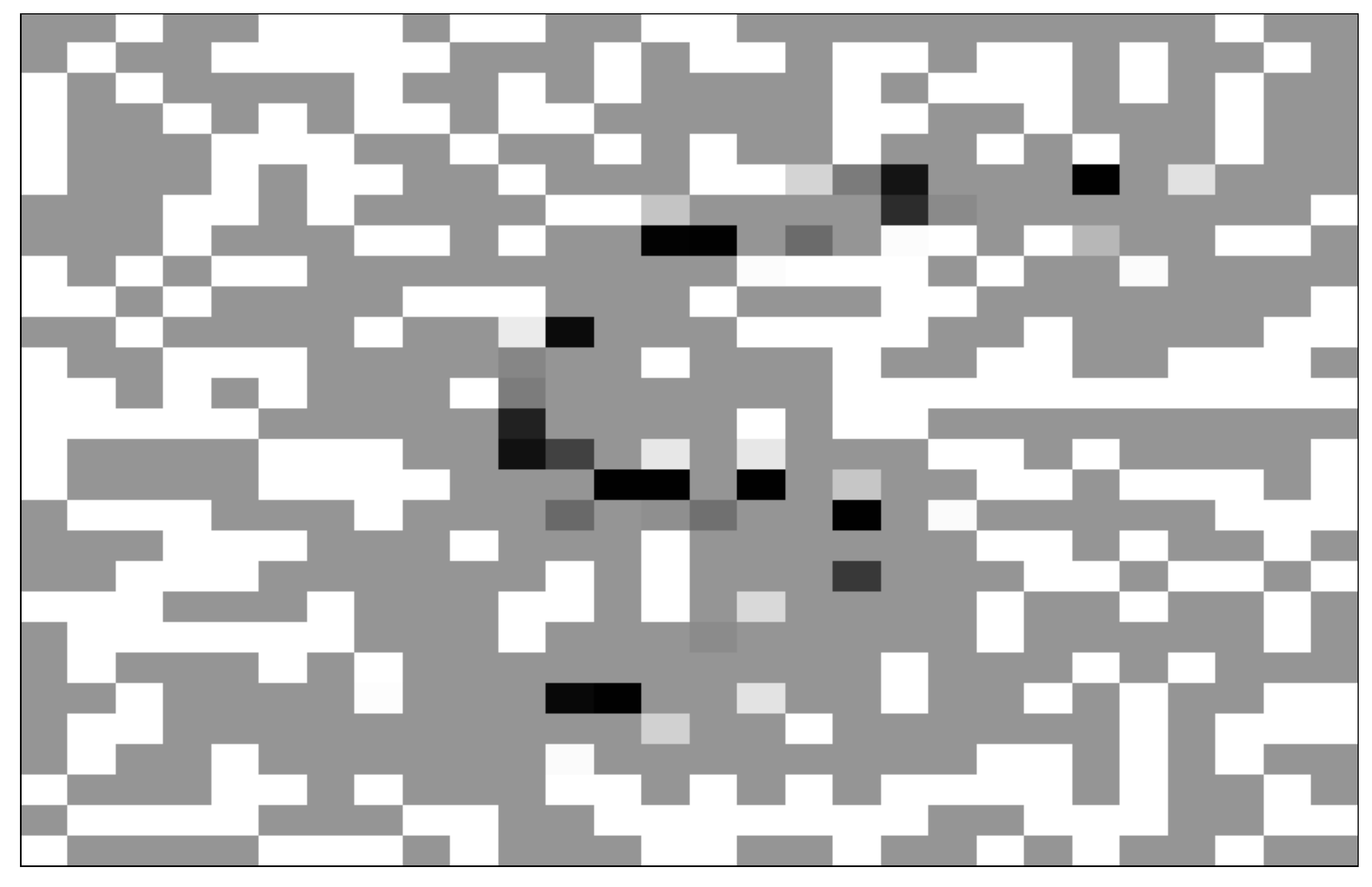}&
			\includegraphics[width=0.13\linewidth, height=2cm, valign=m]{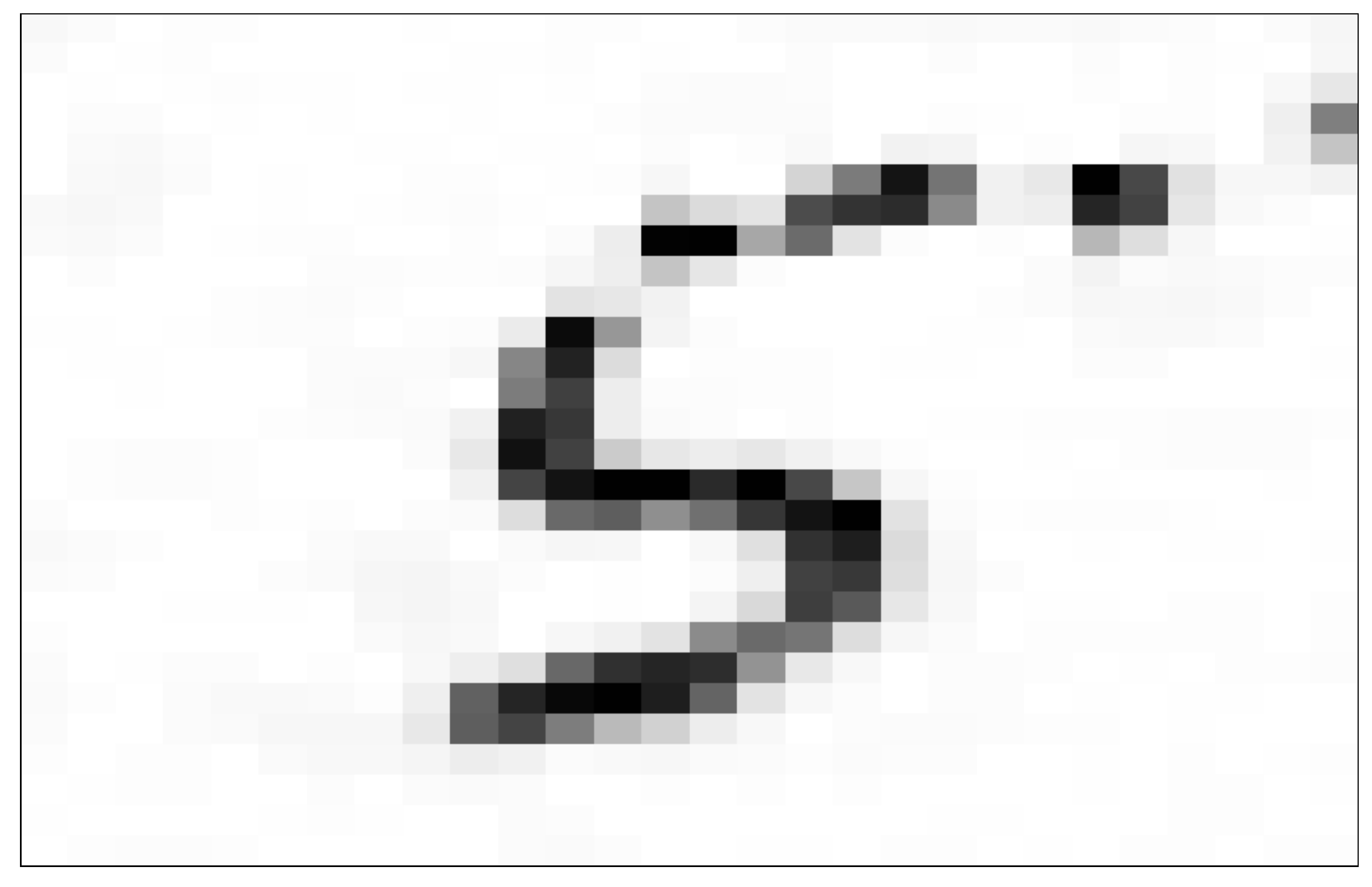}\\[1.5cm]
			16&\includegraphics[width=0.13\linewidth, height=2cm, valign=m]{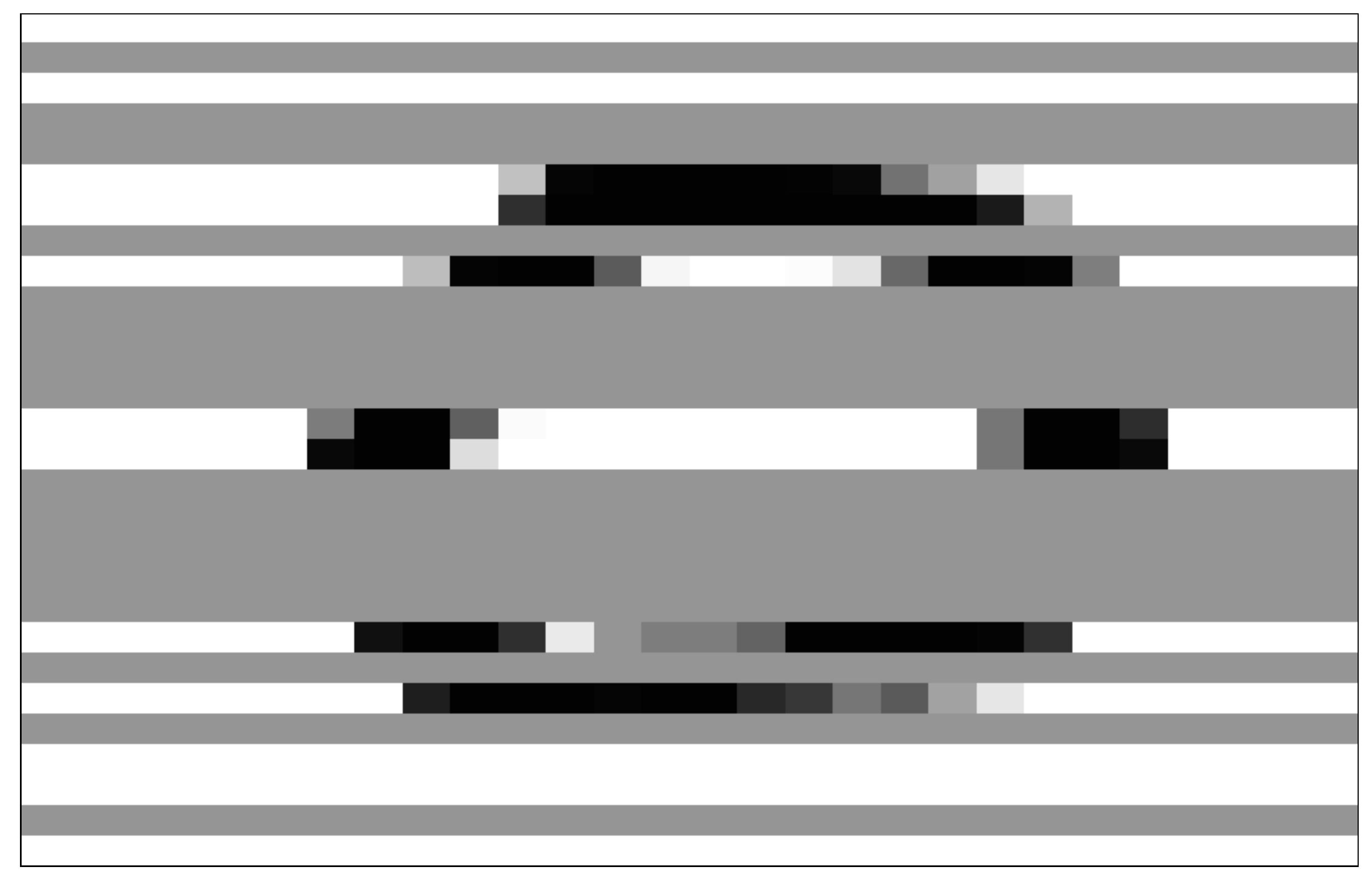}&
			\includegraphics[width=0.13\linewidth, height=2cm, valign=m]{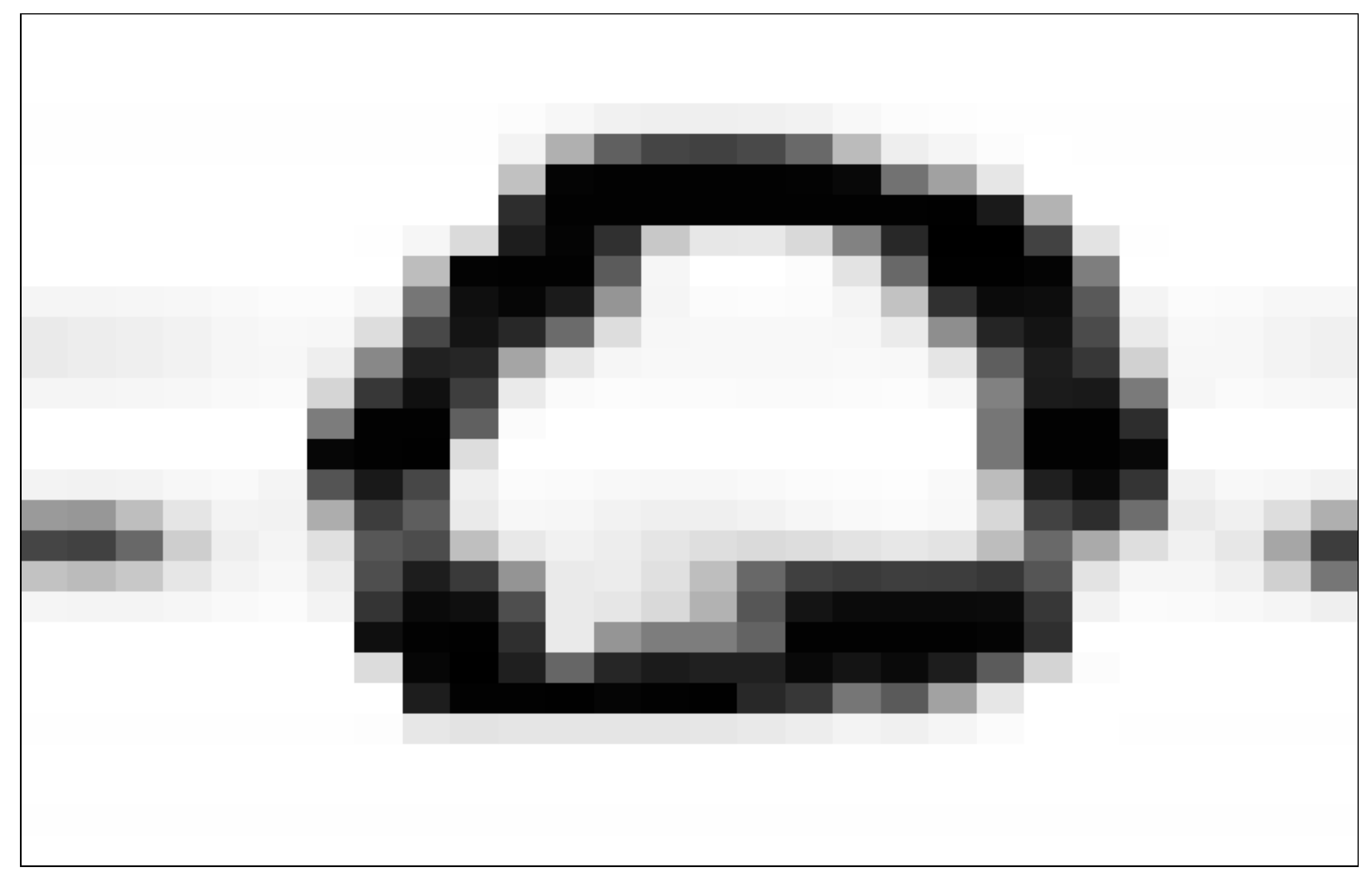}&
			$70\%$&\includegraphics[width=0.13\linewidth, height=2cm, valign=m]{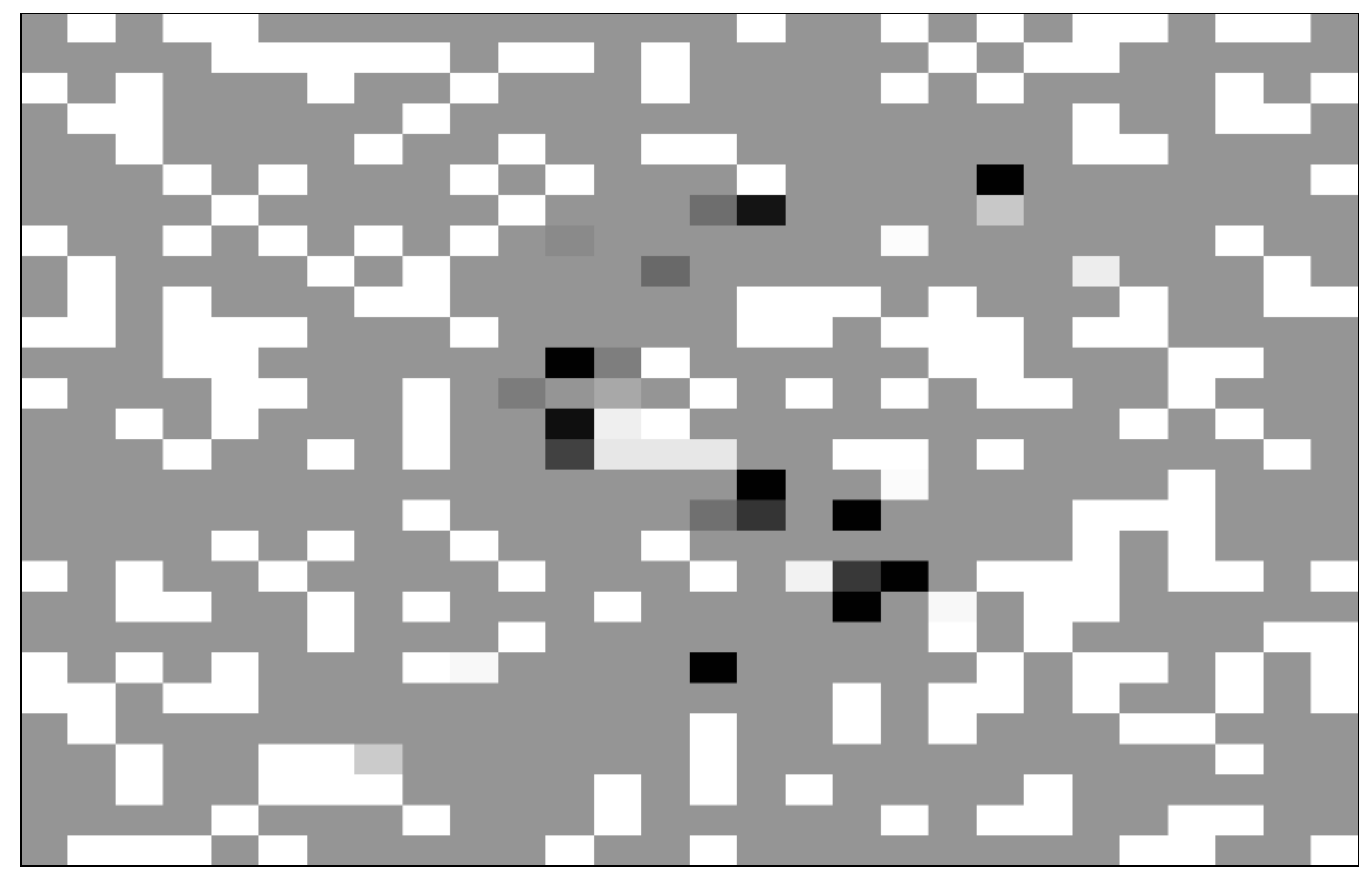}&
			\includegraphics[width=0.13\linewidth, height=2cm, valign=m]{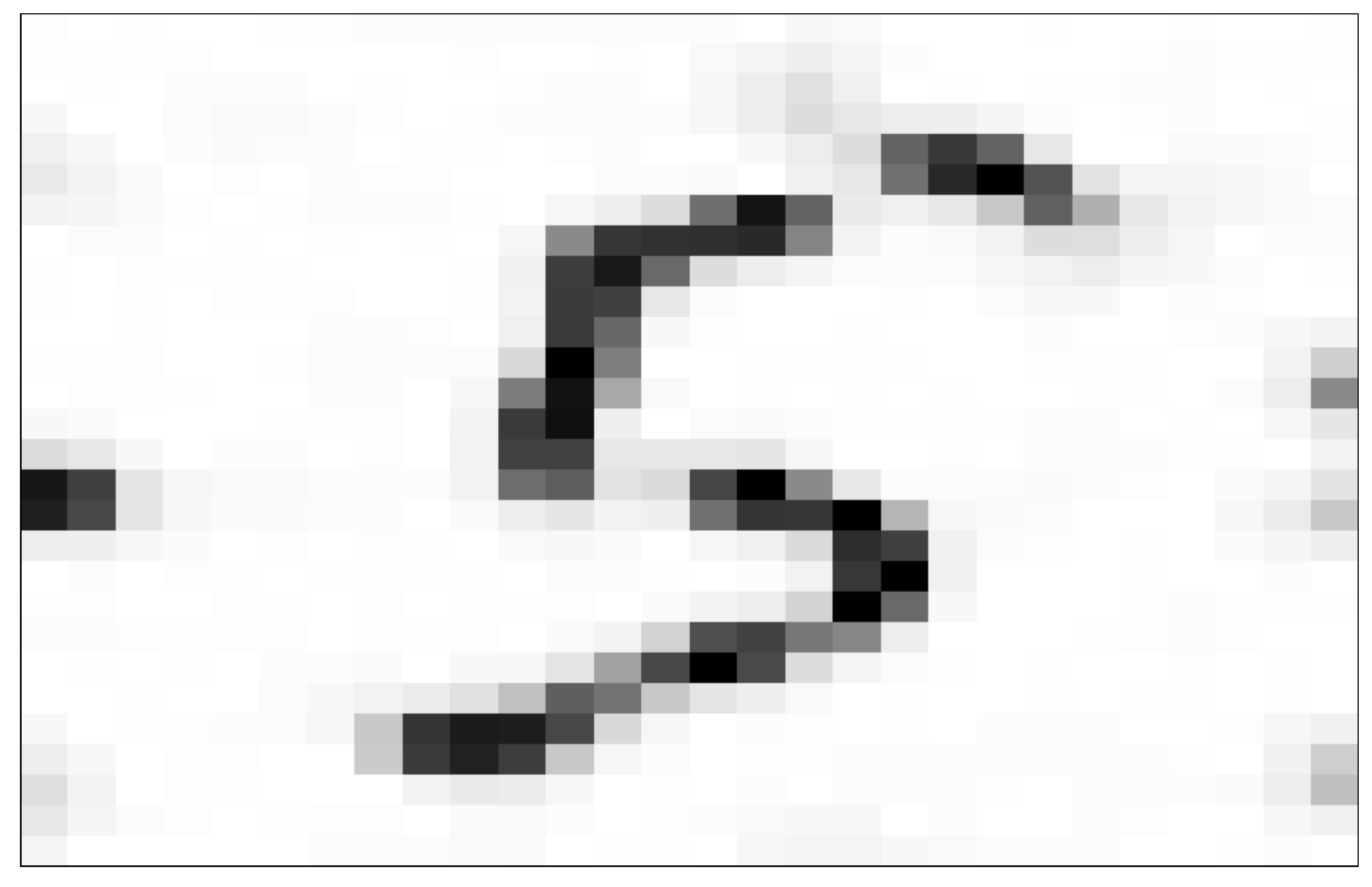}\\[1.5cm]
			18&\includegraphics[width=0.13\linewidth, height=2cm, valign=m]{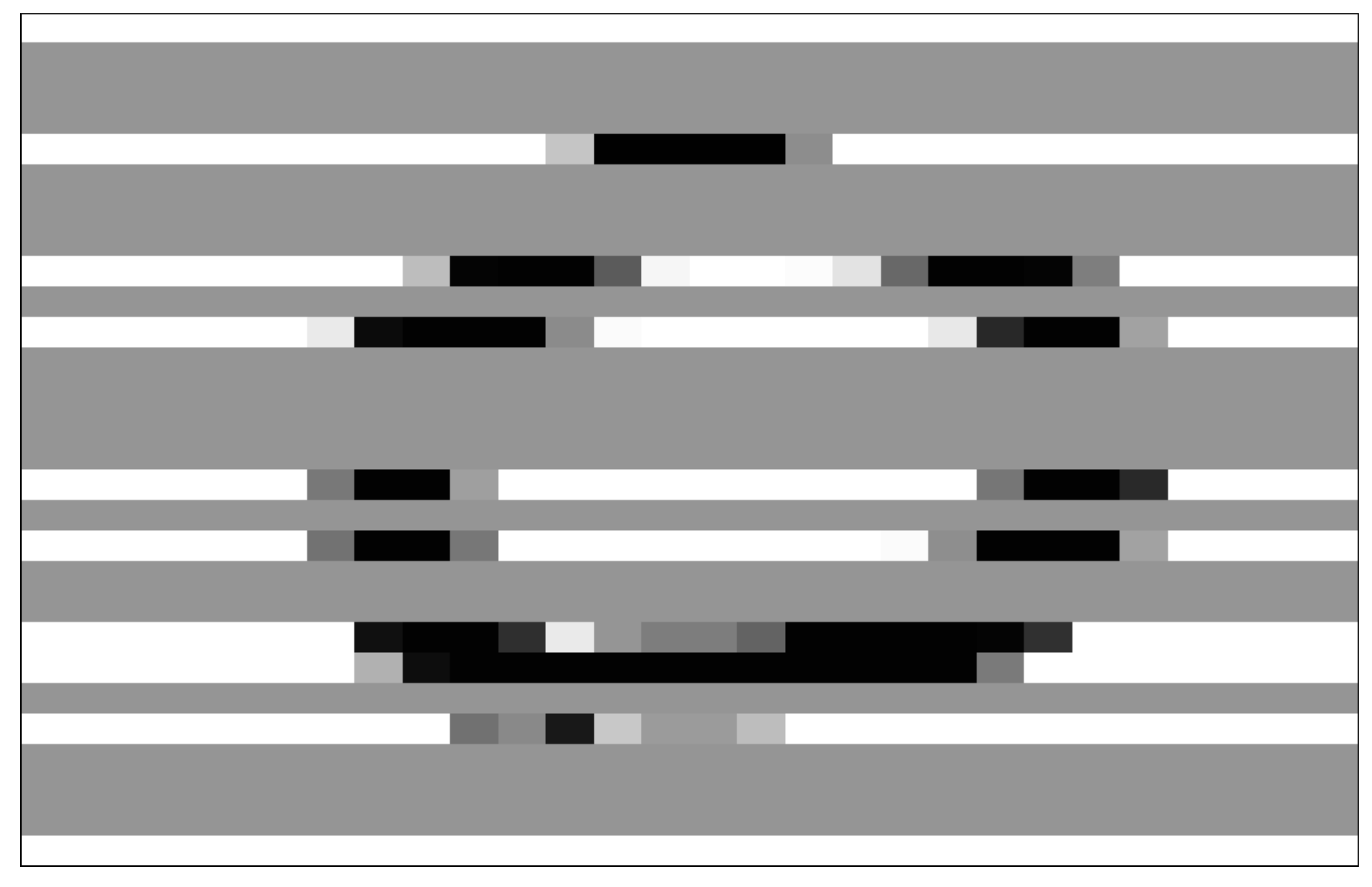}&
			\includegraphics[width=0.13\linewidth, height=2cm, valign=m]{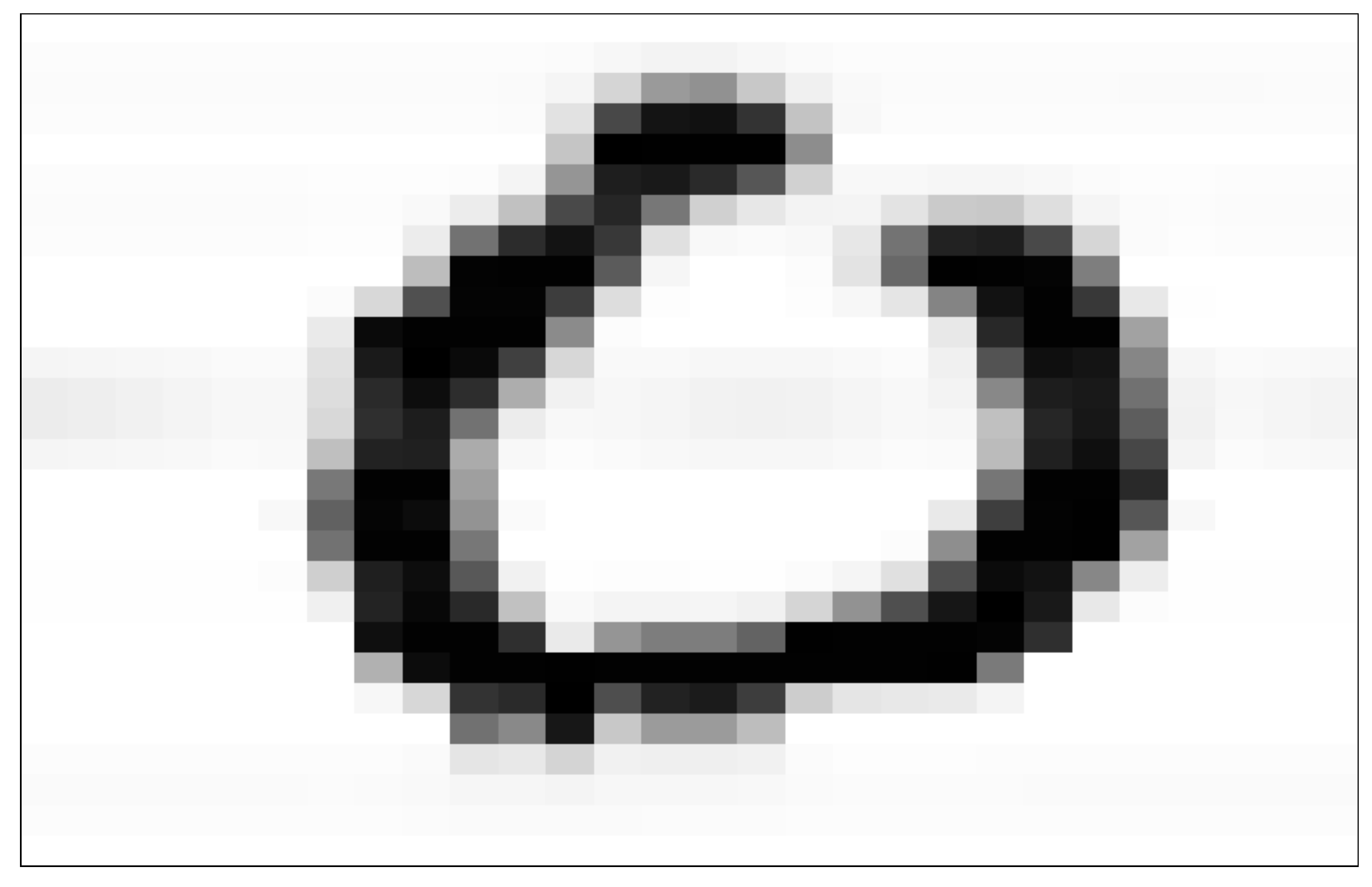}&
			$80\%$&\includegraphics[width=0.13\linewidth, height=2cm, valign=m]{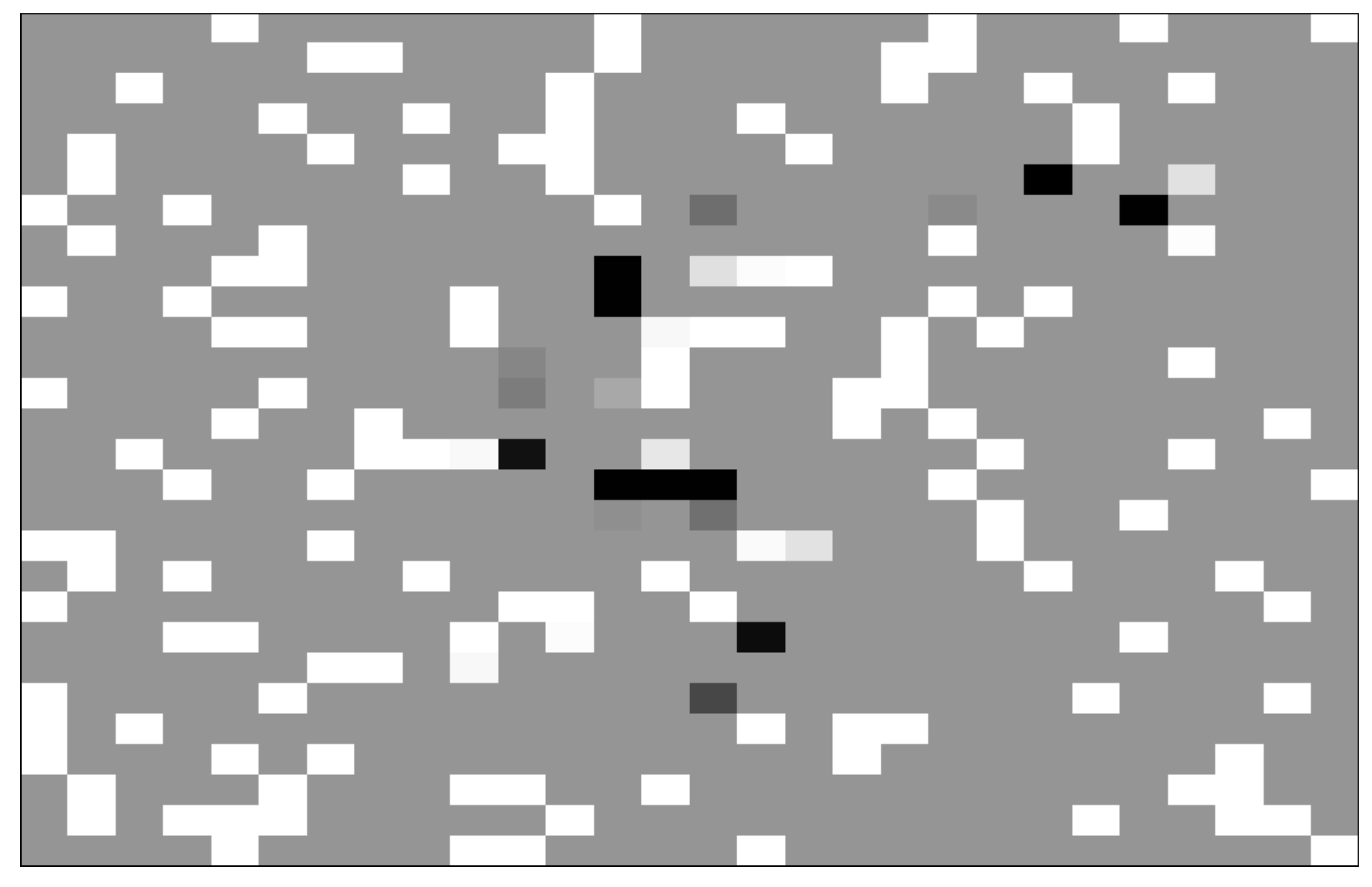}&
			\includegraphics[width=0.13\linewidth, height=2cm, valign=m]{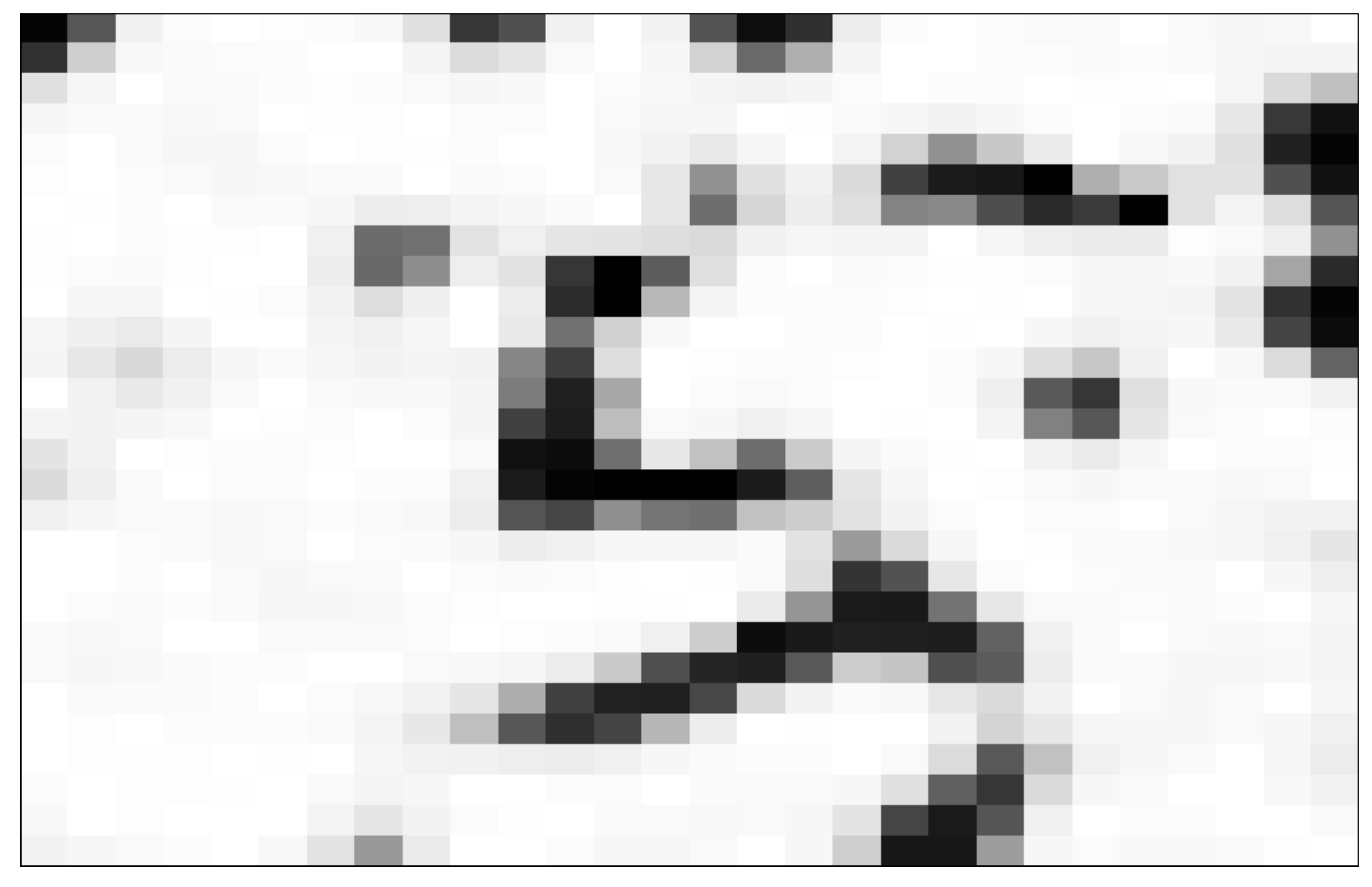}\\[1cm]
			
			\bottomrule
		\end{tabular}
		
		\caption{Examples of image inpainting for random damage in whole horizontal rows and in individual pixels. The column entitled
			``Damaged rows" marks the number of rows randomly selected for damage in the $28\times28$ images. The column entitled ``Damaged pixels"
			marks the percentage of ramdonly damaged pixels over the whole image.
			For higher levels of damage intensity the inpaintings lose more information. }
		\label{fig:random}
	\end{figure}

	Here we discuss the types of damage inserted into the MNIST test set, with
	the objective of subsequently applying the CH filter developed in
	\autoref{subsec:fvm_mod} for image inpainting. The varied damage employed
	aims to represent a mock case of damage that may be encountered in an image
	in need for restoration. As a result, we decide to employ two kinds of damage
	with different intensities: customized damage affecting particular regions of
	the image, and random damage selecting arbitrary pixels or horizontal lines
	in the image. The details of both are:
	
	\begin{enumerate}[label=\alph*)]
		\item Customized damage: this type of damage is applied in four
		different fashions, as shown in \autoref{fig:damageexample}. The basic idea
		is to turn vertical or horizontal lines of pixels into a uniform grey
		intensity between black and white color. In \autoref{fig:damageexample} we
		show the outcome of applying the CH filter to the damaged images. It can be
		seen that our model is able to recover the images from the different types
		of damage, albeit with varying degrees of success. For instance, the damage
		introduced in \autoref{fig:dexample3} is a thick horizontal line which
		implied a considerable loss of information from the original image,
		compared to the other types of damage. As a result, the inpainted image
		filter for this type of damage is not as effective as the other ones, as it
		can be seen from the inpaintings in \autoref{fig:damageexample}.

		\begin{figure}[h]
			\centering
			
			\begin{minipage}{0.24\textwidth}
				\includegraphics[width=\linewidth, height=3.5cm]{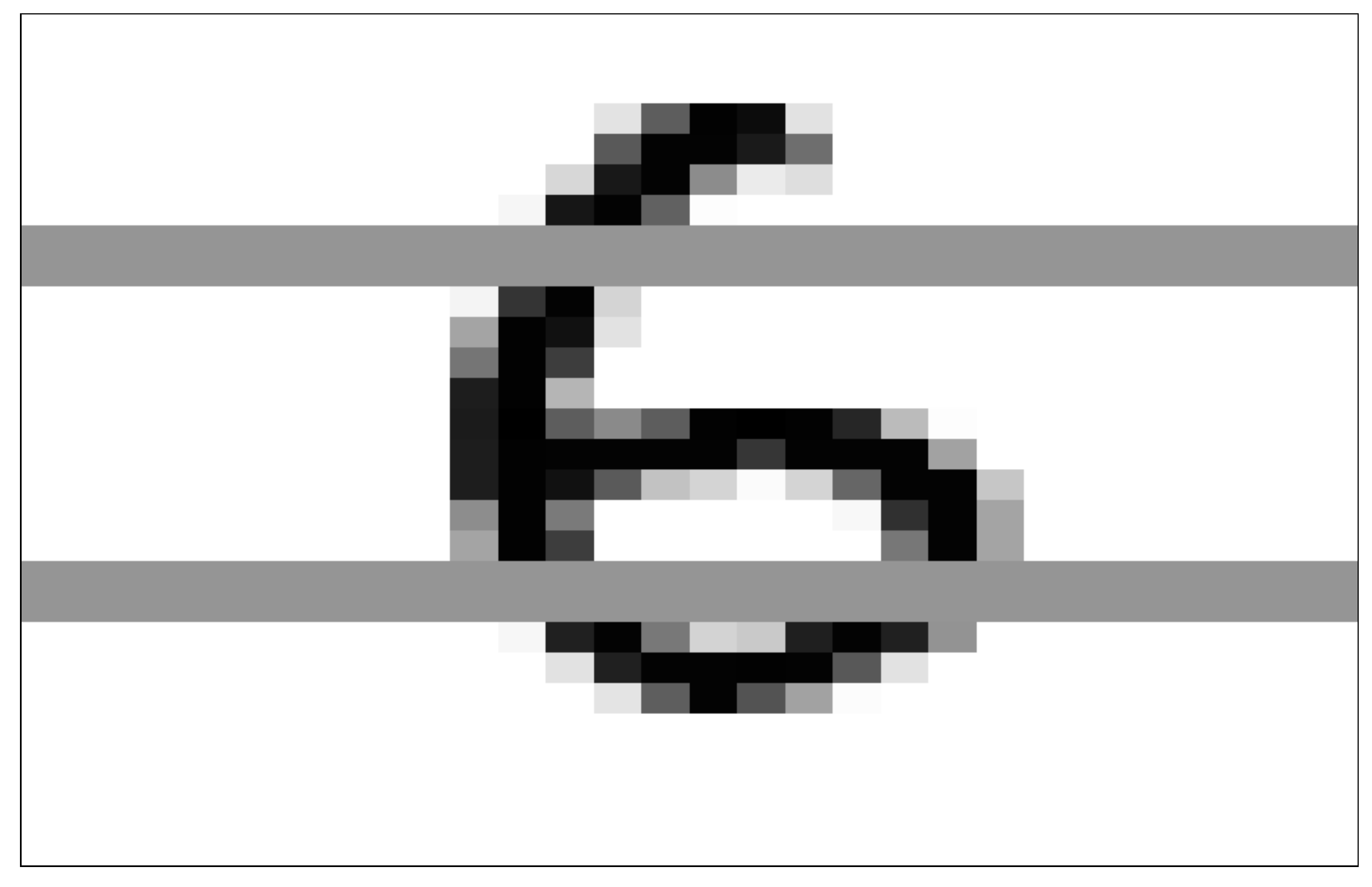}
				\subcaption{}
				\label{fig:dexample1}
			\end{minipage}
			\hfill
			\begin{minipage}{0.24\textwidth}
				\includegraphics[width=\linewidth, height=3.5cm]{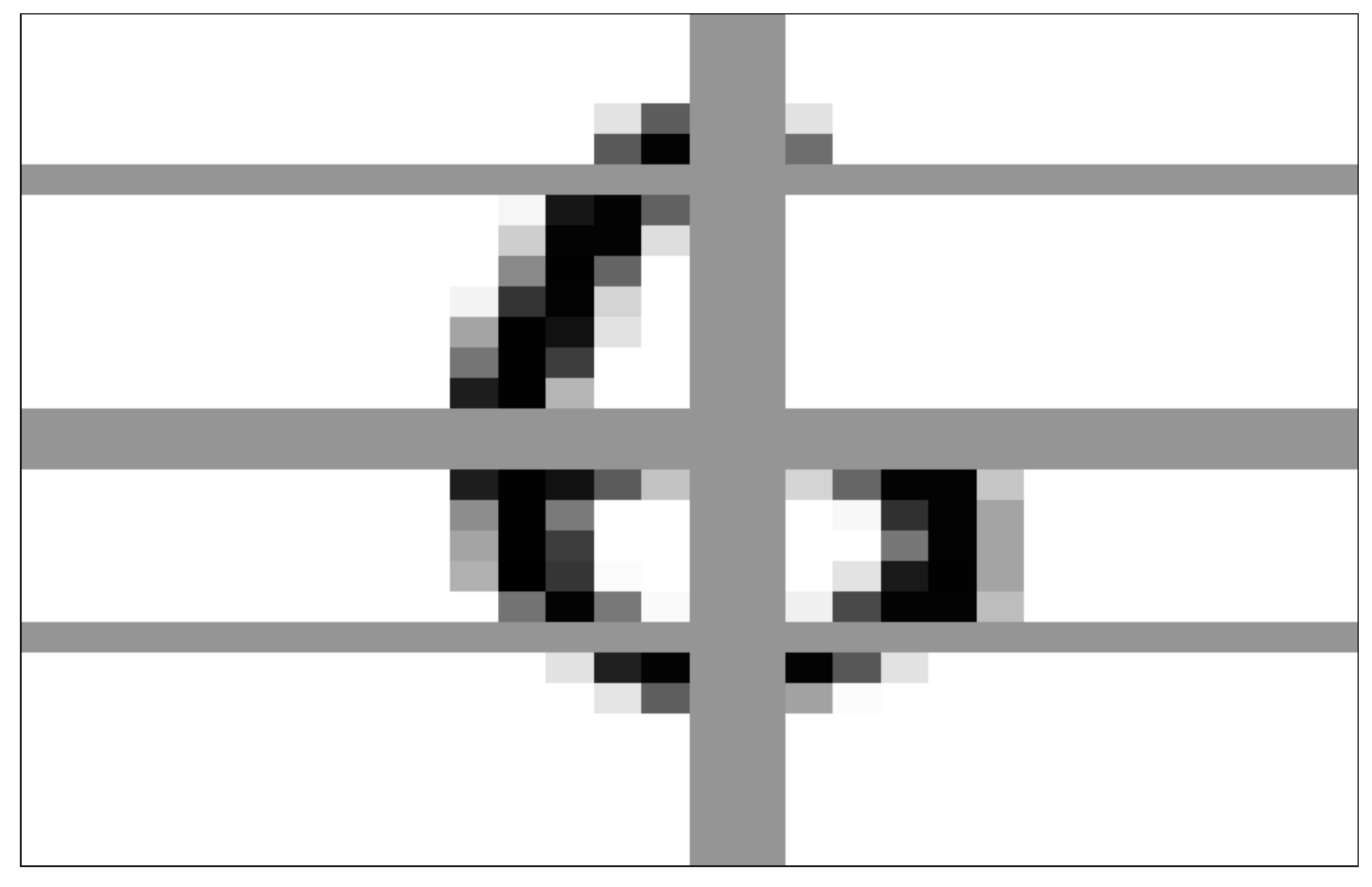}
				\subcaption{}
				\label{fig:dexample2}
			\end{minipage}
			\hfill
			\begin{minipage}{0.24\textwidth}
				\includegraphics[width=\linewidth, height=3.5cm]{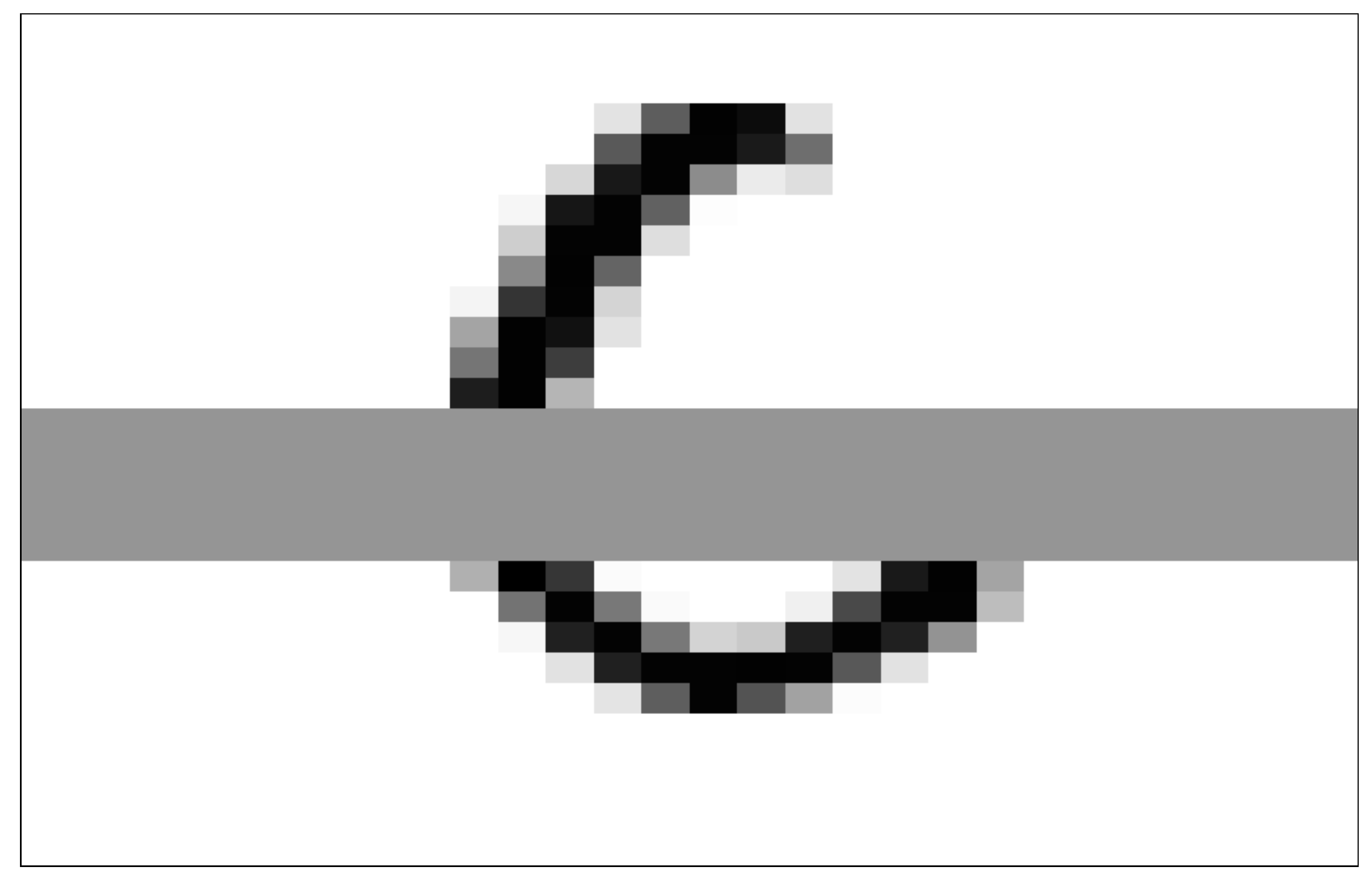}
				\subcaption{}
				\label{fig:dexample3}
			\end{minipage}
			\hfill
			\begin{minipage}{0.24\textwidth}
				\includegraphics[width=\linewidth, height=3.5cm]{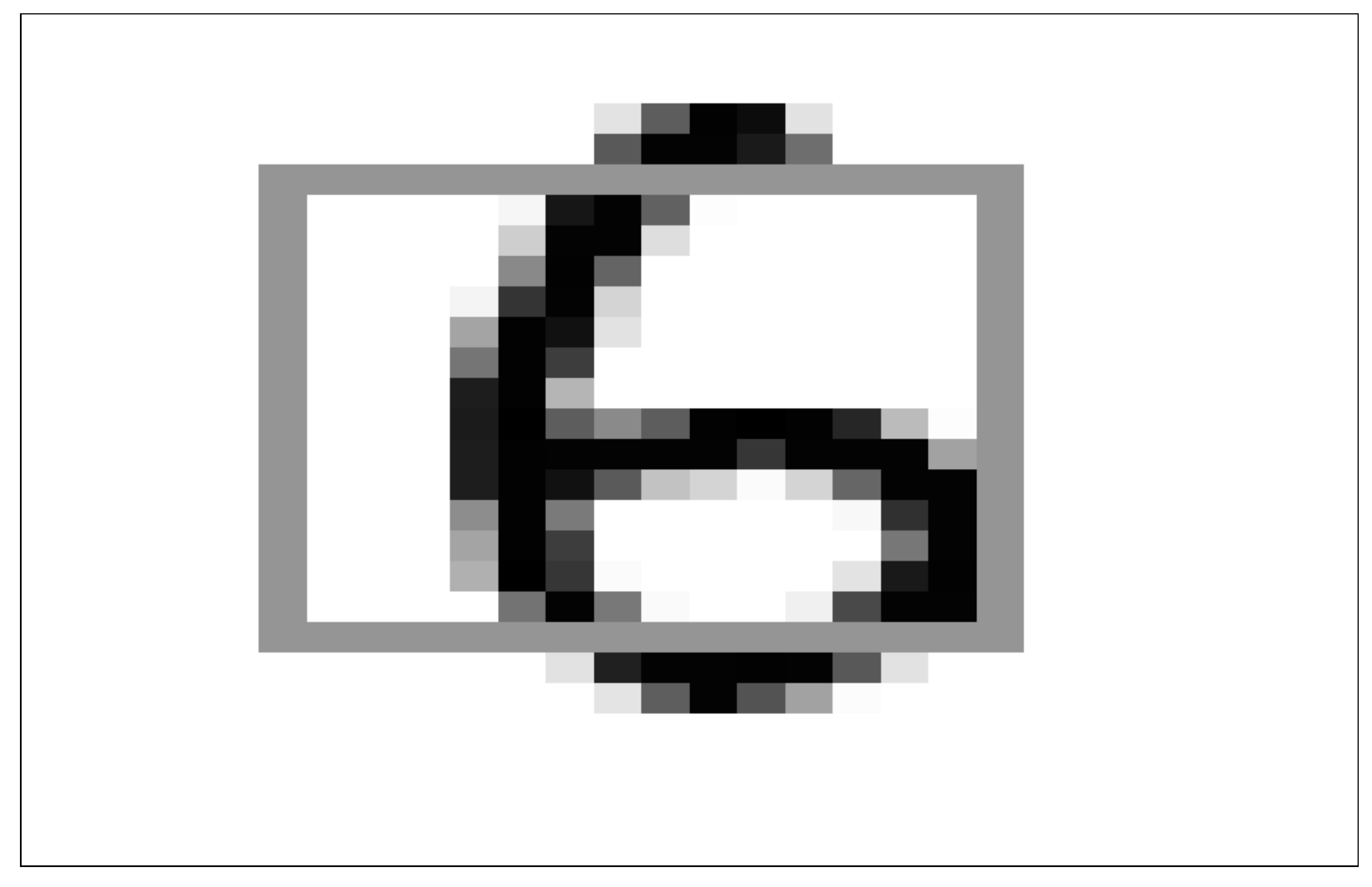}
				\subcaption{}
				\label{fig:dexample5}
			\end{minipage}

			\begin{minipage}{0.24\textwidth}
				\includegraphics[width=\linewidth, height=3.5cm]{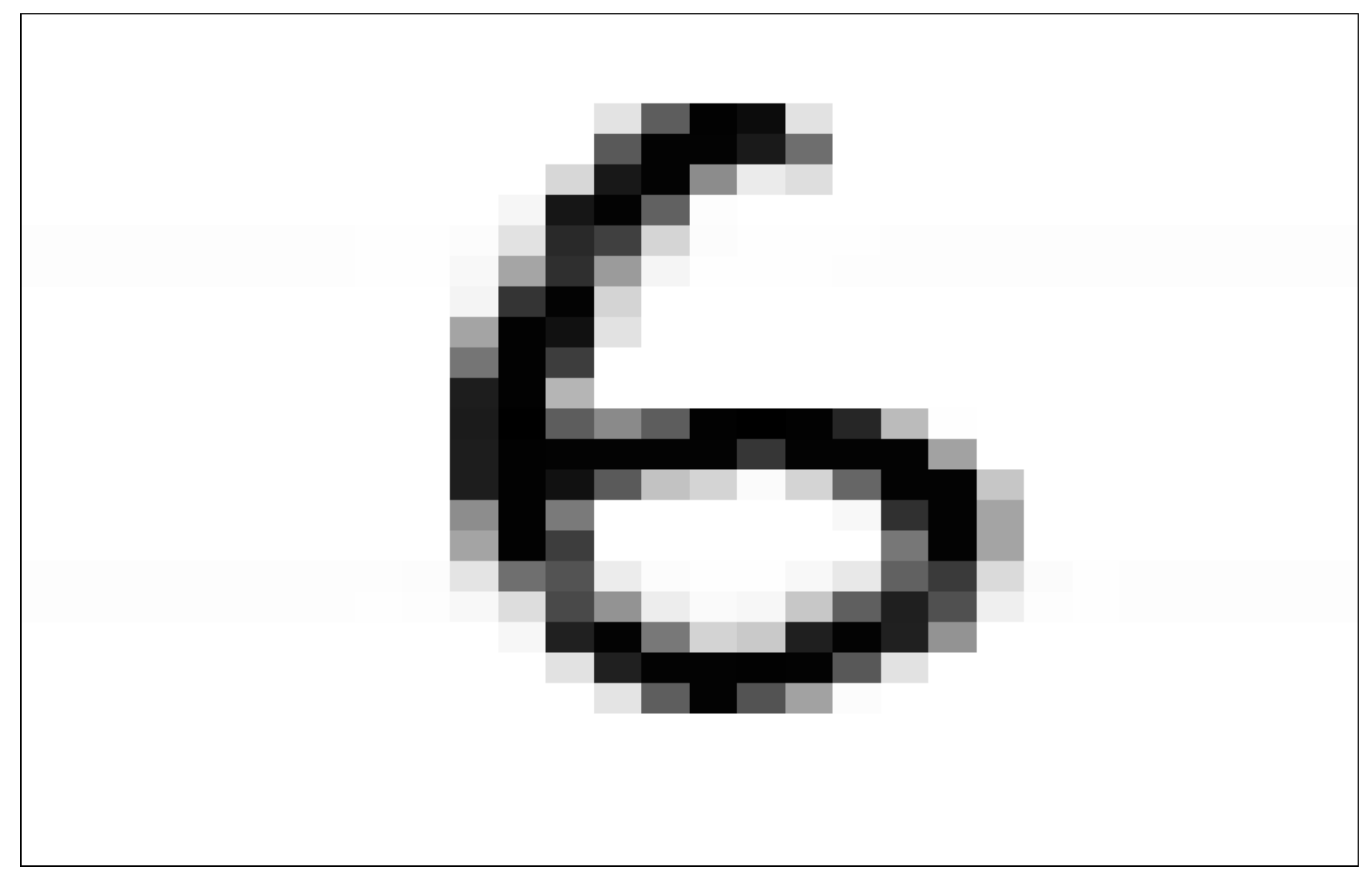}
				\subcaption{}
				\label{fig:exampleinpaint1}
			\end{minipage}
			\hfill
			\begin{minipage}{0.24\textwidth}
				\includegraphics[width=\linewidth, height=3.5cm]{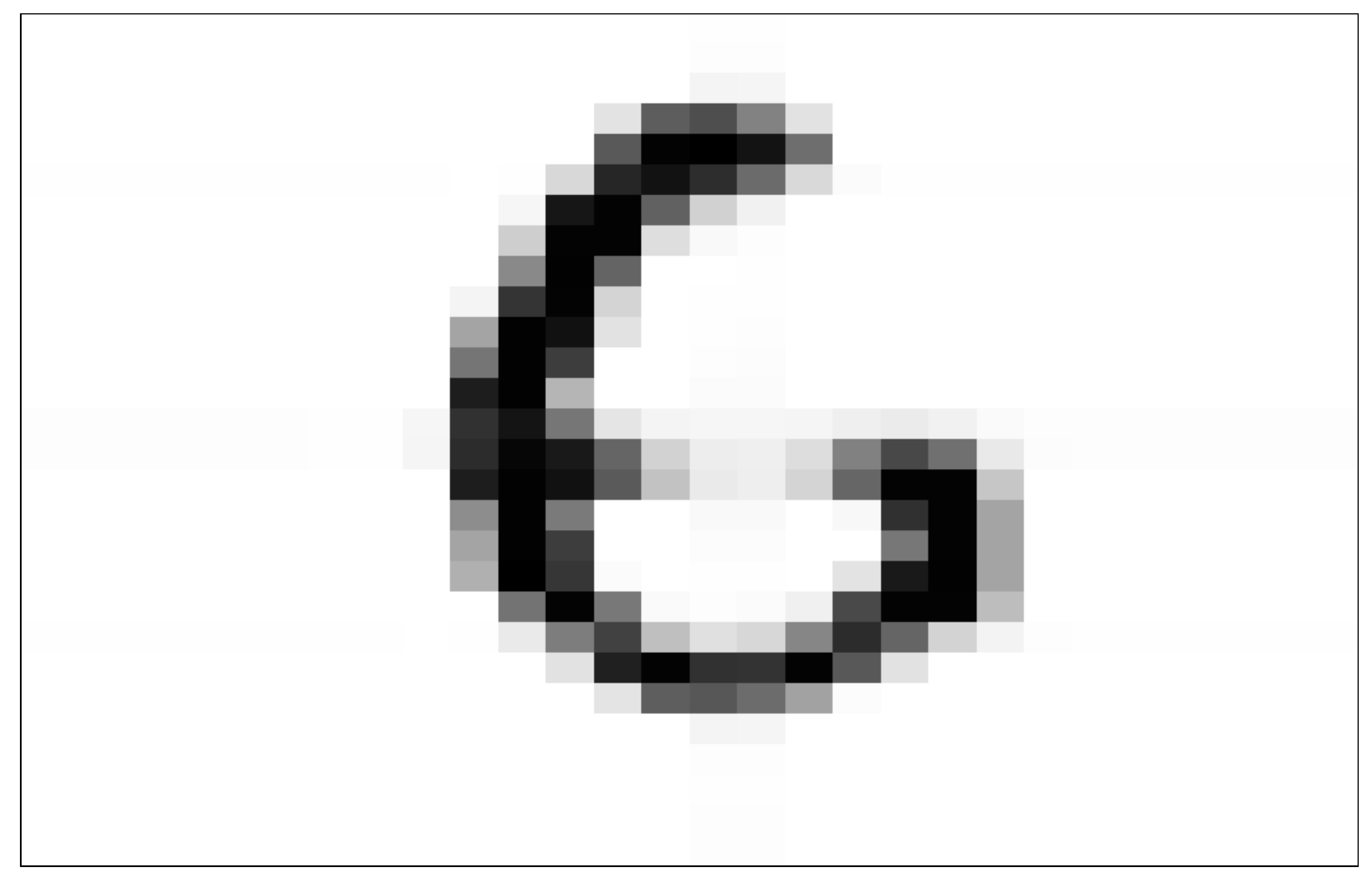}
				\subcaption{}
				\label{fig:exampleinpaint2}
			\end{minipage}
			\hfill
			\begin{minipage}{0.24\textwidth}
				\includegraphics[width=\linewidth, height=3.5cm]{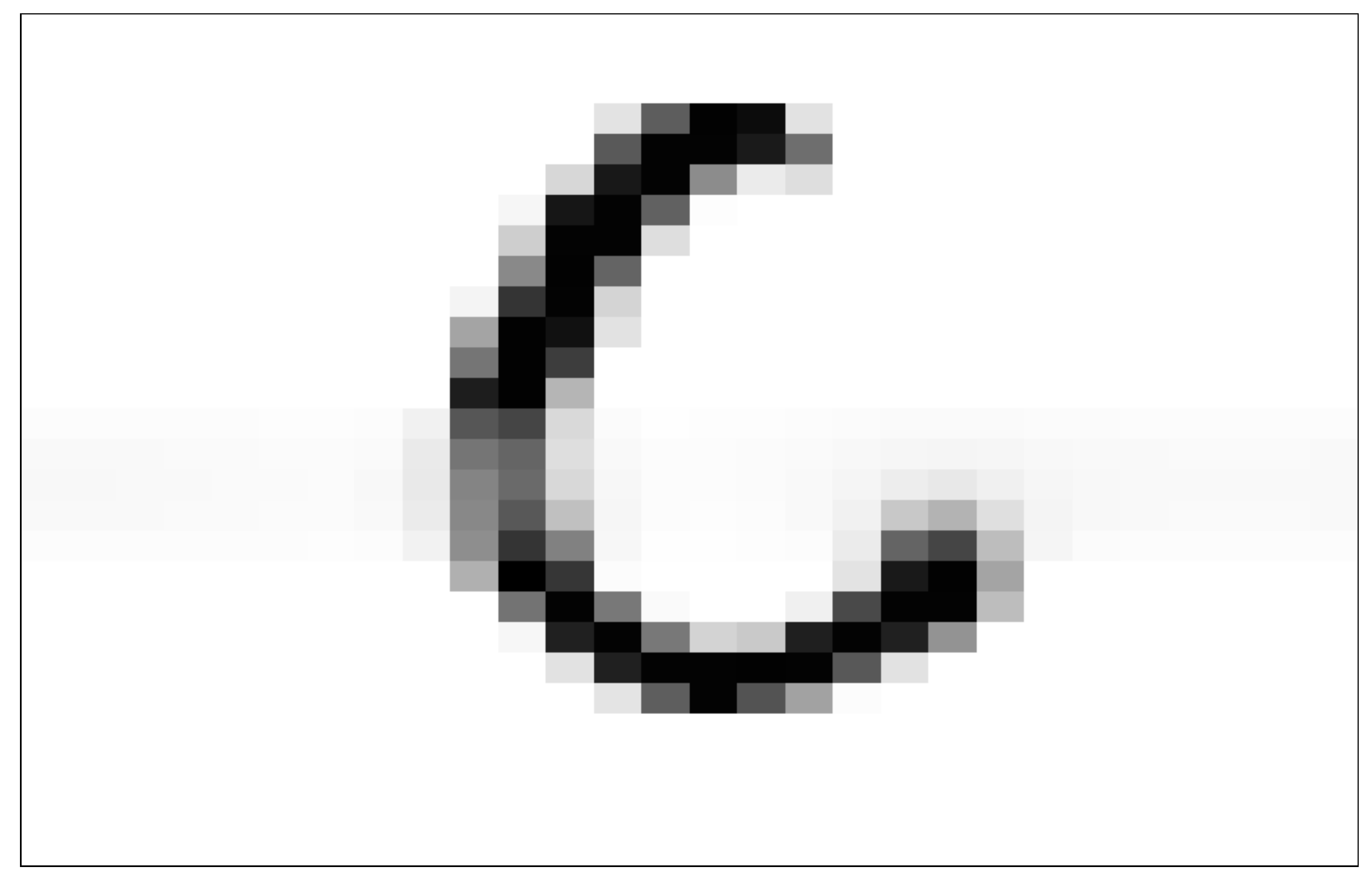}
				\subcaption{}
				\label{fig:exampleinpaint3}
			\end{minipage}
			\hfill
			\begin{minipage}{0.24\textwidth}
				\includegraphics[width=\linewidth, height=3.5cm]{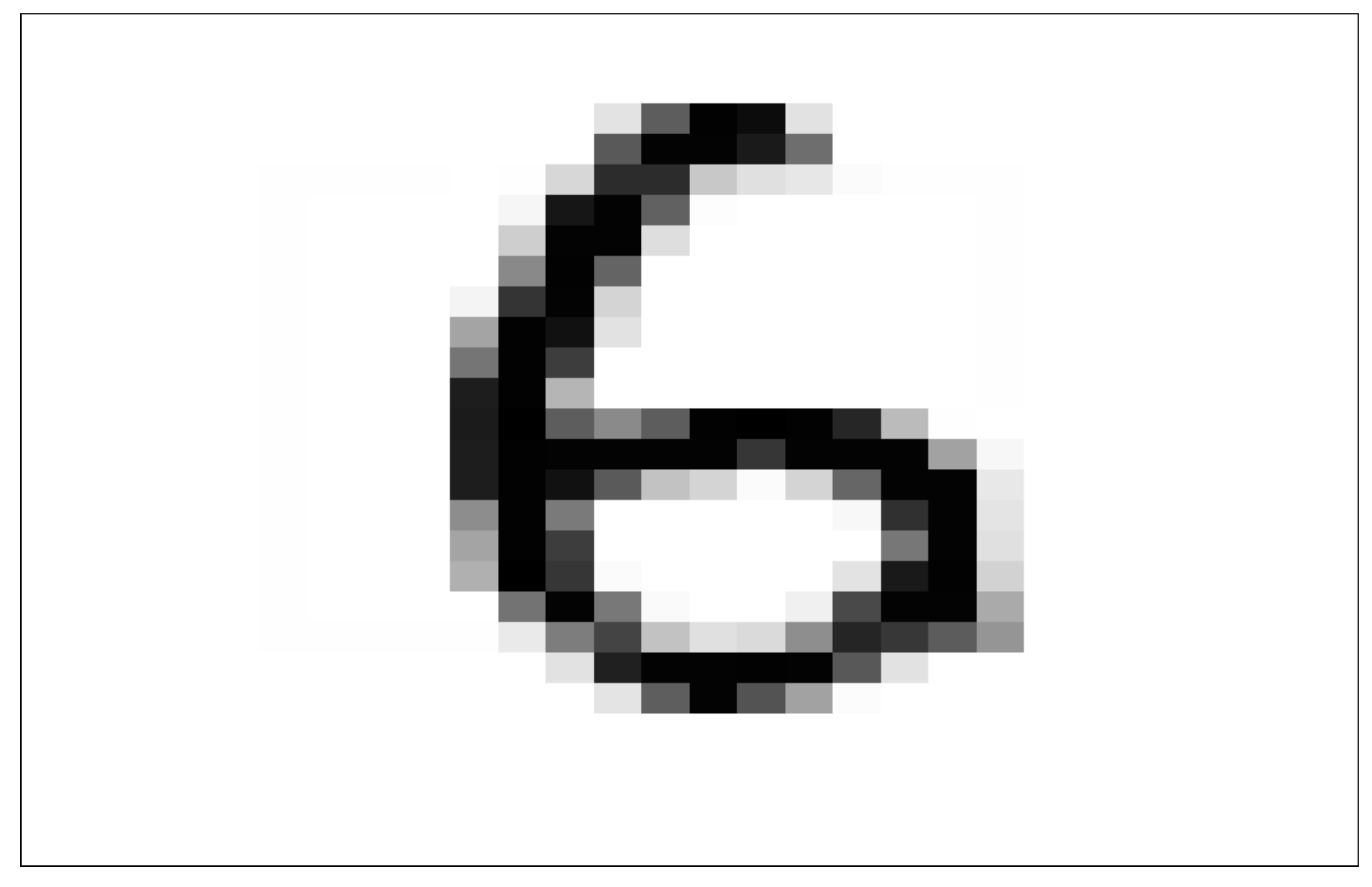}
				\subcaption{}
				\label{fig:exampleinpaint5}
			\end{minipage}
			\hfill
			
			\caption{Customized damage applied to a particular sample of the MNIST dataset. (A)-(D): the sample with four different types of damage; (E)-(F): The outcome of applying image inpainting to the damaged samples.}
			\label{fig:damageexample}
		\end{figure}
		
		\item Random damage: this second type of damage is inserted in a
		random fashion and with different levels of intensity. Two ways of randomly
		creating damage are considered: one makes use of randomly selecting whole
		horizontal rows of pixels, while the other is obtained by randomly
		selecting individual pixels. In addition, for both types of random damage
		we employ different levels of damage intensity, so that a higher percentage
		of the image contains damage if the intensity rises. This allows us to test
		how our image inpainting algorithm behaves with increasing levels of damage
		in the image. Examples of these damages are shown in \autoref{fig:random}.
		Similarly to the case of customized damage, the higher the intensity of
		damage the more information is lost in the inpainting, as we can see for
		example in the case of $80\%$ pixel damage in \autoref{fig:random}. But
		despite of this, our image inpainting algorithm renders recognisable images
		even with relative high amounts of damage.
	\end{enumerate}
	
	\begin{table}[h]
		\centering
		
		\begin{tabular}{|c|c|c|c|c|}
			\hline
			\multicolumn{2}{|c|}{Parameters} & $\epsilon_1$ & $\epsilon_2$ & $\lambda$ \\
			\hline
			\multicolumn{2}{|c|}{Customized damage} & 1.5 & 0.5 & 1000 \\
			\hline
			\multirow{2}{*}{Random damage} & Rows & 1.5 & 0.5 & 1000 \\
			\cline{2-5}
			& Pixels & 1.5 & 0.5 & 9000 \\
			\hline
		\end{tabular}
		\caption{Optimal parameters of the two-step algorithm applied in the MNIST dataset. $\epsilon_1$ and $\epsilon_2$ denote the values of $\epsilon$ used in the first and second step respectively, as explained in \autoref{subsec:two-step method}.}
		\label{tab:parameters}
	\end{table}
	
	The parameter values of $\lambda$, $\epsilon_1$ and $\epsilon_2$ are gathered
	in \autoref{tab:parameters}, and follow the reasoning discussed in
	\autoref{subsec:damage}. This choice of parameters in our finite-volume
	scheme in \autoref{subsec:fvm_mod} leads to an effective image-inpainting
	algorithm capable of restoring images with damage of varied nature as shown
	in \autoref{fig:damageexample} and \autoref{fig:random}. {Further details about the selection of $\lambda$, $\epsilon_1$ and $\epsilon_2$ are provided in \autoref{app:sen}, where a sensitivity analysis is provided in order to justify the parameter choices in \autoref{tab:parameters}}.
	The next step is to
	integrate this image-inpainting algorithm within a pattern recognition
	framework for the MNIST dataset.

	%
	%

	\subsection{Pattern recognition for inpainted images}\label{subsec:prediction}
	We now apply the neural network described in \autoref{subsec:neural network}
	to predict labels of damaged images with and without image inpainting, with
	the aim of quantifying the improvement of accuracy following the application of the CH
	filter. This study is completed for the different types and intensities of
	the noise depicted in \autoref{subsec:damage}.
	
	We begin by adding the types of damage in \autoref{subsec:damage} to the
	10,000 samples of the MNIST test dataset. The next step is to apply the CH
	filter and two-step method to each one of them, while also saving copies of
	the test images with the damage. Eventually, for each type of damage we get
	two batches of 10,000 images: one still with the damage, and another one with
	image inpainting applied. Given that the neural network of
	\autoref{subsec:neural network} is already trained with the 60,000 samples of
	the MNIST training dataset, we can directly compute the accuracy of each
	of the two batches. This way we are able to assess the improvement in
	accuracy thanks to applying image inpainting to restore the damage.
	
	Here for the validation we just employ the accuracy metric, which is defined as follows
	\begin{equation*}
	\text{Accuracy} \equiv \frac{\text{Number of correct predictions}}{\text{Number of total predictions}}.
	\end{equation*}
	There are however many other metrics apart from the accuracy one
	that play a vital role in other classification problems: recall, precision, F1 score, true positive
	rate and so on.
	Here we believe that the accuracy metric is enough to draw conclusions about how the image inpainting is improving the predictions with
	respect to the damage images. This is due to the fact that the MNIST dataset is a balanced dataset, where there is generally no preference
	between false positives and false negatives.
	
	The measure of improvement between the batch of samples with image inpainting and the damaged ones without it is computed as
	\begin{equation}\label{eq:improvement}
	\text{Improvement} \equiv \frac{\text{Accuracy}_{\text{with CH filter}}-\text{Accuracy}_{\text{without CH filter}}}{\text{Accuracy}_{\text{without CH filter}}},
	\end{equation}
	and it basically represents the percentage of improvement that results from adding the CH filter to the prediction process.
	
	The results for all the types of damage under consideration are displayed
	in \autoref{tab:predict_cust}, \autoref{tab:predict_rows} and
	\autoref{tab:predict_pixel}. In \autoref{tab:predict_cust} we gather
	the prediction accuracies for the four customized damages displayed in
	\autoref{fig:damageexample}, as well as the prediction accuracy for the
	unmodified MNIST test set, which for our neural network architecture is 0.97.
	We observe that for the types of more intense customized damage B and C the
	accuracy prediction for the damaged images without CH filter drops to 0.71
	and 0.64, respectively. By applying the filter we find  that the accuracy
	predictions can significantly escalate to 0.93 and 0.82, leading to
	improvements of 31\% and 28\% respectively. The other two types of customized
	damage A and D are not as pervasive as B and C, and as a result the accuracy
	predictions are high even without applying the CH filter.
	
	\begin{table}[H]
		\centering
		
		\begin{tabular}{cccc}
			\toprule
			Customized damage & Without CH filter & With CH filter & Improvement \\
			\midrule
			A & 0.84 & 0.96  & 14\%\\
			B & 0.71 & 0.93  & 31\%\\
			C & 0.64 & 0.82  & 28\%\\
			D & 0.90 & 0.96  & 7\%\\
			Initial test images & - & 0.97 &-  \\
			\bottomrule
		\end{tabular}
		
		\caption{Accuracy for the test dataset of MNIST without and with the CH filter, for the customized damage in \autoref{fig:damageexample}. The improvement is computed following \eqref{eq:improvement}.}
		\label{tab:predict_cust}
	\end{table}
	
	In \autoref{tab:predict_rows} and \autoref{tab:predict_pixel} we test the accuracies for random damage with various levels of intensities.
	The objective here is to analyse how the CH filter responds when the damage occupies more and more space in the images,
	both for the case of rows or pixels, as displayed in \autoref{fig:random}. In \autoref{tab:predict_rows} we show the accuracies for a
	range of damaged rows between 6 and 26, bearing in mind that the dimensions of the MNIST images are $28\times28$.
	We observe that for low numbers of damaged rows the accuracy prediction even without the CH filter is high and it
	does not improve significantly by adding the filter. But then the improvement surges until reaching a maximum value of 47\% for 16 random
	damaged rows, where the prediction without CH filter is 0.55 and with CH
	filter 0.81. From larger numbers of damaged rows the accuracies drastically
	drop due to the large amount of information lost, and not even the image
	inpainting process is able to achieve decent accuracies. In the limit of
	damaged number of rows tending to 28 we observe that the accuracies are close
	to the ones of a dummy classifier with one out of ten chances of rightly
	guessing the label. In this limit there is no difference between adding the CH filter or not,
	and it turns out that the improvements are even negative.
	
	\begin{table}[h!]
		\centering
		\begin{tabular}{cccc}
			\toprule
			Damaged rows & Without CH filter & With CH filter & Improvement \\
			\midrule
			6 & 0.89 & 0.96 & 8\%   \\
			8 & 0.82 & 0.93 & 13\%   \\
			10 & 0.73 & 0.91 & 25\%  \\
			12 & 0.66 & 0.87 & 32\%  \\
			14 & 0.6 & 0.87 & 45\%  \\
			16 & 0.55 & 0.81 & 47\%  \\
			18 & 0.47 & 0.68 & 45\%  \\
			20 & 0.40 & 0.48 & 20\%  \\
			22 & 0.39 & 0.45 & 15\%  \\
			24 & 0.33 & 0.26 & -21\%  \\
			26 & 0.20 & 0.12 & -40\%  \\
			\bottomrule
		\end{tabular}
		\caption{Accuracy for the test dataset of MNIST without and with the CH filter, for the case of random damage in rows. The improvement is computed following \eqref{eq:improvement}.}
		\label{tab:predict_rows}
	\end{table}
	
	We observe a similar pattern for the case of random damaged pixels in \autoref{tab:predict_pixel}.
	For low percentages of damaged pixels the improvement of adding the CH filter is negligible and the accuracies with and without the
	filter are quite high. As we increase the percentage of damaged pixels the improvement escalates until it reaches 45\% for a scenario
	with 80\% of the pixels randomly damaged. For this case the accuracy prediction without the filter is just 0.55, but thanks to
	the filter it significantly increases to a decent value of 0.8.  For larger percentage of pixels the improvement and accuracies drop,
	and in the limit towards 100\% of damaged pixels we get close to the accuracy of a dummy classifier. This is due to the large loss of
	information that the original images have suffered.
	
	\begin{table}[h!]
		\centering
		\begin{tabular}{cccc}
			\toprule
			Damaged pixels & Without CH filter & With CH filter & Improvement\\
			\midrule
			$30\%$ & 0.93 & 0.93 & 0\%  \\
			$40\%$ & 0.96 & 0.96 & 0\%  \\
			$50\%$ & 0.91 &  0.95 & 4\% \\
			$60\%$ & 0.8 & 0.94 & 18\% \\
			$70\%$ & 0.75 & 0.93 & 24\% \\
			$80\%$ & 0.55 & 0.8 & 45\% \\
			$90\%$ & 0.39 & 0.46 & 18\% \\
			$92\%$ & 0.32 & 0.37 & 16\% \\
			$94\%$ & 0.33 & 0.34 & 3\% \\
			$96\%$ & 0.20 & 0.23 & 15\% \\
			\bottomrule
		\end{tabular}
		\caption{Accuracy for the test dataset of MNIST without and with the CH filter, for the case of random damage in pixels.
			The improvement is computed following \eqref{eq:improvement}.}
		\label{tab:predict_pixel}
	\end{table}
	
\begin{figure}[h!]
		\centering
		\begin{tabular}{cccccc}
			\toprule
			Initial Image & True label & Damaged image & $\quad P_D \quad$ & Inpainted image & $\quad P_I \quad$ \\
			\midrule
			& & & & & \\[-0.5cm]
			\includegraphics[width=0.13\linewidth, height=2cm, valign=m]{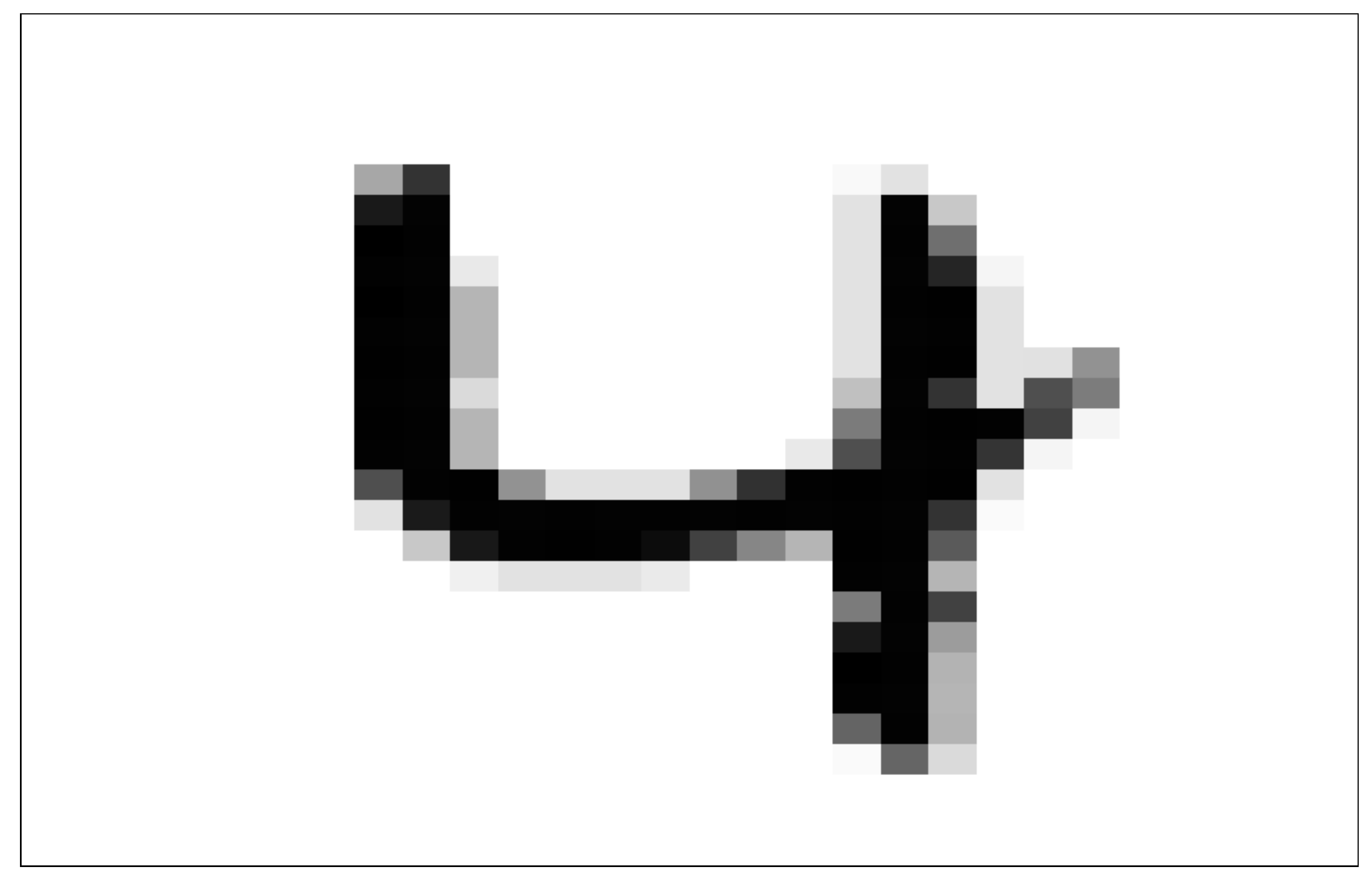}&4&
			\includegraphics[width=0.13\linewidth, height=2cm, valign=m]{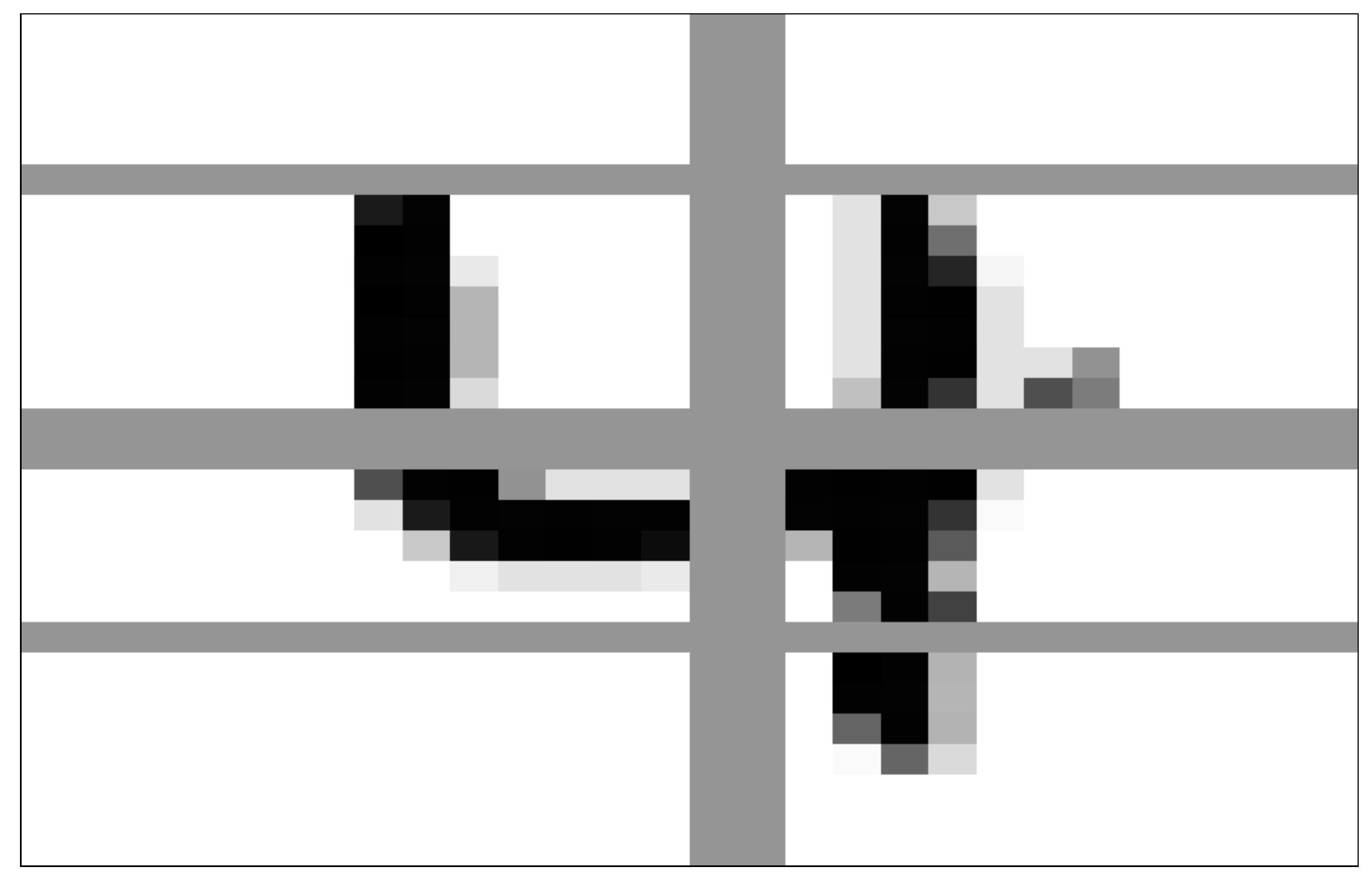}&6 &
			\includegraphics[width=0.13\linewidth, height=2cm, valign=m]{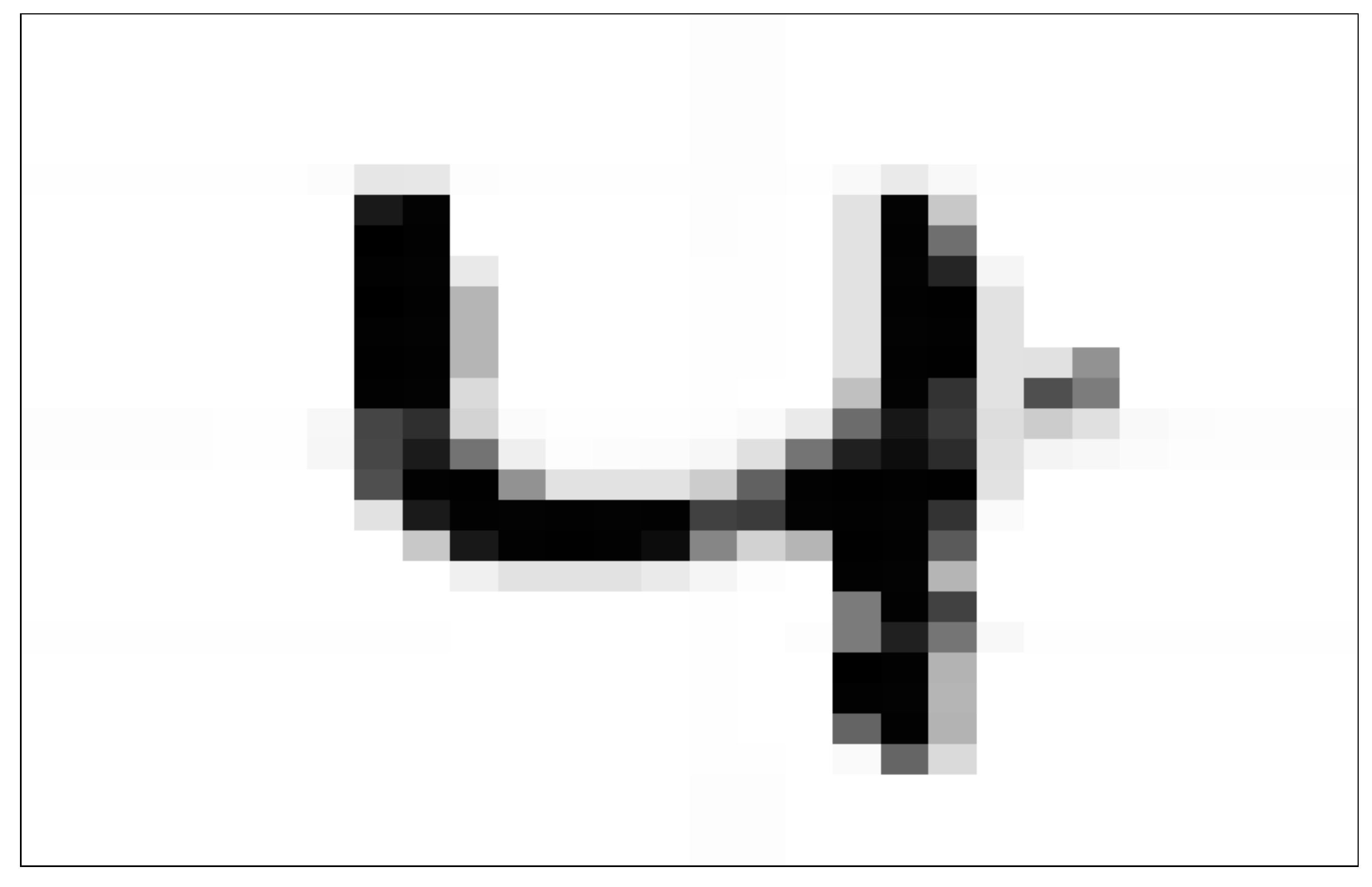}&4\\[1.5cm]
			\includegraphics[width=0.13\linewidth, height=2cm, valign=m]{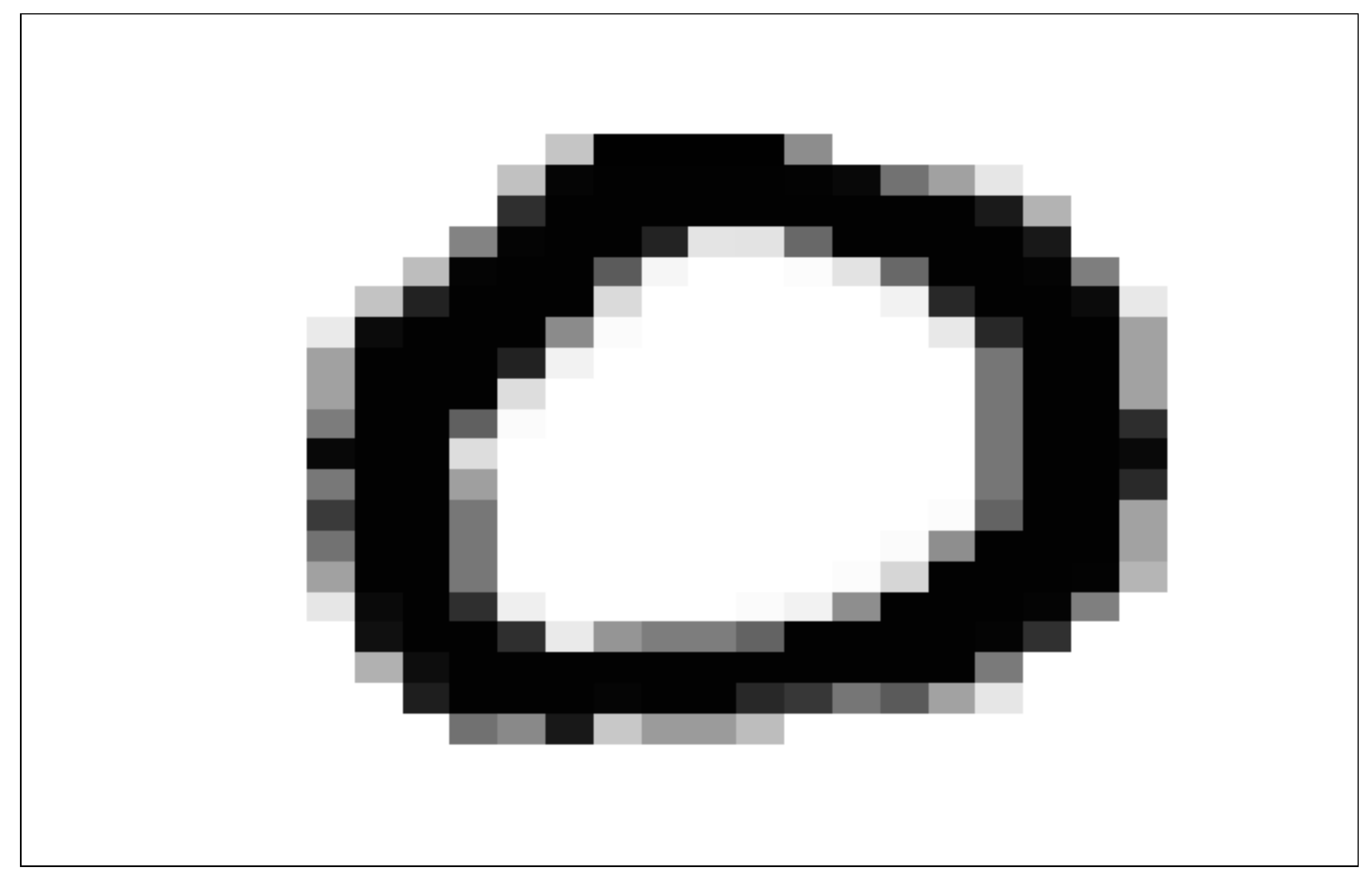}&0&
			\includegraphics[width=0.13\linewidth, height=2cm, valign=m]{damage_71_16R.pdf}& 8 &
			\includegraphics[width=0.13\linewidth, height=2cm, valign=m]{inpainting_71_16R.pdf}&0\\[1.5cm]
			\includegraphics[width=0.13\linewidth, height=2cm, valign=m]{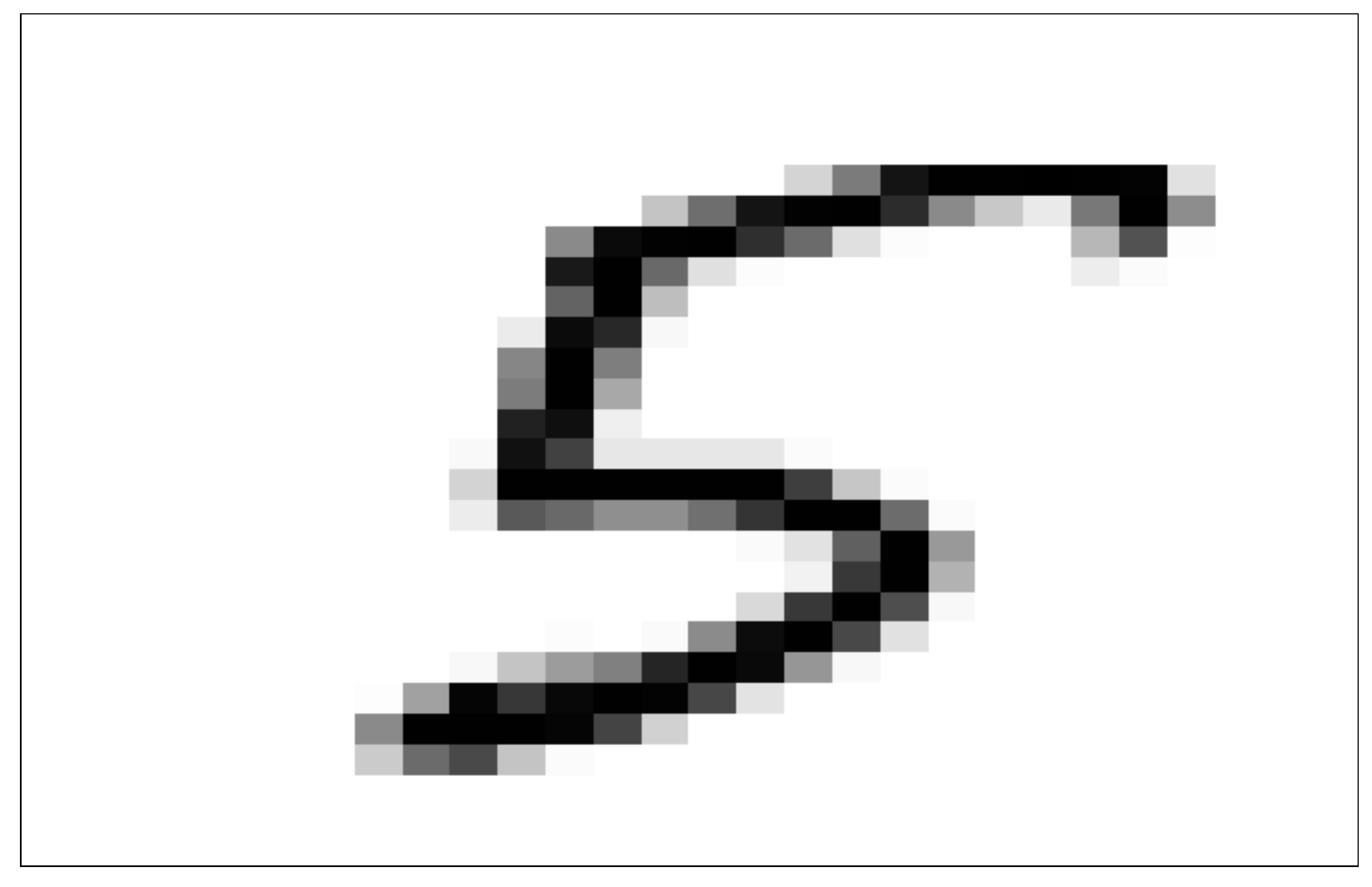}&5&
			\includegraphics[width=0.13\linewidth, height=2cm, valign=m]{damage_23_07.pdf}& 4 &
			\includegraphics[width=0.13\linewidth, height=2cm, valign=m]{inpainting_23_07.pdf}&5\\[1.5cm]
			
			\bottomrule
			
		\end{tabular}
		
		\caption{Particular examples of label predictions for MNIST samples with various types of damage: customized damage C, random damage in 16
			rows and random damage in $70\%$ of pixels. The label of the damaged image is wrongly predicted, while the label for the inpainted image is
			correctly predicted. $P_D$ and $P_I$ represent the label predictions of the damaged and inpainted images, respectively.}
		\label{fig:wrong_predictions} 
	\end{figure}

	In \autoref{fig:wrong_predictions} we depict some specific examples for
	which the label is only predicted correctly after applying the CH filter to
	the damaged image. These are just some particular samples out of the 10,000 images
	contained in the test dataset of MNIST, {and for some of them the opposite
		effect can occur: that the label is correctly predicted for the
		damaged image but following the inpainting process it is predicted
		incorrectly. This only occurs for really severe damages, where there is no difference between adding the CH filter or not and the accuracies are
		close to the ones of a dummy classifier guessing randomly.}  However, we have
	shown in
	\autoref{tab:predict_rows} and
	\autoref{tab:predict_pixel} that overall the CH filter increases the global
	accuracy for images with low to moderate damage, and consequently we expect
	that scenarios such as the ones displayed in \autoref{fig:wrong_predictions}
	are much more common than the opposite ones.
	
	{In \autoref{fig:wrong_predictions} we also appreciate that some greyscale pixels remain in the inpainted image after applying the modified CH filter. This is due to the fact that the finite-volume scheme has not run for long enough to turn those intensities into the stable white ($\ph=-1$) or black ($\ph=1$) phase field. This is prone to happening for images with high levels of damage, and one should try to let the simulation run for as long as possible to avoid possible misclassification issues due to those greyscale regions.}
	
	%
	%
	
	\section{Discussion and conclusions}\label{sec:con_fwork}

	We have quantified the prediction improvement of employing a CH image
	inpainting filter to restore damaged images which are then passed into a
	neural network. We combined a finite-volume scheme with a neural network for
	pattern recognition to develop an integrated algorithm summing up the process
	of adding damage to the images and then predicting their label. Our results
	for the MNIST dataset suggest that, in general, the accuracy is improved for a
	wide range of low to moderate damages, while for some particular cases we
	reach improvements of up to 50\%. We also provide the image inpainting
	outcome of multiple damage scenarios and the benefits of adding the CH filter
	to predict the label of the image are easily visible.
	
	We believe that our results employing the MNIST dataset lay the foundations
	towards the application of image inpainting in more complex datasets. Here we
	have demonstrated the benefit of combining the fields of image inpainting
	with machine learning, and we believe that many applications can take
	advantage of it. For instance there are applications such as medical images
	from MRI or satellite observations where there is typically some inherent
	noise or damage involved and where there may be potential to employ tools
	from machine learning as was done here.
	
	MNIST is one of the most well-known and convenient benchmark datasets. It is possible
	that applying our methodology to increasingly complex datasets might
	bring about new challenges. At the same time we have relied on two main assumptions
	about the damage and the images: the first one is that the images are binary,
	leading to the standard CH potential in \eqref{eq:doublewellpot} that has
	only two wells (i.e. one for each of the two colours); the second one is that
	the damage is not blind, meaning that the location of the damage is known.
	Performing image inpainting without these two assumptions becomes
	substantially more involved as already pointed out in
	\cite{cherfils2016cahn,burger2009cahn,xie2012image}. We will be exploring
	these and related questions in future works.
	
	\vspace{-0.3cm}
	
	%
	%
	\section*{Acknowledgements}
	We are indebted to P. Yatsyshin and A. Russo from the Chemical Engineering Department of Imperial College (IC) for numerous stimulating discussions on machine learning and image inpainting. S.~P.~Perez acknowledges financial support from the IC
	President's PhD Scholarship. JAC was partially supported by the EPSRC through grant no. EP/P031587 and
	the ERC through Advanced Grant no. 883363. SK was partially supported by
	the EPSRC through grant no. EP/L020564 and the ERC through Advanced Grant
	no. 247031.

	
	\appendix

	\section{{Sensitivity analysis of the parameters $\epsilon_1$, $\epsilon_2$ and $\lambda$}}\label{app:sen}
	{Adequately tuning the two values of $\epsilon_1$,  $\epsilon_2$ and
		$\lambda$, is vital to complete a successful image inpainting. These values
		have to be chosen empirically and depend on the dataset and type of damage.
		Here we provide further details about finding an adequate combination of
		these parameters for images of MNIST. }
	
\begin{figure}[ht!]
		\begin{center}
			\begin{minipage}{0.48\textwidth}
				\begin{center}
					\includegraphics[height=3cm,width=3cm]{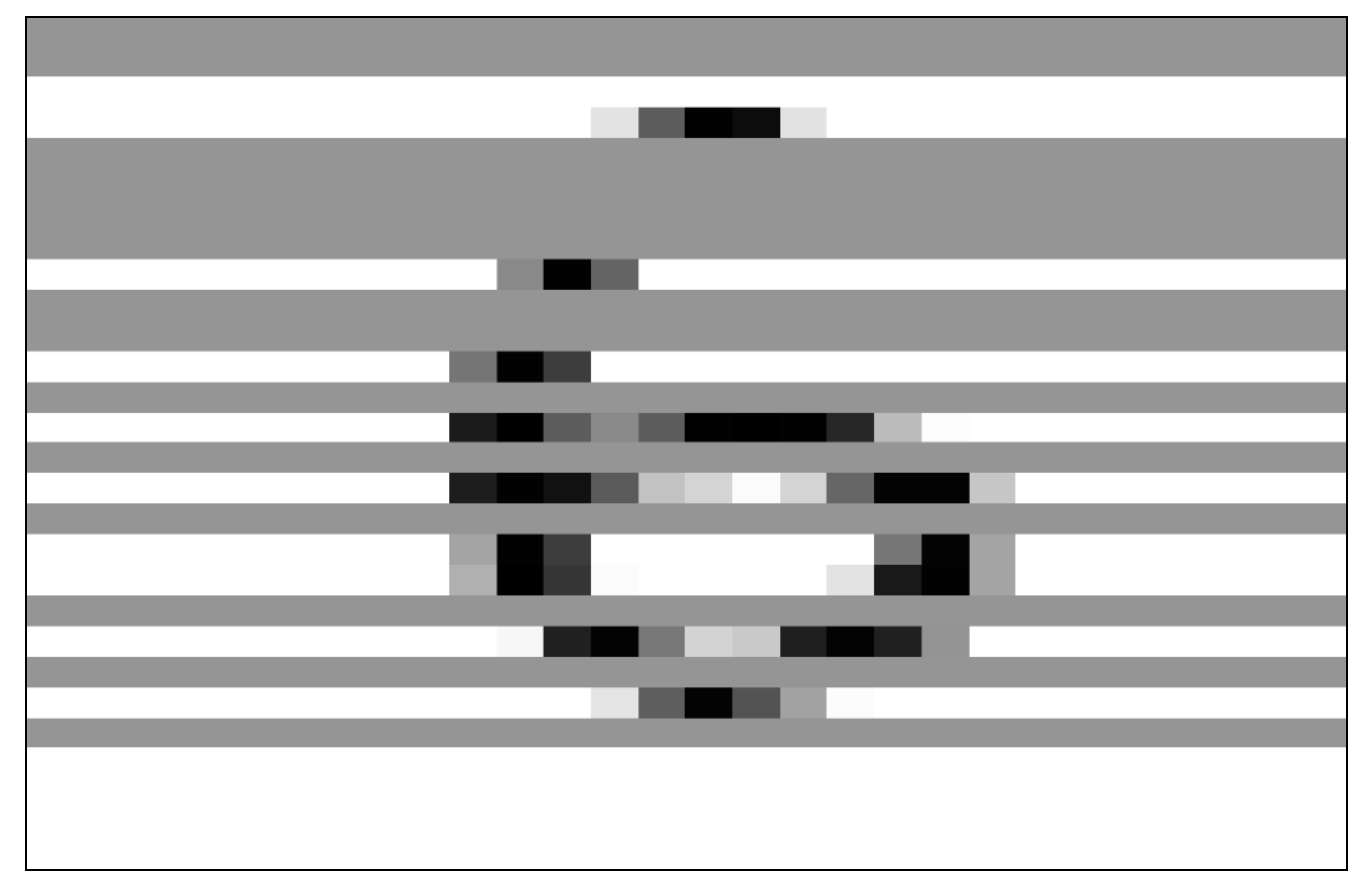}
					\subcaption{Original image with damage}
					\label{subfig:sen1}
				\end{center}
			\end{minipage}
			\begin{minipage}{0.48\textwidth}
				\begin{center}
					\includegraphics[height=3cm,width=3cm]{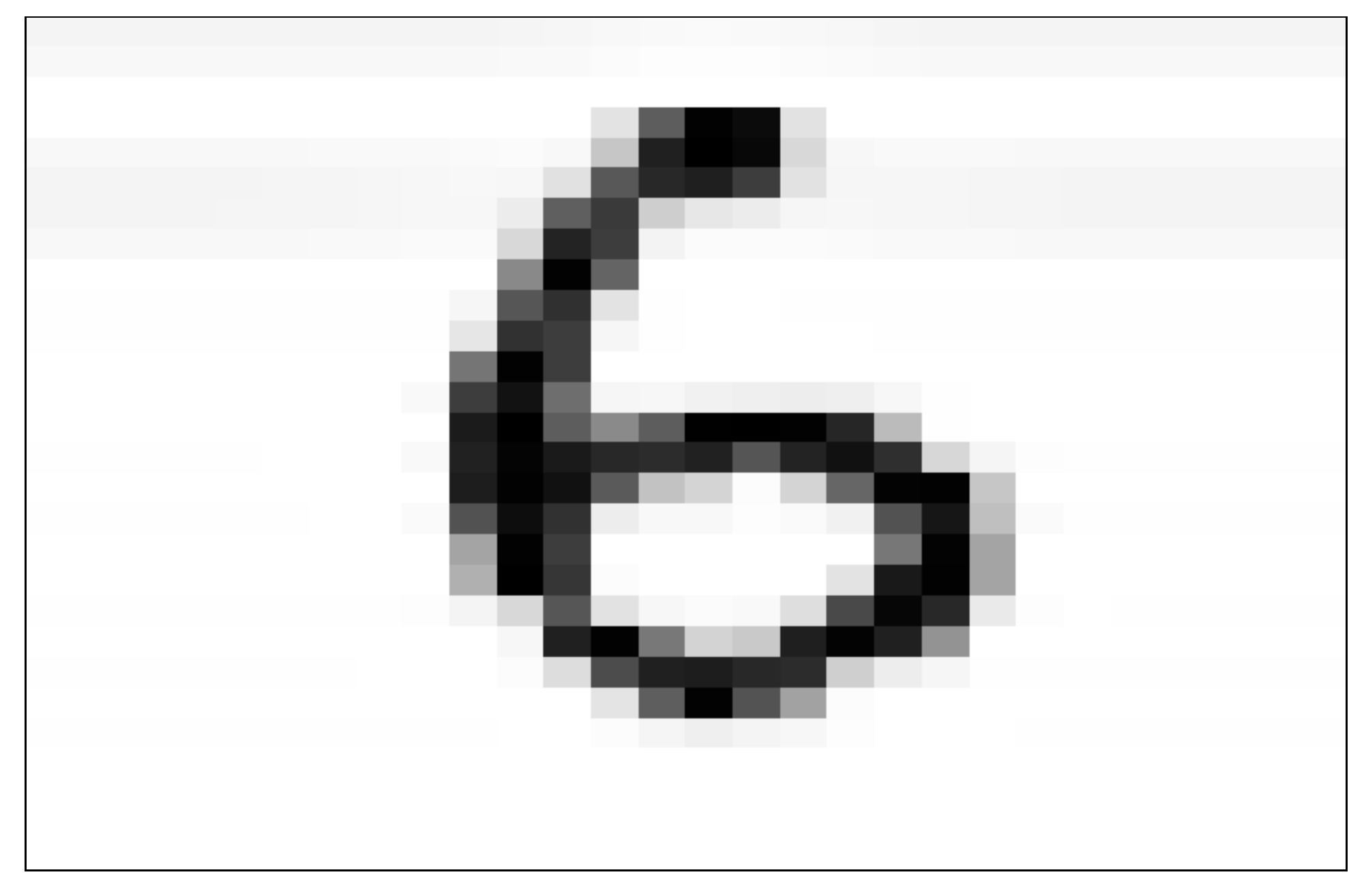}
					\subcaption{$\epsilon_1=1.5$, $\epsilon_2=0.5$ and $\lambda=1000$}
					\label{subfig:sen2}
				\end{center}
			\end{minipage}
			\hfill
			\vspace{0.5cm}
			
			\begin{minipage}{0.48\textwidth}
				\begin{center}
					\includegraphics[height=3cm,width=3cm]{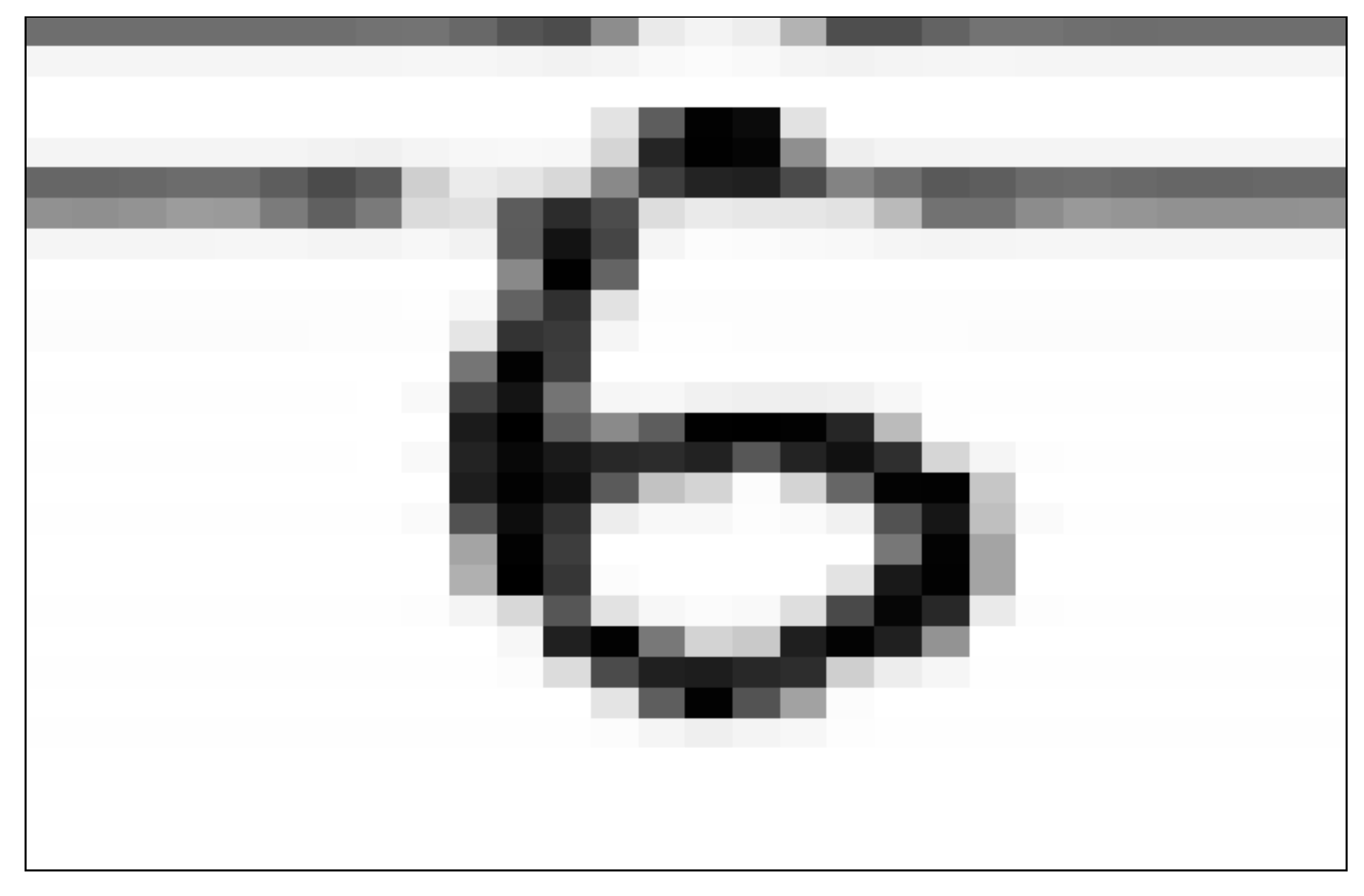}
					\subcaption{$\epsilon_1=0.5$, $\epsilon_2=0.5$ and $\lambda=1000$}
					\label{subfig:sen3}
				\end{center}
			\end{minipage}
			\begin{minipage}{0.48\textwidth}
				\begin{center}
					\includegraphics[height=3cm,width=3cm]{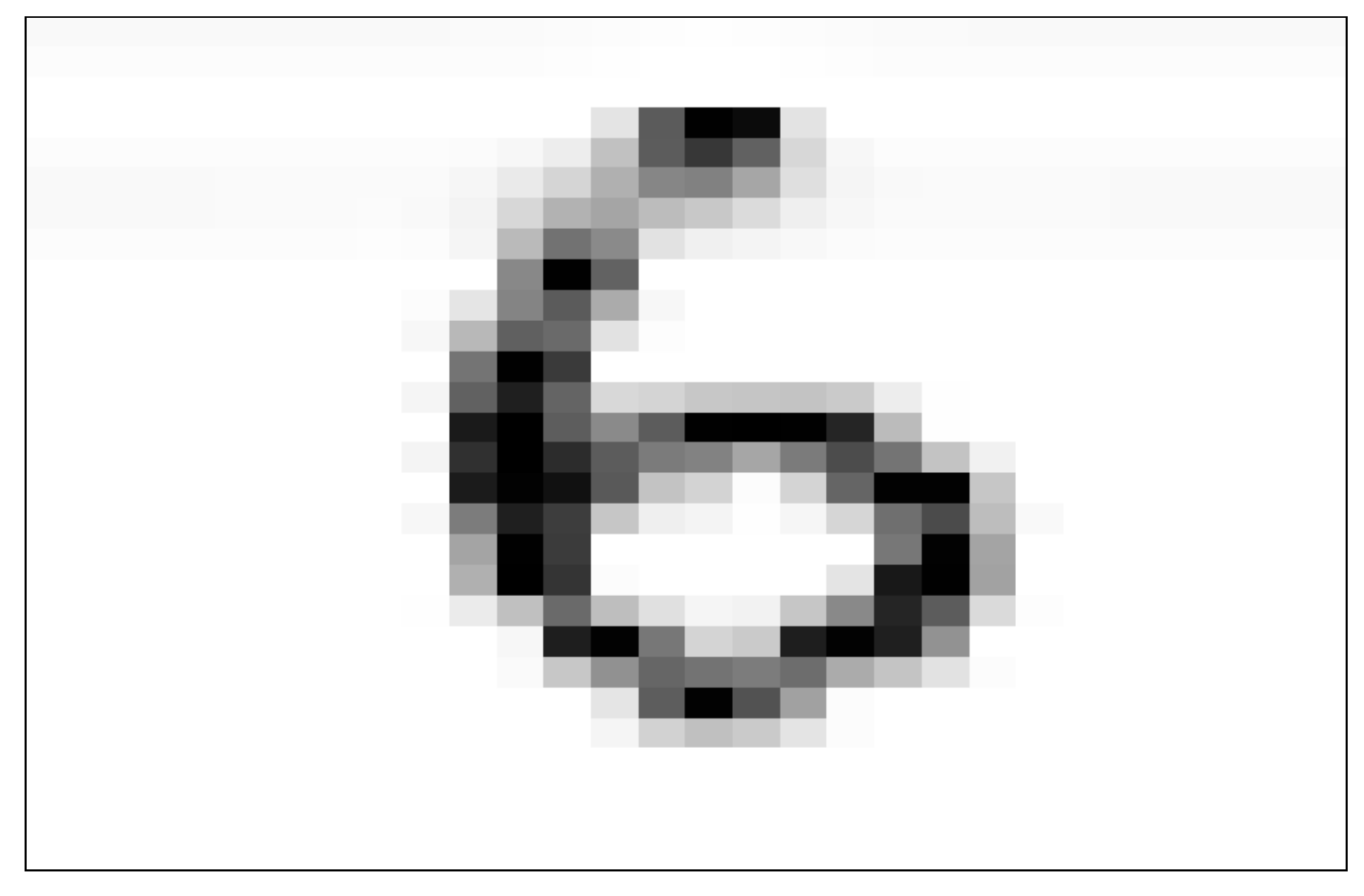}
					\subcaption{$\epsilon_1=1.5$, $\epsilon_2=1.5$ and $\lambda=1000$}
					\label{subfig:sen4}
				\end{center}
			\end{minipage}
			\vspace{0.5cm}
			
			\begin{center}
				\begin{minipage}{0.28\textwidth}
					\begin{center}
						\includegraphics[height=3cm,width=3cm]{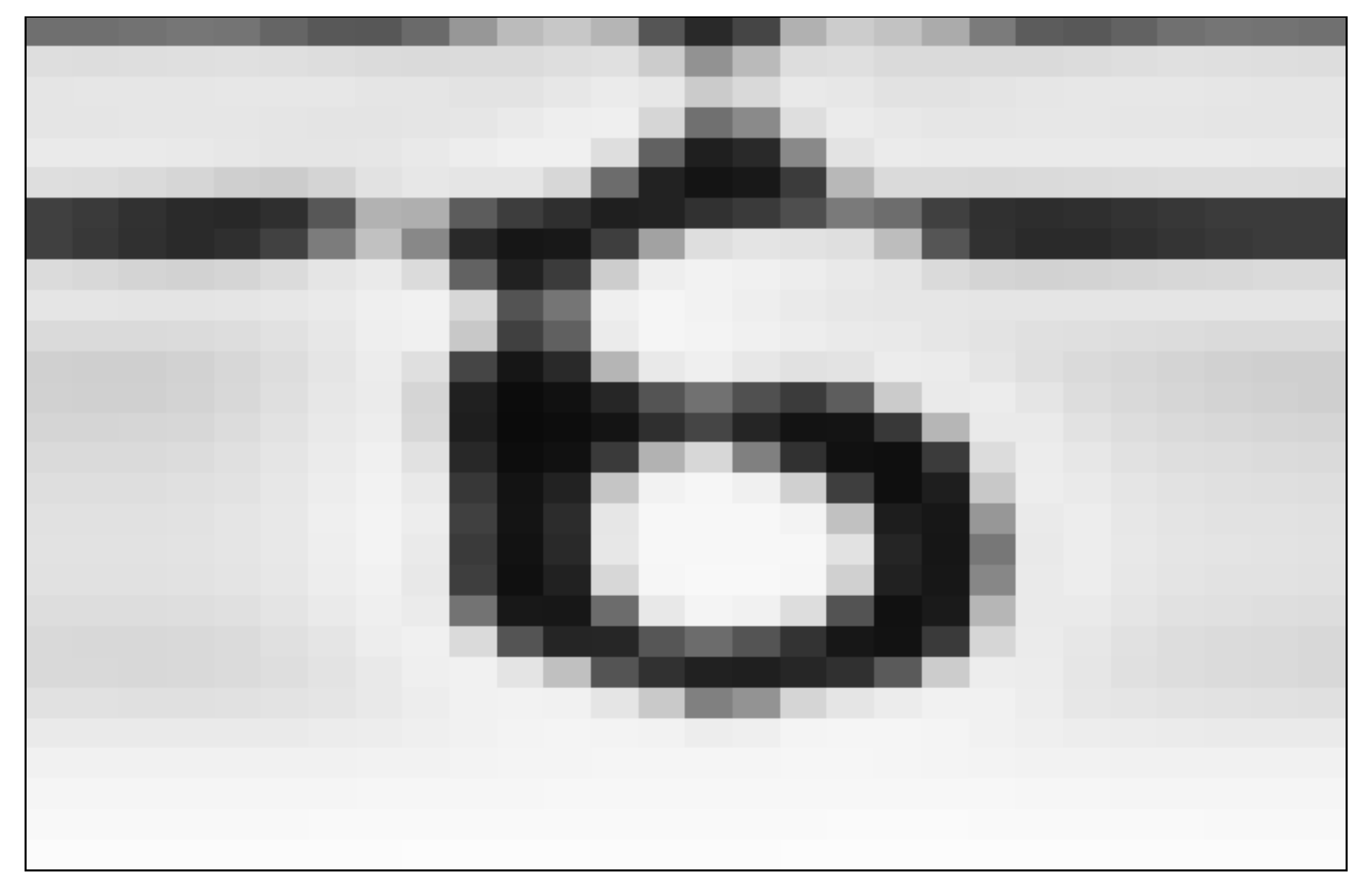}
						\subcaption{$\epsilon_1=1.5$, $\epsilon_2=0.5$ and $\lambda=0.01$}
						\label{subfig:sen5}
					\end{center}
				\end{minipage}
				\hspace{0.3cm}
				\begin{minipage}{0.28\textwidth}
					\begin{center}
						\includegraphics[height=3cm,width=3cm]{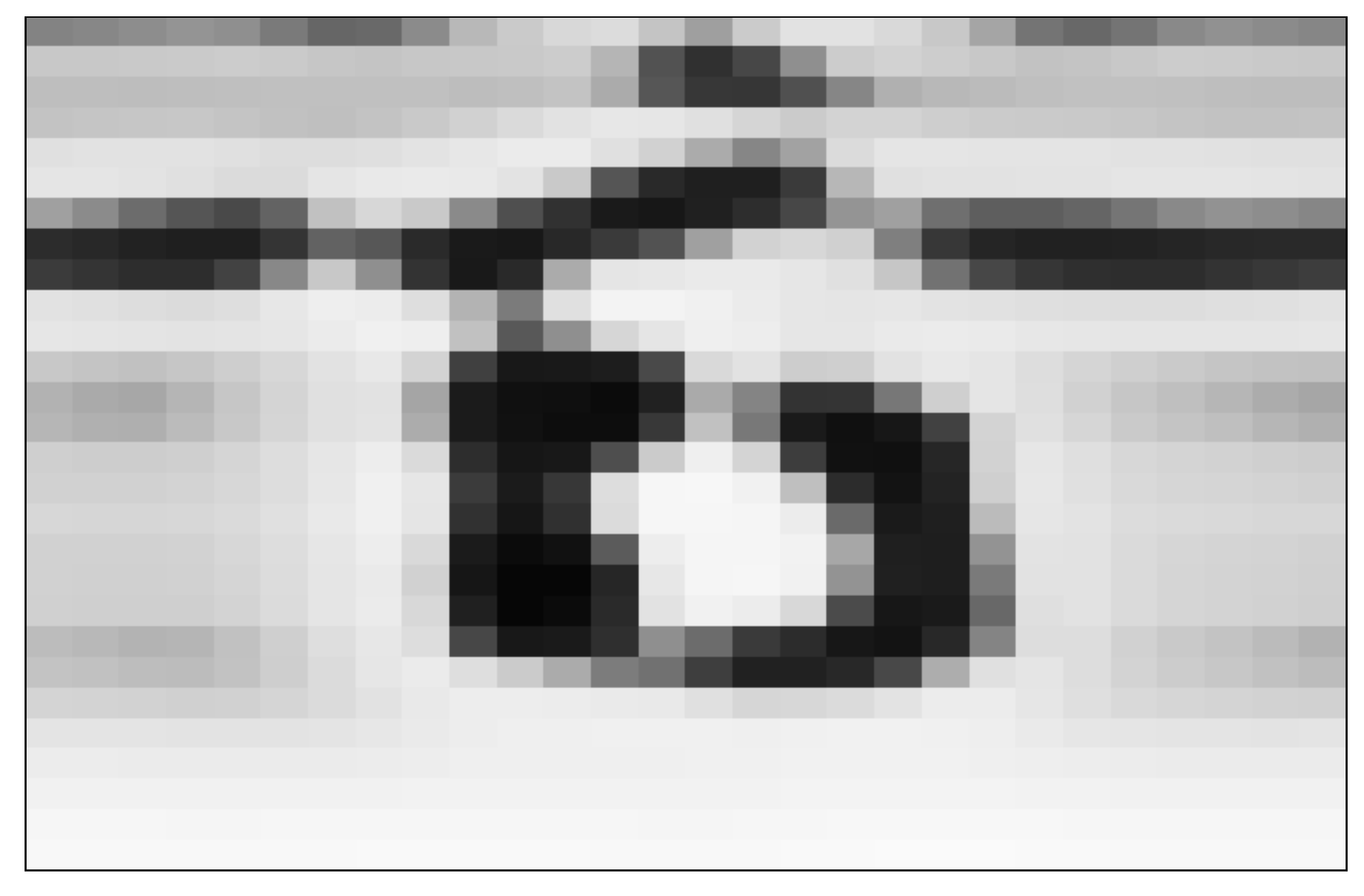}
						\subcaption{$\epsilon_1=1.5$, $\epsilon_2=0.5$ and $\lambda=0.1$}
						\label{subfig:sen6}
					\end{center}
				\end{minipage}
				\hspace{0.3cm}
				\begin{minipage}{0.28\textwidth}
					\begin{center}
						\includegraphics[height=3cm,width=3cm]{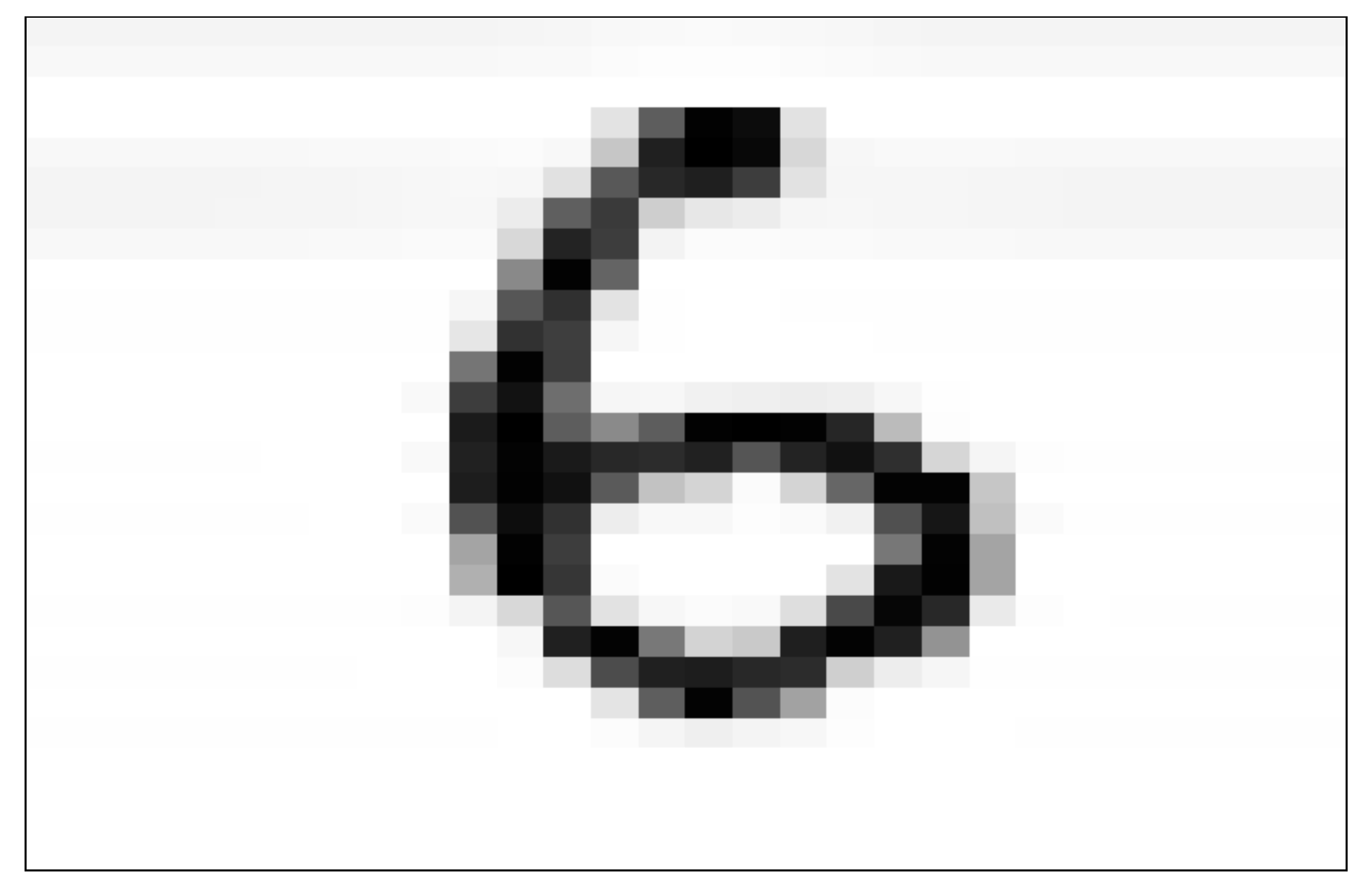}
						\subcaption{$\epsilon_1=1.5$, $\epsilon_2=0.5$ and $\lambda=10000$}
						\label{subfig:sen7}
					\end{center}
				\end{minipage}
			\end{center}
			
		\end{center}
		\protect\protect\caption{\label{fig:sen} Sensitivity analysis for different values of $\epsilon_1$, $\epsilon_2$ and $\lambda$. (A): original image with damage; (B): optimal choice of parameters; (C): short value of $\epsilon_1$ and $\epsilon_2$; (D): large value of $\epsilon_1$ and $\epsilon_2$; (E): really small value of $\lambda$; (F): small value of $\lambda$; (G): large value of $\lambda$.}
		
	\end{figure}

	{On the one hand, the parameter $\epsilon$ is related to the interphase
		thickness of the solution of the modified CH equation in \eqref{eq:ch_modif}.
		The larger the $\epsilon$, the more diffused the resulting image. However, a
		large $\epsilon$ allows for a large-scale topological reconnection of shapes,
		which is convenient when the damage breaks a connected color phase. The
		strategy proposed in \cite{bertozzi2006inpainting} and followed here,
		consists of employing a two-stage methodology where a large $\epsilon$ is
		employed first (denoted as $\epsilon_1$), followed by a smaller $\epsilon$
		(denoted as $\epsilon_2$) in order to sharpen the final image. In
		\autoref{fig:sen} we depict some of the image-inpainted images resulting from
		various combinations of $\epsilon_1$ and $\epsilon_2$. The original image
		with damage is shown in \autoref{subfig:sen1}, and the optimal combination of
		$\epsilon_1$ and $\epsilon_2$ following the values in
		\autoref{tab:parameters} is depicted in \autoref{subfig:sen2}. For this
		example the optimal parameters are $\epsilon_1=1.5$ and $\epsilon_2=0.5$, and
		depend on the mesh size that in this case is $\Delta x= \Delta y =1$. A
		larger mesh size requires a larger value of $\epsilon$ to produce the same
		outcome, implying that both parameters are interconnected and have to be of
		the same order. In order to evaluate the impact of $\epsilon_1$ and
		$\epsilon_2$ we run two experiments: in the first one we choose both of them
		with the small value of $\epsilon_1=\epsilon_2=0.5$, and in the second one
		both parameters take the large value $\epsilon_1=\epsilon_2=1.5$. The outcome
		with the small value is depicted in \autoref{subfig:sen3}, and clearly there
		is not an efficient large-scale topological reconnection of the white
		regions, resulting in some damage left in the image. The outcome with the
		large value is depicted in \autoref{subfig:sen4}, and the drawback here is
		that the image is too diffused in comparison to the optimal combination in
		\autoref{subfig:sen2}.}
	
	{On the other hand, the parameter $\lambda$ in the finite-volume scheme
		in \eqref{eq:fv_2} ensures that the undamaged pixels are not modified during
		the image inpainting. The choice of $\lambda$ has to be sufficiently large to
		counterbalance the fluxes of the scheme and act as a penalty term that keeps
		the undamaged pixels invariant. However, $\lambda$ cannot be too large since
		otherwise the finite-volume scheme becomes a singularly perturbed problem,
		and the convergence of the implicit scheme is deteriorated. The value of
		$\lambda$ is related to the time step $\Delta t$, and greater values of
		$\lambda$ are possible if $\Delta t$ is refined. In our case, with the choice
		of $\Delta t=0.1$, our finite-volume scheme does not yield any result and
		breaks down during the simulation for values of $\lambda\notin[1,10000]$. To
		evaluate the impact of $\lambda$ we run three experiments: in the first one
		we choose the really small value $\lambda=0.01$, in the second the small
		$\lambda=0.1$ and in the third the large value $\lambda=10000$. The outcomes
		with the two small values are depicted in \autoref{subfig:sen5} and
		\autoref{subfig:sen6}, and we observe that the undamaged pixels have been
		altered and the overall appearance is not clean. The outcome with the large
		value is depicted in \autoref{subfig:sen7}, and the image is comparable to
		the optimal one in \autoref{subfig:sen2}. There is a threshold-$\lambda$
		value above which there is no convergence of the implicit finite-volume
		scheme. For the particular image we found this threshold can be approximated
		by $\lambda=10000$.}

	\bibliographystyle{siam}
	\bibliography{references}

\begin{thebibliography}{10}

\bibitem{Kaggle}
{\em Digit recognizer competition in {K}aggle}.
\newblock https://www.kaggle.com/c/digit-recognizer/overview.
\newblock Accessed: 2020-03-05.

\bibitem{TensorFlow}
{\em Official website of {T}ensor{F}low}.
\newblock https://www.tensorflow.org/.
\newblock Accessed: 2020-03-05.

\bibitem{Ala-Nissila2004}
{\sc T.~Ala-Nissila, S.~Majaniemi, and K.~Elder}, {\em Phase-field modeling of
  dynamical interface phenomena in fluids}, Lect. Notes Phys., 640 (2004),
  pp.~357--388.

\bibitem{aymard2019linear}
{\sc B.~Aymard, U.~Vaes, M.~Pradas, and S.~Kalliadasis}, {\em A linear,
  second-order, energy stable, fully adaptive finite element method for
  phase-field modelling of wetting phenomena}, Journal of Computational
  Physics: X, 2 (2019), p.~100010.

\bibitem{bailo2018fully}
{\sc R.~Bailo, J.~A. Carrillo, and J.~Hu}, {\em Fully discrete
  positivity-preserving and energy-dissipating schemes for
  aggregation-diffusion equations with a gradient flow structure}, arXiv
  preprint arXiv:1811.11502, to appear in Comm. Math. Sci.,  (2018).

\bibitem{refCH}
{\sc R.~Bailo, J.~A. Carrillo, S.~Kalliadasis, and S.~P. Perez}, {\em
  Energy-stable and bounded finite volume schemes for the {C}ahn-{H}illiard
  equation}, work in preparation.

\bibitem{barrett1999finite}
{\sc J.~W. Barrett, J.~F. Blowey, and H.~Garcke}, {\em Finite element
  approximation of the {C}ahn--{H}illiard equation with degenerate mobility},
  SIAM Journal on Numerical Analysis, 37 (1999), pp.~286--318.

\bibitem{bertalmio2001navier}
{\sc M.~Bertalmio, A.~L. Bertozzi, and G.~Sapiro}, {\em Navier-{S}tokes, fluid
  dynamics, and image and video inpainting}, in Proceedings of the 2001 IEEE
  Computer Society Conference on Computer Vision and Pattern Recognition. CVPR
  2001, vol.~1, IEEE, 2001, pp.~I--I.

\bibitem{bertalmio2000image}
{\sc M.~Bertalmio, G.~Sapiro, V.~Caselles, and C.~Ballester}, {\em Image
  inpainting}, in Proceedings of the 27th annual conference on Computer
  graphics and interactive techniques, 2000, pp.~417--424.

\bibitem{bertozzi2007analysis}
{\sc A.~Bertozzi, S.~Esedoglu, and A.~Gillette}, {\em Analysis of a two-scale
  {C}ahn--{H}illiard model for binary image inpainting}, Multiscale Modeling \&
  Simulation, 6 (2007), pp.~913--936.

\bibitem{bertozzi2006inpainting}
{\sc A.~L. Bertozzi, S.~Esedoglu, and A.~Gillette}, {\em Inpainting of binary
  images using the {C}ahn--{H}illiard equation}, IEEE Transactions on image
  processing, 16 (2006), pp.~285--291.

\bibitem{bessemoulin2012finite}
{\sc M.~Bessemoulin-Chatard and F.~Filbet}, {\em A finite volume scheme for
  nonlinear degenerate parabolic equations}, SIAM Journal on Scientific
  Computing, 34 (2012), pp.~B559--B583.

\bibitem{bosch2014fast}
{\sc J.~Bosch, D.~Kay, M.~Stoll, and A.~J. Wathen}, {\em Fast solvers for
  {C}ahn--{H}illiard inpainting}, SIAM Journal on Imaging Sciences, 7 (2014),
  pp.~67--97.

\bibitem{bosch2015fractional}
{\sc J.~Bosch and M.~Stoll}, {\em A fractional inpainting model based on the
  vector-valued {C}ahn--{H}illiard equation}, SIAM Journal on Imaging Sciences,
  8 (2015), pp.~2352--2382.

\bibitem{burger2009cahn}
{\sc M.~Burger, L.~He, and C.-B. Sch{\"o}nlieb}, {\em {C}ahn--{H}illiard
  inpainting and a generalization for grayvalue images}, SIAM Journal on
  Imaging Sciences, 2 (2009), pp.~1129--1167.

\bibitem{cahn1958free}
{\sc J.~W. {C}ahn and J.~E. {H}illiard}, {\em Free energy of a nonuniform
  system. i. interfacial free energy}, The Journal of chemical physics, 28
  (1958), pp.~258--267.

\bibitem{carrillo2015finite}
{\sc J.~A. Carrillo, A.~Chertock, and Y.~Huang}, {\em A finite-volume method
  for nonlinear nonlocal equations with a gradient flow structure},
  Communications in Computational Physics, 17 (2015), pp.~233--258.

\bibitem{caselles1998axiomatic}
{\sc V.~Caselles, J.-M. Morel, and C.~Sbert}, {\em An axiomatic approach to
  image interpolation}, IEEE Transactions on image processing, 7 (1998),
  pp.~376--386.

\bibitem{cates2005physical}
{\sc M.~Cates, J.-C. Desplat, P.~Stansell, A.~Wagner, K.~Stratford,
  R.~Adhikari, and I.~Pagonabarraga}, {\em Physical and computational scaling
  issues in lattice {B}oltzmann simulations of binary fluid mixtures},
  Philosophical Transactions of the Royal Society A: Mathematical, Physical and
  Engineering Sciences, 363 (2005), pp.~1917--1935.

\bibitem{chen2019positivity}
{\sc W.~Chen, C.~Wang, X.~Wang, and S.~M. Wise}, {\em Positivity-preserving,
  energy stable numerical schemes for the {C}ahn-{H}illiard equation with
  logarithmic potential}, Journal of Computational Physics: X, 3 (2019),
  p.~100031.

\bibitem{cherfils2016cahn}
{\sc L.~Cherfils, H.~Fakih, and A.~Miranville}, {\em A {C}ahn--{H}illiard
  system with a fidelity term for color image inpainting}, Journal of
  Mathematical Imaging and Vision, 54 (2016), pp.~117--131.

\bibitem{cherfils2017complex}
\leavevmode\vrule height 2pt depth -1.6pt width 23pt, {\em A complex version of
  the {C}ahn--{H}illiard equation for grayscale image inpainting}, Multiscale
  Modeling \& Simulation, 15 (2017), pp.~575--605.

\bibitem{choksi2009phase}
{\sc R.~Choksi, M.~A. Peletier, and J.~Williams}, {\em On the phase diagram for
  microphase separation of diblock copolymers: an approach via a nonlocal
  {C}ahn--{H}illiard functional}, SIAM Journal on Applied Mathematics, 69
  (2009), pp.~1712--1738.

\bibitem{choo2005discontinuous}
{\sc S.~Choo and Y.~Lee}, {\em A discontinuous {G}alerkin method for the
  {C}ahn-{H}illiard equation}, Journal of Applied Mathematics and Computing, 18
  (2005), pp.~113--126.

\bibitem{coppersmith1987matrix}
{\sc D.~Coppersmith and S.~Winograd}, {\em Matrix multiplication via arithmetic
  progressions}, in Proceedings of the nineteenth annual ACM symposium on
  Theory of computing, 1987, pp.~1--6.

\bibitem{criminisi2003object}
{\sc A.~Criminisi, P.~Perez, and K.~Toyama}, {\em Object removal by
  exemplar-based inpainting}, in 2003 IEEE Computer Society Conference on
  Computer Vision and Pattern Recognition, 2003. Proceedings., vol.~2, IEEE,
  2003, pp.~II--II.

\bibitem{criminisi2004region}
{\sc A.~Criminisi, P.~P{\'e}rez, and K.~Toyama}, {\em Region filling and object
  removal by exemplar-based image inpainting}, IEEE Transactions on image
  processing, 13 (2004), pp.~1200--1212.

\bibitem{cueto2008time}
{\sc L.~Cueto-Felgueroso and J.~Peraire}, {\em A time-adaptive finite volume
  method for the {C}ahn--{H}illiard and kuramoto--sivashinsky equations},
  Journal of Computational Physics, 227 (2008), pp.~9985--10017.

\bibitem{deng2012mnist}
{\sc L.~Deng}, {\em The {MNIST}database of handwritten digit images for machine
  learning research}, IEEE Signal Processing Magazine, 29 (2012), pp.~141--142.

\bibitem{elliott1989cahn}
{\sc C.~M. Elliott}, {\em The {C}ahn-{H}illiard model for the kinetics of phase
  separation}, in Mathematical models for phase change problems, Springer,
  1989, pp.~35--73.

\bibitem{eyre1998unconditionally}
{\sc D.~J. Eyre}, {\em An unconditionally stable one-step scheme for gradient
  systems}, Unpublished article,  (1998), pp.~1--15.

\bibitem{furihata2001stable}
{\sc D.~Furihata}, {\em A stable and conservative finite difference scheme for
  the {C}ahn-{H}illiard equation}, Numerische Mathematik, 87 (2001),
  pp.~675--699.

\bibitem{gu2006x}
{\sc J.~Gu, L.~Zhang, G.~Yu, Y.~Xing, and Z.~Chen}, {\em X-ray ct metal
  artifacts reduction through curvature based sinogram inpainting}, Journal of
  X-ray Science and Technology, 14 (2006), pp.~73--82.

\bibitem{guo2016h2}
{\sc J.~Guo, C.~Wang, S.~M. Wise, and X.~Yue}, {\em An h2 convergence of a
  second-order convex-splitting, finite difference scheme for the
  three-dimensional {C}ahn-{H}illiard equation}, Commun. Math. Sci, 14 (2016),
  pp.~489--515.

\bibitem{lee2012robust}
{\sc J.~Lee, D.-K. Lee, and R.-H. Park}, {\em Robust exemplar-based inpainting
  algorithm using region segmentation}, IEEE Transactions on Consumer
  Electronics, 58 (2012), pp.~553--561.

\bibitem{liu2015stabilized}
{\sc F.~Liu and J.~Shen}, {\em Stabilized semi-implicit spectral deferred
  correction methods for allen--{C}ahn and {C}ahn--{H}illiard equations},
  Mathematical Methods in the Applied Sciences, 38 (2015), pp.~4564--4575.

\bibitem{masnou1998level}
{\sc S.~Masnou and J.-M. Morel}, {\em Level lines based disocclusion}, in
  Proceedings 1998 International Conference on Image Processing. ICIP98 (Cat.
  No. 98CB36269), IEEE, 1998, pp.~259--263.

\bibitem{GithubSergio}
{\sc S.~P. Perez}, {\em Code repository to reproduce the results of this work}.
\newblock https://github.com/sergiopperez/Image\_Inpainting, 2021.

\bibitem{schmuck2012upscaled}
{\sc M.~Schmuck, M.~Pradas, G.~A. Pavliotis, and S.~Kalliadasis}, {\em Upscaled
  phase-field models for interfacial dynamics in strongly heterogeneous
  domains}, Proceedings of the Royal Society A: Mathematical, Physical and
  Engineering Sciences, 468 (2012), pp.~3705--3724.

\bibitem{schonlieb2015partial}
{\sc C.-B. Sch{\"o}nlieb}, {\em Partial differential equation methods for image
  inpainting}, vol.~29, Cambridge University Press, 2015.

\bibitem{shen2019new}
{\sc J.~Shen, J.~Xu, and J.~Yang}, {\em A new class of efficient and robust
  energy stable schemes for gradient flows}, SIAM Review, 61 (2019),
  pp.~474--506.

\bibitem{tierra2015numerical}
{\sc G.~Tierra and F.~Guill{\'e}n-Gonz{\'a}lez}, {\em Numerical methods for
  solving the {C}ahn--{H}illiard equation and its applicability to related
  energy-based models}, Archives of Computational Methods in Engineering, 22
  (2015), pp.~269--289.

\bibitem{vollmayr2003fast}
{\sc B.~P. Vollmayr-Lee and A.~D. Rutenberg}, {\em Fast and accurate coarsening
  simulation with an unconditionally stable time step}, Physical Review E, 68
  (2003), p.~066703.

\bibitem{wells2006discontinuous}
{\sc G.~N. Wells, E.~Kuhl, and K.~Garikipati}, {\em A discontinuous {G}alerkin
  method for the {C}ahn--{H}illiard equation}, Journal of Computational
  Physics, 218 (2006), pp.~860--877.

\bibitem{wise2009energy}
{\sc S.~M. Wise, C.~Wang, and J.~S. Lowengrub}, {\em An energy-stable and
  convergent finite-difference scheme for the phase field crystal equation},
  SIAM Journal on Numerical Analysis, 47 (2009), pp.~2269--2288.

\bibitem{wu2014stabilized}
{\sc X.~Wu, G.~Van~Zwieten, and K.~Van~der Zee}, {\em Stabilized second-order
  convex splitting schemes for {C}ahn--{H}illiard models with application to
  diffuse-interface tumor-growth models}, International journal for numerical
  methods in biomedical engineering, 30 (2014), pp.~180--203.

\bibitem{Marc2012}
{\sc C.~Wylock, M.~Pradas, B.~Haut, P.~Colinet, and S.~Kalliadasis}, {\em
  Disorder-induced hysteresis and nonlocality of contact line motion in
  chemically heterogeneous environments}, Phys. Fluids, 24 (2012), p.~032108.

\bibitem{xia2007local}
{\sc Y.~Xia, Y.~Xu, and C.-W. Shu}, {\em Local discontinuous {G}alerkin methods
  for the {C}ahn--{H}illiard type equations}, Journal of Computational Physics,
  227 (2007), pp.~472--491.

\bibitem{xie2012image}
{\sc J.~Xie, L.~Xu, and E.~Chen}, {\em Image denoising and inpainting with deep
  neural networks}, in Advances in neural information processing systems, 2012,
  pp.~341--349.

\bibitem{yeh2017semantic}
{\sc R.~A. Yeh, C.~Chen, T.~Yian~Lim, A.~G. Schwing, M.~Hasegawa-Johnson, and
  M.~N. Do}, {\em Semantic image inpainting with deep generative models}, in
  Proceedings of the IEEE conference on computer vision and pattern
  recognition, 2017, pp.~5485--5493.

\bibitem{yu2018generative}
{\sc J.~Yu, Z.~Lin, J.~Yang, X.~Shen, X.~Lu, and T.~S. Huang}, {\em Generative
  image inpainting with contextual attention}, in Proceedings of the IEEE
  conference on computer vision and pattern recognition, 2018, pp.~5505--5514.

\end{thebibliography}
	
	\printindex

\end{document}